%% file: sample-acmtog-SIGGRAPH-submission.tex
\documentclass[acmtog,authorversion]{acmart}

\usepackage{booktabs}
\usepackage{subcaption}
\usepackage{multirow}
\usepackage{xcolor}
\usepackage{soul}
\usepackage{xcolor,colortbl}
\usepackage{listings}

\definecolor{orange}{HTML}{FF8000}
\definecolor{red}{HTML}{FF0000}
\definecolor{blue}{HTML}{004C99}

\usepackage[ruled]{algorithm2e} 

\SetAlFnt{\small}
\SetAlCapFnt{\small}
\SetAlCapNameFnt{\small}
\SetAlCapHSkip{0pt}

\makeatletter
\let\@authorsaddresses\@empty
\makeatother

\copyrightyear{2025}
\acmYear{2025}
\setcopyright{rightsretained}
\acmConference[SIGGRAPH Conference Papers '25]{Special Interest Group on Computer Graphics and Interactive Techniques Conference Conference Papers }{August 10--14, 2025}{Vancouver, BC, Canada}
\acmBooktitle{Special Interest Group on Computer Graphics and Interactive Techniques Conference Conference Papers (SIGGRAPH Conference Papers '25), August 10--14, 2025, Vancouver, BC, Canada}
\acmDOI{10.1145/3721238.3730676}
\acmISBN{979-8-4007-1540-2/2025/08}

\begin{document}
\title{Spline Deformation Field}

\author{Mingyang Song}
\affiliation{%
 \institution{DisneyResearch|Studios}
 \country{Switzerland}}
\affiliation{%
 \institution{ETH Zürich}
 \country{Switzerland}}
\orcid{0009-0002-5354-7045}
\email{mingyang.song@inf.eth.ch}

\author{Yang Zhang}
\affiliation{%
 \institution{DisneyResearch|Studios}
 \country{Switzerland}}
\orcid{0000-0002-2381-6067}
\email{yang.zhang@disneyresearch.com}

\author{Marko Mihajlovic}
\affiliation{%
 \institution{ETH Zürich}
 \country{Switzerland}}
\orcid{0000-0001-6305-3896}
\email{marko.mihajlovic@inf.ethz.ch}

\author{Siyu Tang}
\affiliation{%
 \institution{ETH Zürich}
 \country{Switzerland}}
\orcid{0000-0002-1015-4770}
\email{siyu.tang@inf.ethz.ch}

\author{Markus Gross}
\affiliation{%
 \institution{DisneyResearch|Studios}
 \country{Switzerland}}
\affiliation{%
 \institution{ETH Zürich}
 \country{Switzerland}}
\orcid{0009-0003-9324-779X}
\email{gross@disneyresearch.com}

\author{Tun\c{c} Ozan Ayd{\i}n}
\affiliation{%
 \institution{DisneyResearch|Studios}
 \country{Switzerland}}
\orcid{0009-0002-0415-1415}
\email{tunc@disneyresearch.com}

\begin{abstract}
Trajectory modeling of dense points usually employs implicit deformation fields, represented as neural networks that map coordinates to relate canonical spatial positions to temporal offsets. However, the inductive biases inherent in neural networks can hinder spatial coherence in ill-posed scenarios. Current methods focus either on enhancing encoding strategies for deformation fields, often resulting in opaque and less intuitive models, or adopt explicit techniques like linear blend skinning, which rely on heuristic-based node initialization. Additionally, the potential of implicit representations for interpolating sparse temporal signals remains under-explored. To address these challenges, we propose a spline-based trajectory representation, where the number of knots explicitly determines the degrees of freedom. This approach enables efficient analytical derivation of velocities, preserving spatial coherence and accelerations, while mitigating temporal fluctuations. To model knot characteristics in both spatial and temporal domains, we introduce a novel low-rank time-variant spatial encoding, replacing conventional coupled spatiotemporal techniques. Our method demonstrates superior performance in temporal interpolation for fitting continuous fields with sparse inputs. Furthermore, it achieves competitive dynamic scene reconstruction quality compared to state-of-the-art methods while enhancing motion coherence without relying on linear blend skinning or as-rigid-as-possible constraints.
\end{abstract}

\begin{teaserfigure}
  \includegraphics[width=\textwidth]{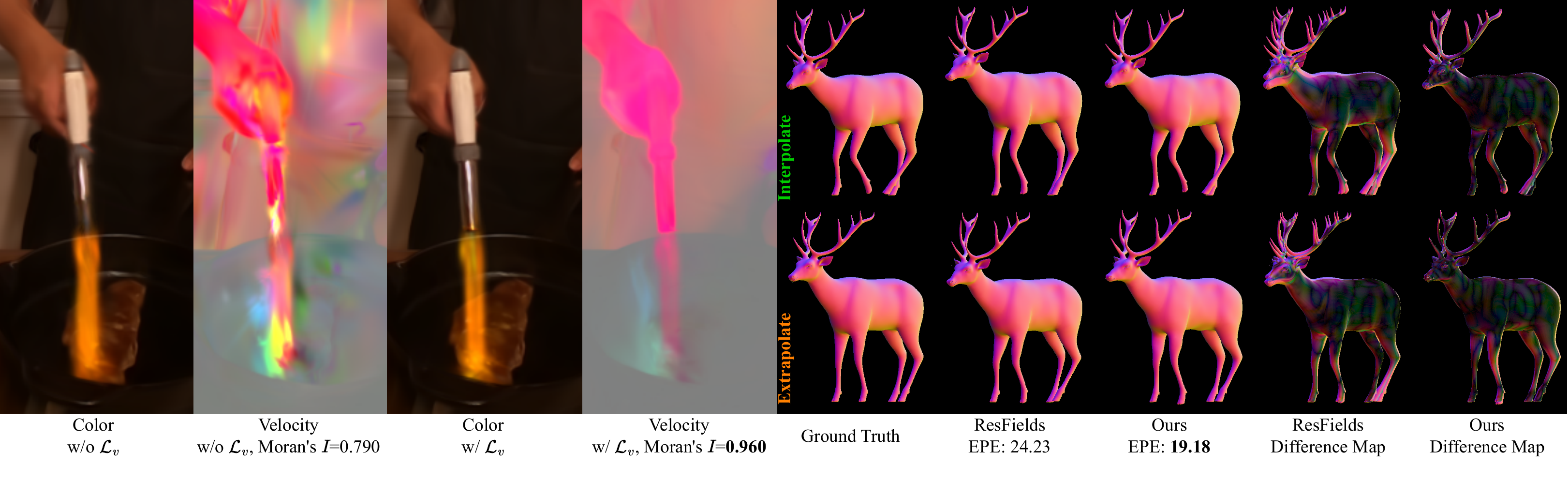}
  \caption{We represent trajectories with splines, in which analytically derived velocity \(v\) enables effective spatial coherency preservation through our novel loss term \(\mathcal{L}_v\). Furthermore, we propose a metric concepting on Moran's \textit{I} to quantify it. Spline interpolation can generate fair interpolated and extended motions, compared to purely relying on the smoothness inductive bias of the implicit deformation field, such as ResFields~\cite{mihajlovic2024ResFields}.}
  \label{fig:teaser}
\end{teaserfigure}

\maketitle

\input{samplebody-journals}

\end{document}

%% file: samplebody-journals.tex
\section{Introduction \& Related Work}

The demand for precise and computationally efficient trajectory modeling has significantly increased in fields such as character animation, motion capture, and 3D reconstruction.
Traditionally, motion modeling was achieved through explicit constraints or handcrafted priors~\cite{sorkine2007rigid, huang2008non, kilian2007geometric, innmann2020nrmvs, eisenberger2018divergence, myronenko2010point}, ensuring spatial coherence and adherence to physical laws. However, such priors are not always practical, particularly when dealing with sparse or noisy data.

To address these limitations, neural network-based deformation modeling approaches have emerged, offering task-agnostic, differentiable, and continuous representations. Among these, coordinate neural networks (Coord.-NN) have gained prominence for their ability to encode continuous signals directly as network weights using query coordinates as input. This is typically achieved through a compact implicit function comprising a 3+1D encoding followed by a Multi-Layer Perceptron (MLP). Coord.-NNs generally produce reasonably smooth results regardless of the encoding scheme (e.g., sinusoidal, dense grid, hash grid), benefiting from an inherent smoothness inductive bias. This bias can be attributed to the continuity of MLP, the smoothness of activation functions, the local coherence of multi-resolution grids interpolation, or their interplay. However, while Coord.-NNs are widely used for modeling various types of temporal changes, it remains unclear which source of bias offers the best trade-off between performance and adaptability across different scenarios.

Recent Dynamic Point Fields (DPF)~\cite{Prokudin_2023_ICCV} leverages the intrinsic regularization power of SIREN~\cite{sitzmann2020implicit} to map unregistered point pairs. DOMA~\cite{zhang2024degrees} extends this further to multi-frame sequences by incorporating time conditioning. While this approach achieved improved temporal consistency compared with per-frame modeling, the coupled 4D signals introduced potential wavy fluctuations. ResFields~\cite{mihajlovic2024ResFields} mitigated such biases by disentangling spatial and temporal signals, enhancing the capacity of MLPs for spatiotemporal signal modeling. However, its reliance on temporal discretization limited its interpolation performance. Constant canonical points and direct supervision are common characteristics across the aforementioned tasks. As a result, SIREN~\cite{sitzmann2020implicit}, with its higher nonlinearity and representational capacity, is typically preferred over sinusoidal positional encoding in such more-constrained scenarios.


In scenarios where the points to be deformed are also optimization targets, such as dynamic scene reconstruction with Gaussian Splatting (GS)~\cite{kerbl20233d}, Coord.-NN's inductive bias preserves spatial consistency fairly for free. In such cases, the canonical positions of 3D Gaussians are optimized alongside the deformation field. 4DGS~\cite{wu20234d} represented the deformation field using Hexplane~\cite{cao2023hexplane, fridovich2023k}, achieving strong photometric results on multi-view datasets. However, it suffered from floaters due to the limited regularization capacity of the total variation loss. Grid4D~\cite{xu2024grid4d} addressed overlapping issues inherent to Hexplanes and introduced a perturbation loss, effectively reducing artifacts in grid-based methods. Similar to D-NeRF~\cite{pumarola2021d}, Deformable3DGS (Def.GS)~\cite{yang2023deformable} utilized MLP to predict offsets, whose global continuity yielded better performance on synthetic datasets, even without extensive regularization. SC-GS~\cite{huang2023sc} tackled the limited rigidity of Def.GS by introducing the concept of sparse-controlled GS, which relies on sampling sparse control nodes from the scene trained in the first stage. Moreover, the ARAP constraints used in SC-GS~\cite{huang2023sc}, which originate from meshes with fixed vertex connectivity, may not perform effectively in large-scale scenes where densification causes significant changes in local neighborhoods.ReMatching~\cite{oblak2024rematching} adopted deformation priors to enhance spatial coherency in a single stage. While effective, the heuristics introduced by multi-stage training and prior choice limit generalization ability.

Inspired by previous works~\cite{chugunov2023neural, knodt2022continuous}, we propose representing trajectories using splines, a classic tool from applied mathematics, that naturally model continuous motion over time. With the timeline discretized by knots, we concept on the design of ResFileds~\cite{mihajlovic2024ResFields}, which decouples spatial and temporal signals, and extend this approach to multiple existing encoding schemes. Our experiments demonstrate that explicit spline interpolation outperforms deep feature interpolation, particularly in scenarios with sparse temporal signals. More importantly, the analytical derivation of kinematic properties from the interpolation function enables effective regularization of spatial coherence and reduces temporal jitter without requiring two-stage training or reliance on priors, thus reducing heuristics in the training process. Additionally, we introduce a novel measurement for spatial coherence that aligns well with other metric evaluations and visual reconstruction results. Finally, we analyze the limitations of our representation, offering insights that may guide future research in this area.

\section{Method}
\begin{figure*}
    \centering
    \includegraphics[width=0.82\linewidth]{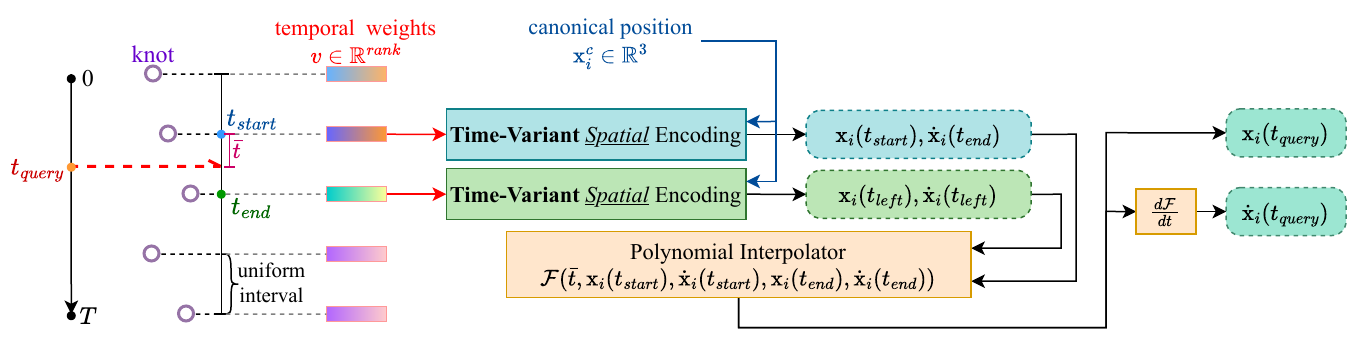}
    \caption{The pipeline of our method. We explain how to query value at arbitrary timestamps through polynomial interpolation in Sec.~\ref{sec:spline_interp}, corresponding to the timelines in the left part and \textcolor{orange}{orange} block. We describe our design of spatiotemporal conditioning in Sec.~\ref{sec:st_cond}, which corresponds to the \textcolor{red}{red} (temporal signal) and \textcolor{blue}{blue} (spatial signal) arrows.}
    \label{fig:frame_work}
\end{figure*}

\subsection{Spline Interpolation}
\label{sec:spline_interp}
A trajectory from \(t=0\) to \(t=1\) is uniformly divided into \(N-1\) time intervals, resulting in \(N\) knots. Unlike global fitting methods that utilize a single polynomial, Fourier series, or their combination~\cite{lin2024gaussian}, the localized nature of splines minimizes potential temporal oscillations, particularly in static regions. To prevent underfitting, the choice of \(N\) must correspond to the number of input temporal steps, ensuring a well-determined system. Specifically, for \(T\) training timestamps, the number of parameters introduced by \(N\) knots must satisfy the following equation:
\begin{equation}
    K\cdot N=T\\,
    \label{eq:determined_eq}
\end{equation}
where \(K\) is a factor determined by the order of the polynomial. Here, we focus on the Cubic Hermite Spline~\cite{wiki:Cubic_Hermite_spline}, which represents a trajectory using consecutive third-order polynomial segments. For a given query timestamp \(t_{query}\) and \(N\) knots, we first locate the time interval to which it belongs and identify the corresponding starting and ending knots, denoted as \(t_{start}\) and \(t_{end}\), respectively. The interval is then normalized to a relative time \(\bar{t}\in[0, 1]\). The interpolation function for each segment is defined as:
\begin{equation}
    \begin{array}{l}
        p(\bar{t})=(2\bar{t}^3-3\bar{t}^2+1)p_0+(\bar{t}^3-2\bar{t}^2+\bar{t})m_0+\\
        \qquad \quad (-2\bar{t}^3+3\bar{t}^2)p_1+(\bar{t}^3-\bar{t}^2)m_1,
    \end{array}
    \label{eq:cubic_spline}
\end{equation}
where \(p_0\) and \(p_1\) represent the properties at the starting and ending knots, while \(m_0\) and \(m_1\) are the corresponding starting and ending tangents. 
We treat the tangents as independent optimizable parameters.
We follow Eq.~\ref{eq:determined_eq} and set \(K=2\), which corresponds to choosing \(N=T/2\), thereby guaranteeing a theoretically well-determined fit. 

To preserve spatial coherency, we employ a Canonical-Deformation design, 
where the parameters \(p\) and \(m\) of points are predicted by a Coord.-NN using spatial coordinates in the canonical space as inputs. This process can be expressed as:
\begin{equation}
    \begin{array}{l}
        \textbf{\text{X}}^{c}=\{\text{x}_i^c,\text{x}_i^c\in\mathbb{R}^3\}_{i=1, ..., N_p}, \\
        \text{x}_i(t_{query})=\mathcal{F}(\bar{t},\text{x}_i(t_{start}),\dot{\text{x}}_i(t_{start}),\text{x}_i(t_{end}),\dot{\text{x}}_i(t_{end})),\\
        \Delta\text{x}_i(t_{start}), \dot{\text{x}}_i(t_{start})=\Phi_{\theta}(\text{x}_i^c,t_{start}),\\
        \text{x}_i(t_{start})=\text{x}_i^c+\Delta\text{x}_i(t_{start}),
    \end{array}
\end{equation}
where \(\text{x}_i^c\) denotes the spatial coordinates of points in the canonical space, \(N_p\) is the total number of points, and \(\mathcal{F}(\cdots)\) represents the interpolation function described in Eq.~\ref{eq:cubic_spline}. 
To enhance clarity, we substitute \(p_0\) with \(\text{x}_i(t_{start})\) and \(m_0\) with \(\dot{\text{x}}_i(t_{start})\). For brevity, we omit the derivation of \(t_{end}\) here and include it in the supp. material. 


\subsection{Time-variant Spatial Encoding (TVSE)}
\label{sec:st_cond}
Previous works~\cite{yang2023deformable, wu20234d, huang2023sc, xu2024grid4d, fridovich2023k, pumarola2021d} treat four-dimensional inputs (three spatial coordinates and one timestamp) equally. Recently, ResFields~\cite{mihajlovic2024ResFields} introduced a novel approach by injecting temporal information into the weights of pure MLP-based spatial encodings through time-variant residual and base weights, enhancing the performance of various previous methods. Its architecture-agnostic design naturally extends to other encoding schemes. The key conceptual difference can expressed as:
\begin{equation}
    \begin{array}{l}
        \text{Previous work: }\Phi_{\theta}(\gamma(x),\gamma(y),\gamma(z),\gamma(t)),\\
        \text{Ours: }\Phi_{\theta}(\gamma_{\varphi(t)}(x),\gamma_{\varphi(t)}(y),\gamma_{\varphi(t)}(z)),
    \end{array}
    \label{eq:res_weights}
\end{equation}
where \(\gamma(\cdot)\) represents the encoding (e.g., Sinusoidal~\cite{mildenhall2021nerf} or grids~\cite{muller2022instant, cao2023hexplane}), \(\gamma_{\varphi(t)}(\cdot)\) denotes time-variant encoding, and \(\varphi(t)\) is the temporal signal injection function, which varies based on the specific design of \(\gamma(\cdot)\). 

\begin{figure}
    \centering
    \includegraphics[width=0.9\linewidth]{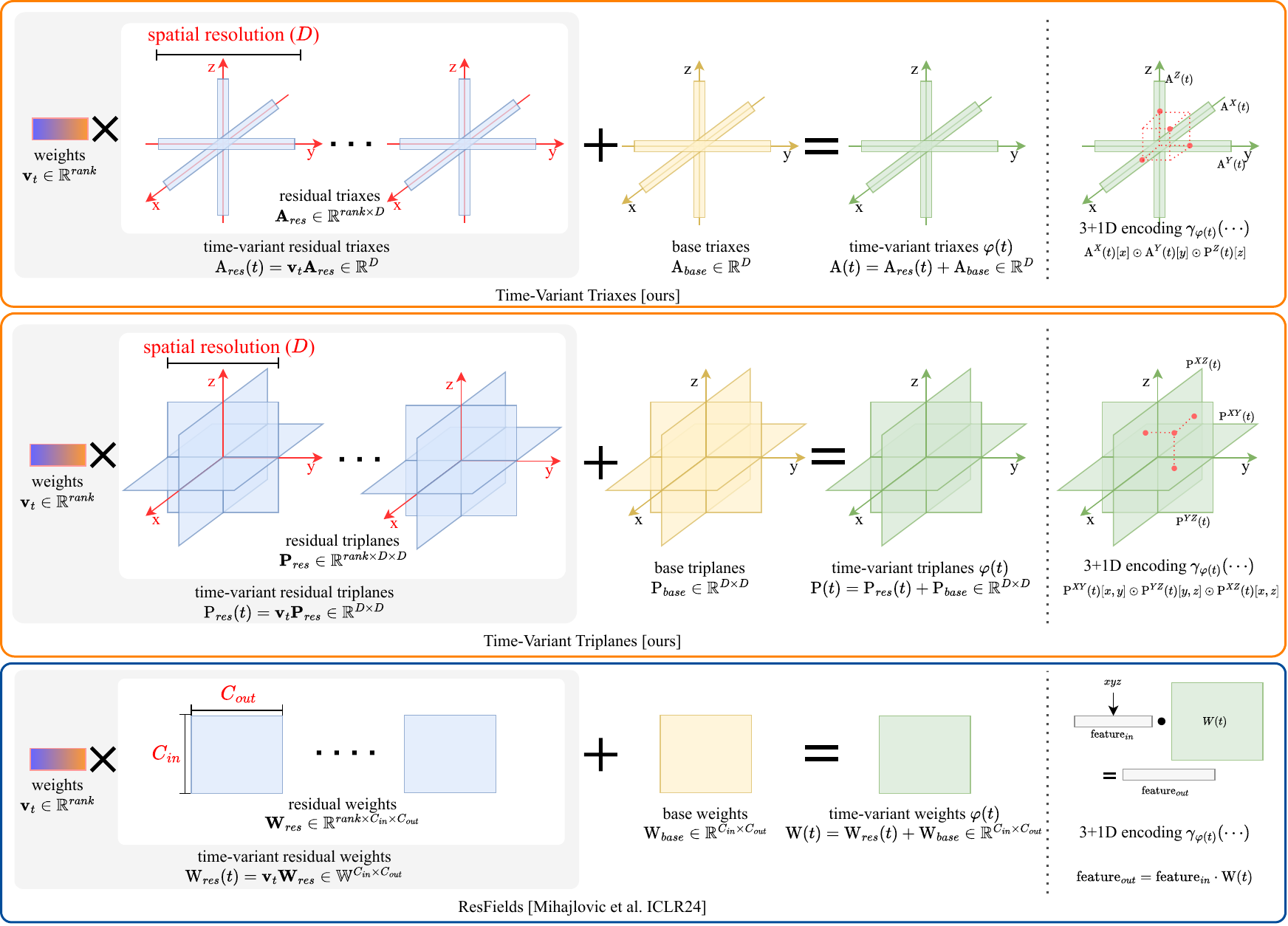}
    \caption{An intuitive diagram of our TVSE. Rounded boxes with different edge colors mark \textcolor{orange}{grid-based} and \textcolor{blue}{MLP} and methods. \(A^{\cdot}(t)[\cdot]\) and \(P^{\cdot\cdot}(t)[\cdot,\cdot]\) denote sampling values at specific position from an axis and plane, respectively. Following \cite{fridovich2023k}, we aggregate features sampled from the three axes and planes using element-wise multiplication \(\odot\).}
    \label{fig:time_variant_encode}
\end{figure}

Building on the concept of ResFields~\cite{mihajlovic2024ResFields}, we incorporate low-rank decomposition in the temporal domain to achieve both compactness and implicit regularization. This approach can be expressed as: 
\begin{equation}
    \phi(t)=\text{b}_{base}+\sum_{r=1}^{rank}\text{v}_t[r]\cdot\text{B}_{res}[r],
    \label{eq:phi_t}
\end{equation}
where \([r]\) denotes element indexing, \(\textbf{v}_t\in\mathbb{R}^{rank}\) represents trainable weights associated with each timestep, \(\textbf{B}_{res}\in\mathbb{R}^{rank\times\cdots}\) are the residual bases, and \(\text{b}_{base}\) corresponds to the time-invariant encoding. These notations carry different meanings depending on the chosen encoding scheme. In this work, we primarily focus on cosine functions and multi-resolution dense grids.
Fig.~\ref{fig:time_variant_encode} provides an intuitive visualization of the design of \(\gamma_{\varphi(t)}\). We refer to the supplementary material for detailed implementations.

\subsection{Velocity and Acceleration Regularization}
\label{sec:vel_acc_reg}
By taking derivative w.r.t time in Eq.~\ref{eq:cubic_spline}, the velocity of a point on the spline can be determined by the following function:
\begin{equation}
    \begin{array}{l}
        v(\bar{t})=(6\bar{t}^2-6\bar{t})p_0+(3\bar{t}^2-4\bar{t}+1)m_0+ \\
        \qquad \quad (-6\bar{t}^2+6\bar{t})p_1+(3\bar{t}^2-2\bar{t})m_1,
    \end{array}
    \label{eq:velocity}
\end{equation}
where \(v(\bar{t})\) has physical meaning, i.e., velocity.
With the above closed-form velocity function, the velocity loss can be formulated as:
\begin{equation}
    \mathcal{L}_v^i=\sum_{j\in\mathcal{N}_k(i)}w_{ij}||v_i-v_j||_2^2,
    \label{eq:vel_loss}
\end{equation}
where \(i\) is the index of points, \(\mathcal{N}_k(i)\) represents the \(k\) nearest neighbors of \(i\), and \(j\) is the local index in neighborhood, \(w_{ij}\) is weight calculated from relative distance.

We propose one additional constraint to reduce high-frequency jitter and alleviate low-frequency oscillation. Specifically, taking derivation w.r.t time in Eq.~\ref{eq:velocity}, we can get the analytical acceleration:
\begin{equation}
    a(\bar{t})=(12\bar{t}-6)p_0+(6\bar{t}-4)m_0+(-12\bar{t}+6)p_1+(6\bar{t}-2)m_1.
    \label{eq:acceleration}
\end{equation}
Then, we regularize the acceleration of points through the following loss function:
\begin{equation}
    \mathcal{L}_{acc}^i=|a_i|.
    \label{eq:acc_loss}
\end{equation}
Finally, the loss function is:
\begin{equation}
    \mathcal{L}=\mathcal{L}_{recon}+\alpha \mathcal{L}_v+\beta \mathcal{L}_{acc},
    \label{eq:final_loss}
\end{equation}
where \(\mathcal{L}_{recon}\) represents supervisions from ground truth to prediction, \(\alpha\) and \(\beta\) are hyperparameters for regularization.

\section{Results}
\subsection{Learning Continuous Deformation Field}
\label{sce:continuous_deformation_field}
\begin{figure*}
    \includegraphics[width=0.85\linewidth]{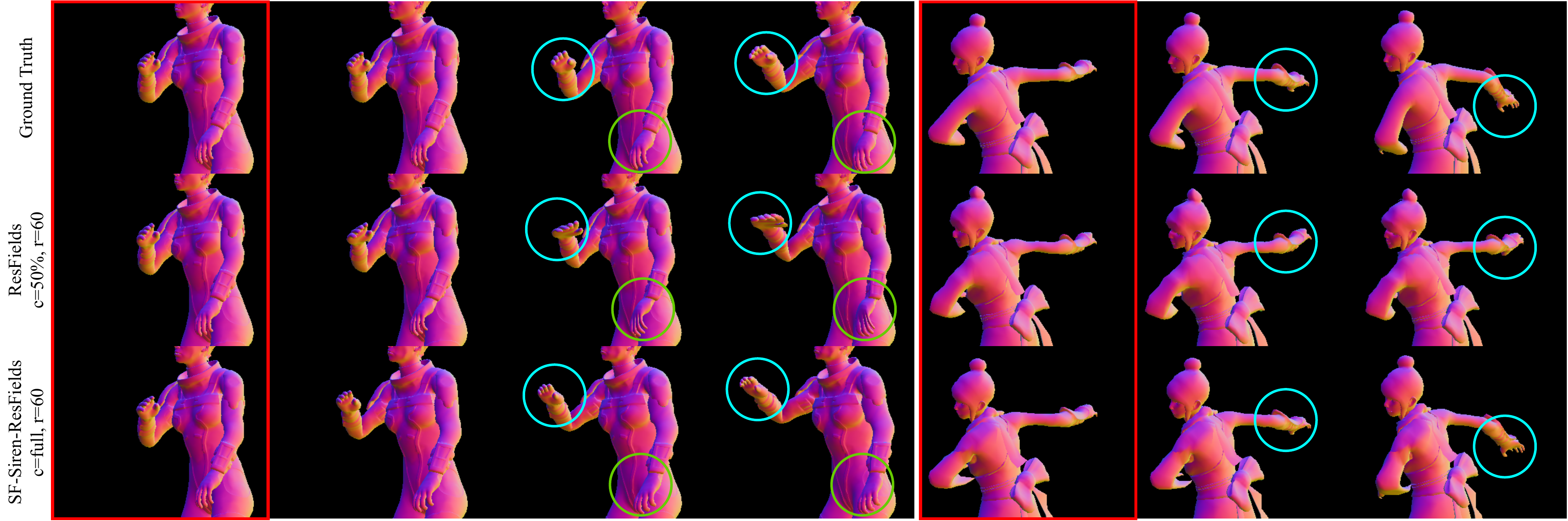}
    \caption{Qualitative comparison of \(\times 6\) (left) and \(\times 4\) (right) scene flow interpolation. Frames boxed in red are training timesteps, and the following ones are interpolated. Our method achieves fair interpolated motions without skinning. We provide more visual results in the extra pages.}
    \label{fig:scene_flow_interp_main}
\end{figure*}

As a proof of concept, we start with a more constrained problem. Given sparsely sampled point trajectories, the goal is to fit an implicit continuous deformation field that can infer the trajectories of unseen points, ensuring alignment with observed data. We focus on a more challenging yet practical scenario where available temporal signals are scarce to demonstrate the strength of explicit spline interpolation. Specifically, we choose nine long sequences (each exceeding 500 frames) from DeformingThings4D (humanoids)~\cite{li20214dcomplete} and train models with every 4th/6th timestep, leaving 75\%/83.3\% timesteps of timesteps for evaluation. Following DOMA~\cite{zhang2024degrees}, 25\% vertices from the starting mesh are sampled to calculate L1 distance for supervision during training. The determination of \(N\) follows Eq.~\ref{eq:determined_eq}. In addition to end point error (EPE), we propose a new metric for evaluating spatial coherence using Moran's \textit{I}~\cite{moran1950notes, mihajlovic2025splatfields}. We provide its detailed implementation in the supplementary material.

We benchmark our method against state-of-the-art approaches in scene flow estimation, as detailed in Tab.~\ref{tab:scene_flow_interp}. DOMA~\cite{zhang2024degrees} employs periodical activation layers, and the coupled spatial and temporal signals introduce significant temporal fluctuation artifacts, which also impairs spatial performance. This issue cannot be resolved simply by adjusting the model's number of parameters. ResFields~\cite{mihajlovic2024ResFields} disentangles time and position coordinates and results in clear improvements, motivating our design in Sec.~\ref{sec:st_cond}. Its performance can be further refined by adjusting temporal capacity and rank or adapting as-isometric-as-possible (AIAP) regularization~\cite{Prokudin_2023_ICCV,qian20243dgs} (besed on setting \(\mathcal{y}\)). For example, c=50\% indicates that the number of temporal weights, \(v\in\mathbb{R}^{rank}\), is reduced to half the number of training timesteps, thereby decreasing the temporal DoF. However, motions generated by interpolating deep features may still exhibit distortions, as shown in Fig.~\ref{fig:scene_flow_interp_main}. In contrast, our method relies on explicit spline interpolation and achieves better performance with c=100\%, reducing the risk of underfitting. Based on setting c=100\%, r=60 (marked as \(\mathcal{z}\)), we conduct ablation studies on \(\mathcal{L}_{v}\), \(\mathcal{L}_{acc}\) and AIAP, and the relative performance shows their effectiveness.

This explicitness also ensures intuitive feedback when adjusting hyperparameters: reducing \(N\) in such large-motion dataset can lead to large misalignment that significantly penalizes reconstruction accuracy. 
The following section shows a practical case where our explicit constraints on \(N\) enhance reconstruction performance by leveraging motion priors.

\begin{table}
\centering
\small
\setlength{\tabcolsep}{2.7pt}
\caption{Interpolation of scene flow on DeformaingTings4D's long sequences. EPE results are multiplied by \(\times 10^{4}\) for better readability. The highlighted rows denote \colorbox[HTML]{F4CCCC}{1st}, \colorbox[HTML]{FCE5CD}{2nd}, and \colorbox[HTML]{FFF2CC}{3rd} best models. Note that when c=90\%, our methods still produce plausible interpolations (shown in Fig.~\ref{fig:scene_flow_interp_extra}).}
\begin{tabular}{l|cc|cc|cc}
\hline
 & \multicolumn{2}{c|}{} & \multicolumn{2}{c|}{x4} & \multicolumn{2}{c}{x6} \\
\multirow{-2}{*}{Method} & \multicolumn{2}{c|}{\multirow{-2}{*}{Settings}} & EPE\(\downarrow\) & M.'s \textit{I}\(\uparrow\) & EPE\(\downarrow\) & M.'s \textit{I}\(\uparrow\) \\ \hline
 & \multicolumn{2}{c|}{\#p=0.03M} & 138.80 & 0.912 & 146.51 & 0.911 \\
 & \multicolumn{2}{c|}{\#p=0.1M} & 92.59 & 0.876 & 105.28 & 0.877 \\
 & \multicolumn{2}{c|}{\#p=1.6M} & 473.17 & 0.249 & 452.23 & 0.254 \\
\multirow{-4}{*}{\begin{tabular}[c]{@{}l@{}}DOMA\\ \cite{zhang2024degrees}\end{tabular}} & \multicolumn{2}{c|}{\#p=2.1M} & 927.26 & 0.115 & 906.67 & 0.118 \\ \hline
 & \multicolumn{1}{c}{} & r=30 & 52.09 & 0.899 & 80.74 & 0.905 \\
 & \multicolumn{1}{c}{} & r=60 & 54.23 & 0.891 & 80.63 & 0.898 \\
 & \multicolumn{1}{c}{\multirow{-3}{*}{c=100\%}} & r=90 & 55.01 & 0.882 & 84.19 & 0.885 \\ \cline{2-7} 
 & \multicolumn{1}{c}{c=90\%} &  & 52.19 & 0.894 & 81.60 & 0.898 \\
 & \multicolumn{1}{c}{c=75\%} &  & 50.75 & 0.896 & 80.56 & 0.898 \\
 & \multicolumn{1}{c}{c=50\%}\(\mathcal{y}\) &  & 45.79 & 0.896 & 74.35 & 0.894 \\
 & \multicolumn{1}{c}{c=25\%} & \multirow{-4}{*}{r=60} & 48.60 & 0.894 & 74.92 & 0.889 \\ \cline{2-7}
 \multirow{-8}{*}{\begin{tabular}[c]{@{}l@{}}ResFields\\ \cite{mihajlovic2024ResFields}\end{tabular}} & \multicolumn{2}{c|}{\(\mathcal{y}\) w/ AIAP} & 45.72 & \cellcolor[HTML]{FCE5CD}0.927 & 73.00 & \cellcolor[HTML]{FFF2CC}0.923 \\ \hline
 & \multicolumn{1}{c}{} & r=30 & 42.98 & 0.917 & \cellcolor[HTML]{FFF2CC}68.73 & 0.927 \\
 & \multicolumn{1}{c}{} & r=60\(\mathcal{z}\) & \cellcolor[HTML]{F4CCCC}40.74 & \cellcolor[HTML]{FFF2CC}0.919 & \cellcolor[HTML]{F4CCCC}68.28 & \cellcolor[HTML]{FCE5CD}0.926 \\
 & \multicolumn{1}{c}{\multirow{-3}{*}{c=100\%}} & r=90 & \cellcolor[HTML]{FFF2CC}41.67 & 0.917 & 69.38 & 0.926 \\ \cline{2-7} 
 & \multicolumn{2}{c|}{\(\mathcal{z}\) w/o \(\mathcal{L}_{v}\)} & 42.15 & 0.916 & \cellcolor[HTML]{FCE5CD}68.40 & 0.924 \\
 & \multicolumn{2}{c|}{\(\mathcal{z}\) w/o \(\mathcal{L}_{acc}\)} & 42.66 & 0.917 & 70.12 & 0.929 \\ 
 & \multicolumn{2}{c|}{\(\mathcal{z}\) w/ AIAP} & \cellcolor[HTML]{FCE5CD}41.64 & \cellcolor[HTML]{F4CCCC}0.938 & 68.93 & \cellcolor[HTML]{F4CCCC}0.943 \\ \cline{2-7}
\multirow{-7}{*}{SF-Siren-ResFields} & \multicolumn{1}{c}{c=90\%} & r=60 & 45.97 & 0.921 & 80.22 & 0.927 \\ \hline
\end{tabular}
\label{tab:scene_flow_interp}
\end{table}

\subsection{Dynamic Scene Reconstruction with GS}
\label{sec:dy_recon}

We evaluate our method against canonical-deformation-based scene reconstruction approaches: Def.3D~\cite{yang2023deformable}, 4DGS~\cite{wu20234d}, SC-GS~\cite{huang2023sc}, and Grid4D~\cite{xu2024grid4d}. Since positional encoding performs better in such an ill-constrained problem, we use it instead of Siren in this section, which we name SF-PE-ResFields. We also implement Eq.~\ref{eq:phi_t} for grid-based methods, which we name SF-Triaxes and SF-Triplanes, and refer to supplementary material for details.

\begin{table*}
\centering
\small
\setlength{\tabcolsep}{2.0pt}
\caption{Quantitative results on NeRF-DS~\cite{yan2023nerf} dataset. Notably, traditional metrics (PSNR and SSIM) penalize sharper but misaligned results over blurry results. We introduce perturbation in Grid4D~\cite{xu2024grid4d} to 4DGS~\cite{wu20234d} since it consistently improves grid-based methods, which we denote with *. SC-GS's~\cite{huang2023sc} metrics except for M.'s \textit{I} are adopted from the official script.}
\begin{tabular}{lcccc|cccc|cccc|cccc}
\hline
 & \multicolumn{4}{c|}{As} & \multicolumn{4}{c|}{Basin} & \multicolumn{4}{c|}{Bell} & \multicolumn{4}{c}{Cup} \\
\multirow{-2}{*}{Method} & PSNR\(\uparrow\) & SSIM\(\uparrow\)  & LPIPS\(\downarrow\) & M.'s \textit{I}\(\uparrow\) & PSNR\(\uparrow\) & SSIM\(\uparrow\)  & LPIPS\(\downarrow\) & M.'s \textit{I}\(\uparrow\) & PSNR\(\uparrow\) & SSIM\(\uparrow\)  & LPIPS\(\downarrow\) & M.'s \textit{I}\(\uparrow\) & PSNR\(\uparrow\) & SSIM\(\uparrow\)  & LPIPS\(\downarrow\) & M.'s \textit{I}\(\uparrow\) \\ \hline
Def.3D~\cite{yang2023deformable} & 26.01 & 87.61 & \cellcolor[HTML]{FFF2CC}12.33 & 0.685 & 19.51 & 77.97 & \cellcolor[HTML]{F4CCCC}13.43 & 0.735 & 25.23 & 83.94 & \cellcolor[HTML]{FCE5CD}12.07 & 0.676 & \cellcolor[HTML]{FCE5CD}24.53 & 88.11 & \cellcolor[HTML]{FFF2CC}11.47 & 0.719 \\
SC-GS~\cite{huang2023sc} & 26.00 & - & \cellcolor[HTML]{F4CCCC}11.40 & \cellcolor[HTML]{FCE5CD}0.924 & \cellcolor[HTML]{FCE5CD}19.60 & - & 15.40 & 0.919 & 25.10 & - & \cellcolor[HTML]{F4CCCC}11.70 & \cellcolor[HTML]{FCE5CD}0.946 & \cellcolor[HTML]{FFF2CC}24.50 & - & 11.50 & \cellcolor[HTML]{FCE5CD}0.955 \\
4DGS*~\cite{wu20234d} & 25.60 & 86.27 & 13.36 & 0.780 & 18.90 & 76.97 & 15.43 & 0.775 & 24.60 & 82.66 & 14.05 & 0.758 & 24.42 & 88.23 & 11.55 & 0.787 \\
4DGS~\cite{wu20234d} & 25.69 & 86.43 & 13.60 & 0.845 & 18.88 & 75.92 & 17.59 & 0.723 & 24.46 & 82.30 & 13.87 & 0.732 & 24.44 & 88.04 & 12.02 & 0.739 \\ \hline
SF-Triaxes & \cellcolor[HTML]{F4CCCC}26.89 & \cellcolor[HTML]{FFF2CC}88.46 & 13.40 & 0.865 & 19.45 & \cellcolor[HTML]{FCE5CD}79.29 & 16.04 & \cellcolor[HTML]{FFF2CC}0.936 & \cellcolor[HTML]{F4CCCC}26.05 & \cellcolor[HTML]{F4CCCC}86.06 & 12.58 & 0.920 & 24.24 & \cellcolor[HTML]{FFF2CC}88.53 & 11.90 & 0.925 \\
SF-Triplanes & \cellcolor[HTML]{FFF2CC}26.34 & \cellcolor[HTML]{FCE5CD}88.75 & 12.43 & \cellcolor[HTML]{FFF2CC}0.891 & \cellcolor[HTML]{FCE5CD}19.61 & \cellcolor[HTML]{F4CCCC}79.85 & \cellcolor[HTML]{FFF2CC}14.67 & \cellcolor[HTML]{FCE5CD}0.937 & \cellcolor[HTML]{FCE5CD}25.68 & \cellcolor[HTML]{FCE5CD}85.27 & 13.01 & \cellcolor[HTML]{FFF2CC}0.928 & 24.40 & \cellcolor[HTML]{FCE5CD}88.60 & \cellcolor[HTML]{FCE5CD}11.33 & \cellcolor[HTML]{FFF2CC}0.932 \\
SF-PE-ResFields & \cellcolor[HTML]{FCE5CD}26.82 & \cellcolor[HTML]{F4CCCC}88.90 & \cellcolor[HTML]{FCE5CD}12.01 & \cellcolor[HTML]{F4CCCC}0.932 & \cellcolor[HTML]{F4CCCC}19.70 & \cellcolor[HTML]{FFF2CC}79.18 & \cellcolor[HTML]{FCE5CD}13.67 & \cellcolor[HTML]{F4CCCC}0.959 & \cellcolor[HTML]{FFF2CC}25.55 & \cellcolor[HTML]{FFF2CC}84.50 & \cellcolor[HTML]{FFF2CC}12.41 & \cellcolor[HTML]{F4CCCC}0.972 & \cellcolor[HTML]{F4CCCC}24.85 & \cellcolor[HTML]{F4CCCC}88.92 & \cellcolor[HTML]{F4CCCC}11.21 & \cellcolor[HTML]{F4CCCC}0.962 \\ \hline
 & \multicolumn{4}{c|}{Plate} & \multicolumn{4}{c|}{Press} & \multicolumn{4}{c|}{Sieve} & \multicolumn{4}{c}{Average} \\
\multirow{-2}{*}{Method} & PSNR\(\uparrow\) & SSIM\(\uparrow\)  & LPIPS\(\downarrow\) & M.'s \textit{I}\(\uparrow\) & PSNR\(\uparrow\) & SSIM\(\uparrow\)  & LPIPS\(\downarrow\) & M.'s \textit{I}\(\uparrow\) & PSNR\(\uparrow\) & SSIM\(\uparrow\)  & LPIPS\(\downarrow\) & M.'s \textit{I}\(\uparrow\) & PSNR\(\uparrow\) & SSIM\(\uparrow\)  & LPIPS\(\downarrow\) & M.'s \textit{I}\(\uparrow\) \\ \hline
Def.3D~\cite{yang2023deformable} & 20.33 & 80.26 & 19.12 & 0.702 & 25.40 & 85.61 & 13.66 & 0.720 & 25.24 & 86.72 & \cellcolor[HTML]{F4CCCC}10.85 & 0.703 & 23.75 & 84.32 & \cellcolor[HTML]{FCE5CD}13.28 & 0.706 \\
SC-GS~\cite{huang2023sc} & 20.20 & - & 20.20 & 0.938 & \cellcolor[HTML]{FCE5CD}26.60 & - & \cellcolor[HTML]{FFF2CC}13.50 & \cellcolor[HTML]{FCE5CD}0.945 & \cellcolor[HTML]{FFF2CC}26.00 & - & 11.40 & \cellcolor[HTML]{FFF2CC}0.927 & 24.10 & - & 14.00 & \cellcolor[HTML]{FCE5CD}0.939 \\
4DGS*~\cite{wu20234d} & 20.02 & 79.81 & 20.19 & 0.869 & 25.61 & 85.03 & 14.41 & 0.744 & 25.76 & 87.47 & 11.51 & 0.801 & 23.56 & 83.78 & 14.36 & 0.788 \\
4DGS~\cite{wu20234d} & 19.32 & 78.37 & 21.29 & 0.846 & 25.04 & 84.35 & 15.82 & 0.682 & 24.94 & 87.01 & 11.83 & 0.765 & 23.25 & 83.20 & 15.15 & 0.762 \\ \hline
SF-Triaxes & \cellcolor[HTML]{FFF2CC}20.92 & \cellcolor[HTML]{FFF2CC}81.74 & \cellcolor[HTML]{FFF2CC}18.63 & \cellcolor[HTML]{FCE5CD}0.986 & 26.22 & \cellcolor[HTML]{FFF2CC}88.21 & 14.09 & 0.908 & \cellcolor[HTML]{F4CCCC}26.71 & \cellcolor[HTML]{FFF2CC}87.68 & 11.47 & \cellcolor[HTML]{FCE5CD}0.943 & \cellcolor[HTML]{FCE5CD}24.36 & \cellcolor[HTML]{FFF2CC}85.71 & 14.01 & 0.926 \\
SF-Triplanes & \cellcolor[HTML]{F4CCCC}21.17 & \cellcolor[HTML]{F4CCCC}82.57 & \cellcolor[HTML]{FCE5CD}18.14 & \cellcolor[HTML]{FFF2CC}0.982 & \cellcolor[HTML]{FFF2CC}26.49 & \cellcolor[HTML]{FCE5CD}88.29 & \cellcolor[HTML]{FCE5CD}13.10 & \cellcolor[HTML]{FFF2CC}0.917 & \cellcolor[HTML]{FCE5CD}26.57 & \cellcolor[HTML]{FCE5CD}87.78 & \cellcolor[HTML]{FFF2CC}11.24 & \cellcolor[HTML]{FFF2CC}0.927 & \cellcolor[HTML]{FFF2CC}24.32 & \cellcolor[HTML]{F4CCCC}85.87 & \cellcolor[HTML]{FFF2CC}13.42 & \cellcolor[HTML]{FFF2CC}0.931 \\
SF-PE-ResFields & \cellcolor[HTML]{FCE5CD}21.02 & \cellcolor[HTML]{FCE5CD}82.51 & \cellcolor[HTML]{F4CCCC}17.41 & \cellcolor[HTML]{F4CCCC}0.993 & \cellcolor[HTML]{F4CCCC}27.09 & \cellcolor[HTML]{F4CCCC}88.33 & \cellcolor[HTML]{F4CCCC}12.29 & \cellcolor[HTML]{F4CCCC}0.967 & 25.80 & \cellcolor[HTML]{F4CCCC}88.32 & \cellcolor[HTML]{FCE5CD}10.96 & \cellcolor[HTML]{F4CCCC}0.965 & \cellcolor[HTML]{F4CCCC}24.40 & \cellcolor[HTML]{FCE5CD}85.81 & \cellcolor[HTML]{F4CCCC}12.85 & \cellcolor[HTML]{F4CCCC}0.964 \\ \hline
\end{tabular}
\label{tab:nerfds_quantitative}
\end{table*}

\begin{figure*}
\centering
\small
\setlength{\tabcolsep}{0.5pt}
\renewcommand{\arraystretch}{0.2}
\begin{tabular}{cccccc}
    4DGS & 4DGS* &  & Def.3D & SC-GS &  \\
    \cite{wu20234d} & \cite{wu20234d} & SF-Triplanes & \cite{yang2023deformable} & \cite{huang2023sc} & SF-PE-ResFields \\
    \includegraphics[width=0.164\linewidth]{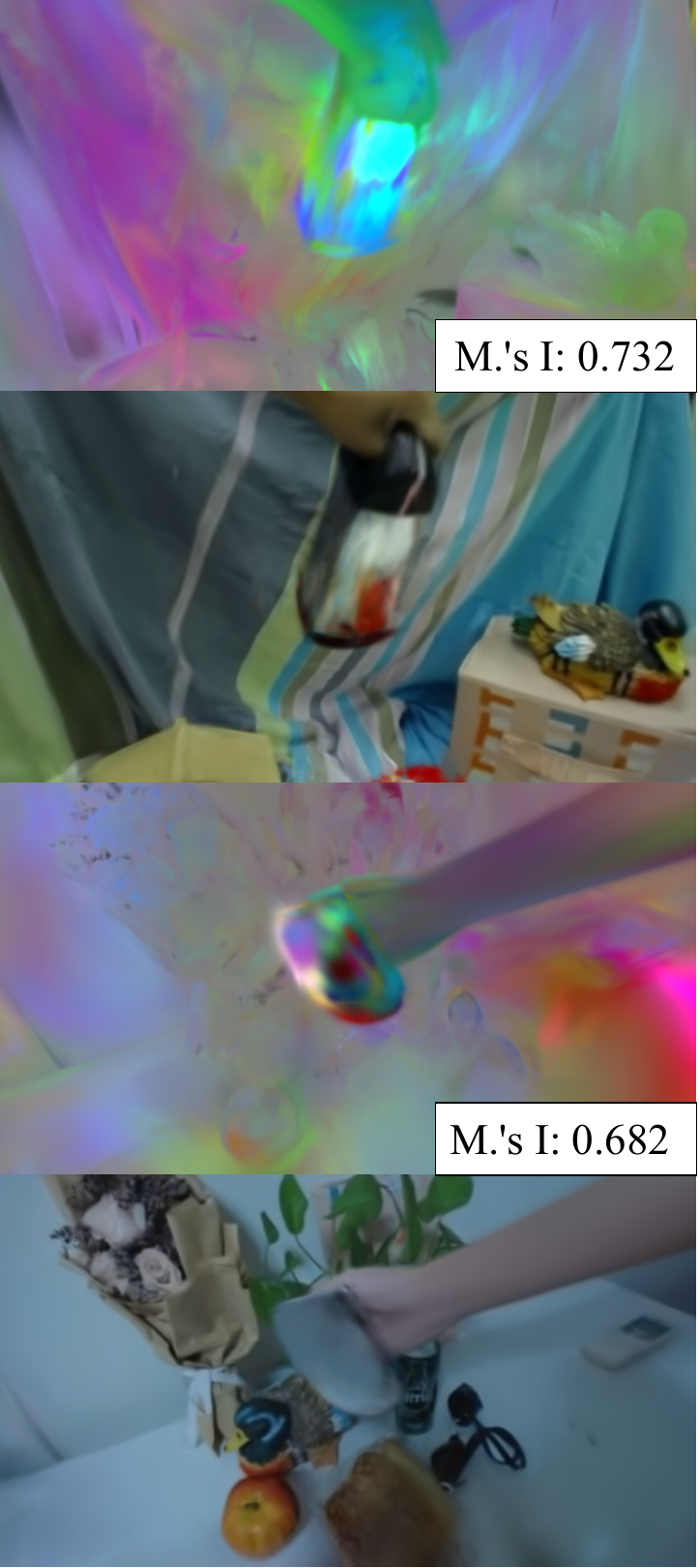} & \includegraphics[width=0.164\linewidth]{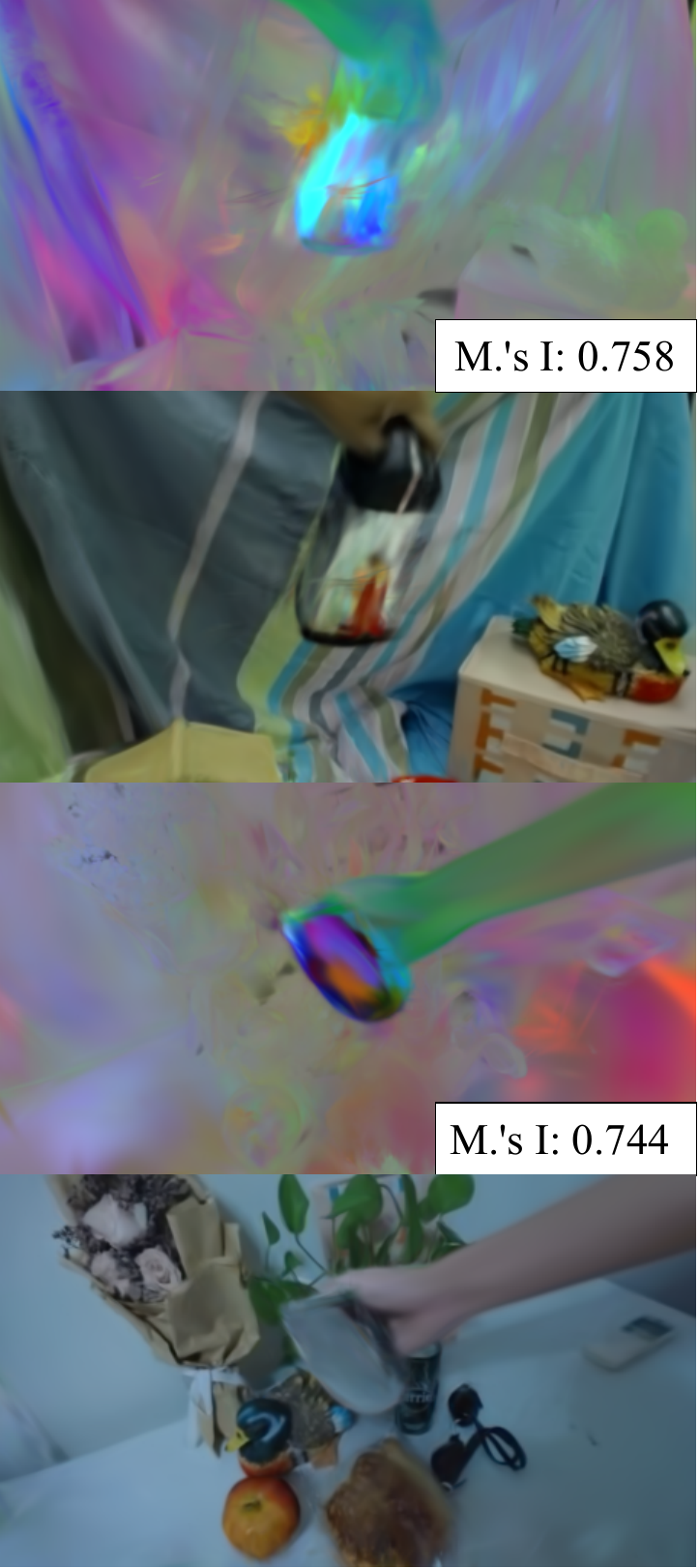} & \includegraphics[width=0.164\linewidth]{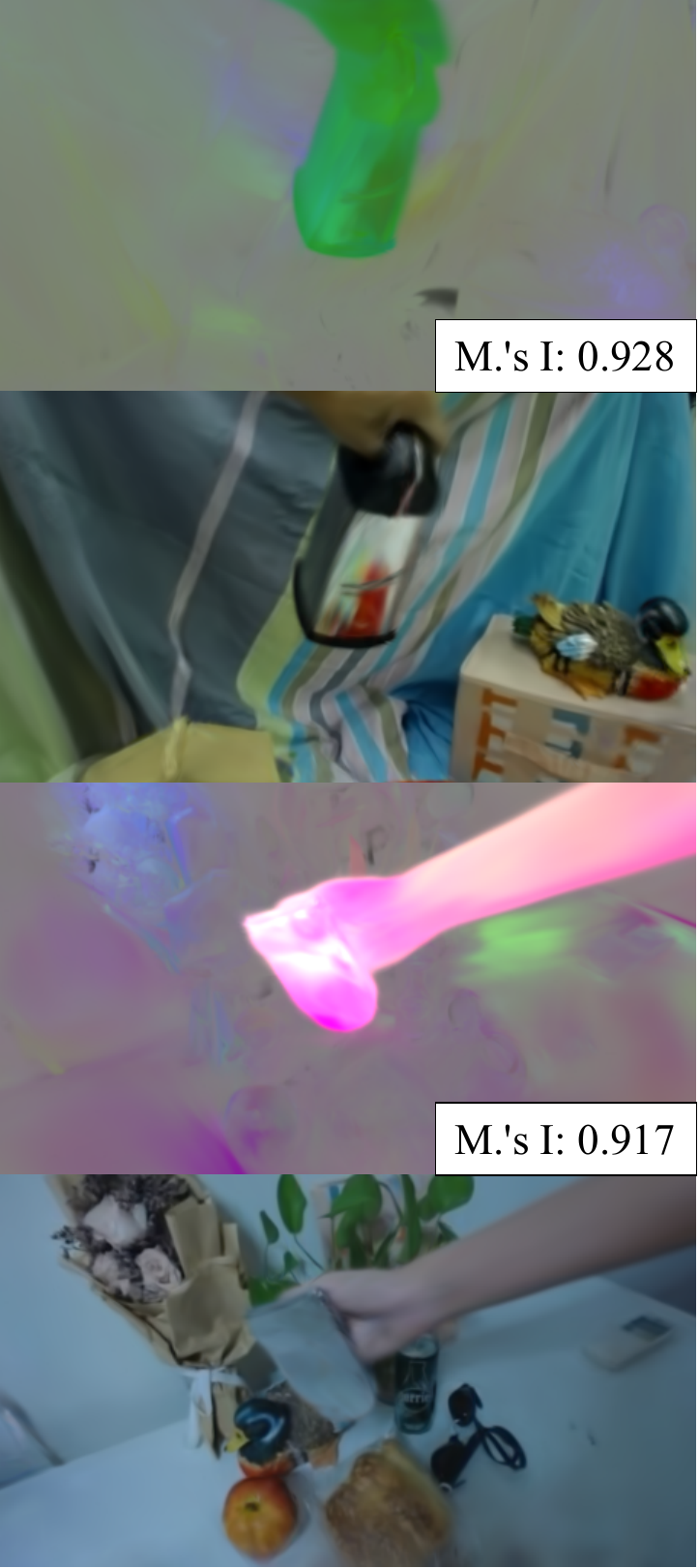} & \includegraphics[width=0.164\linewidth]{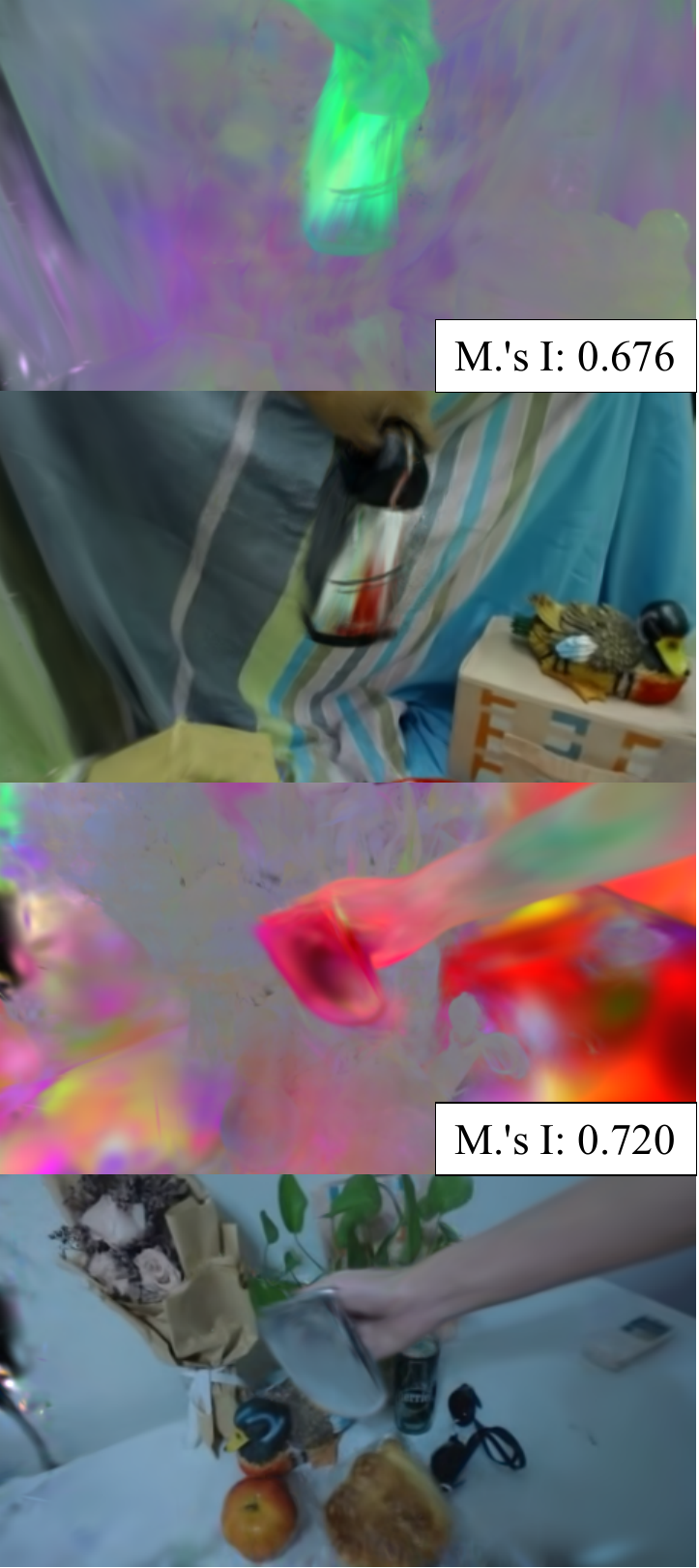} & 
    \includegraphics[width=0.164\linewidth]{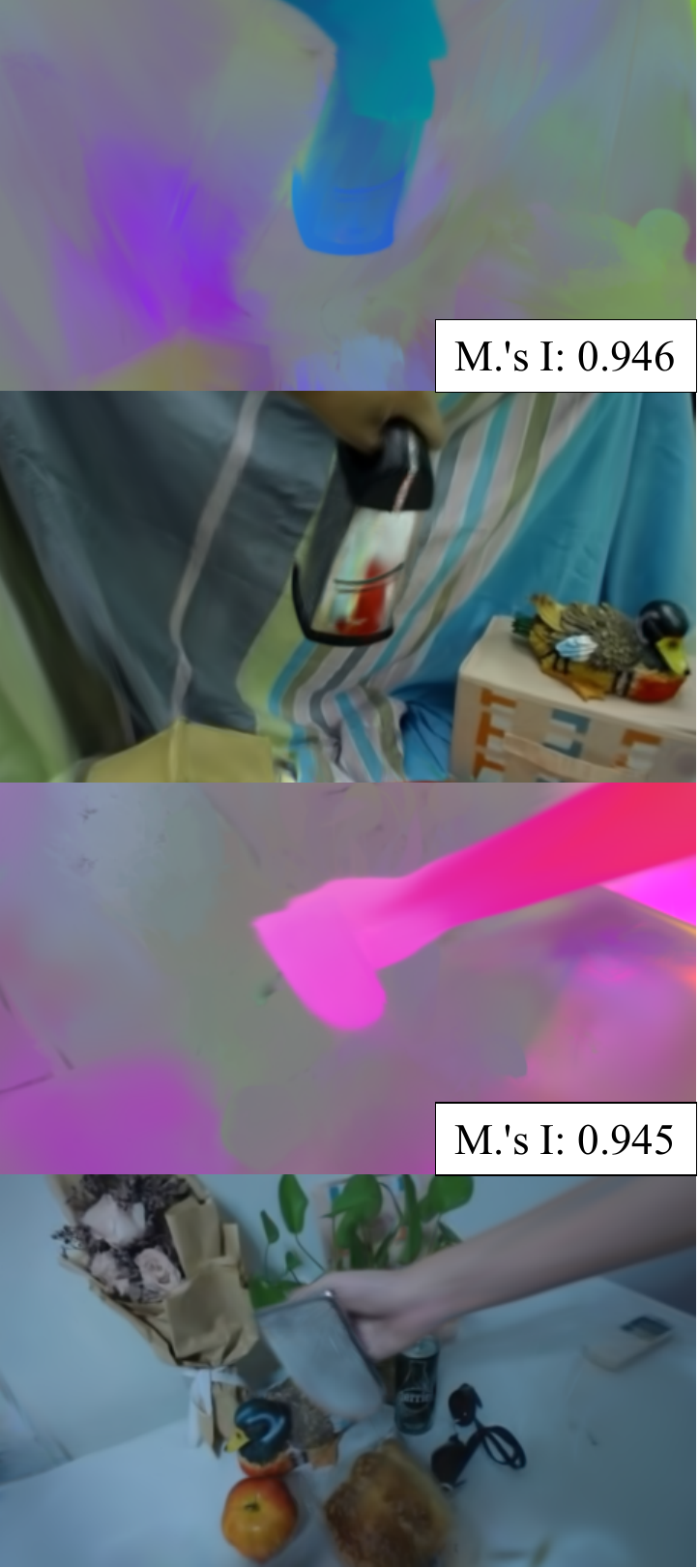} & \includegraphics[width=0.164\linewidth]{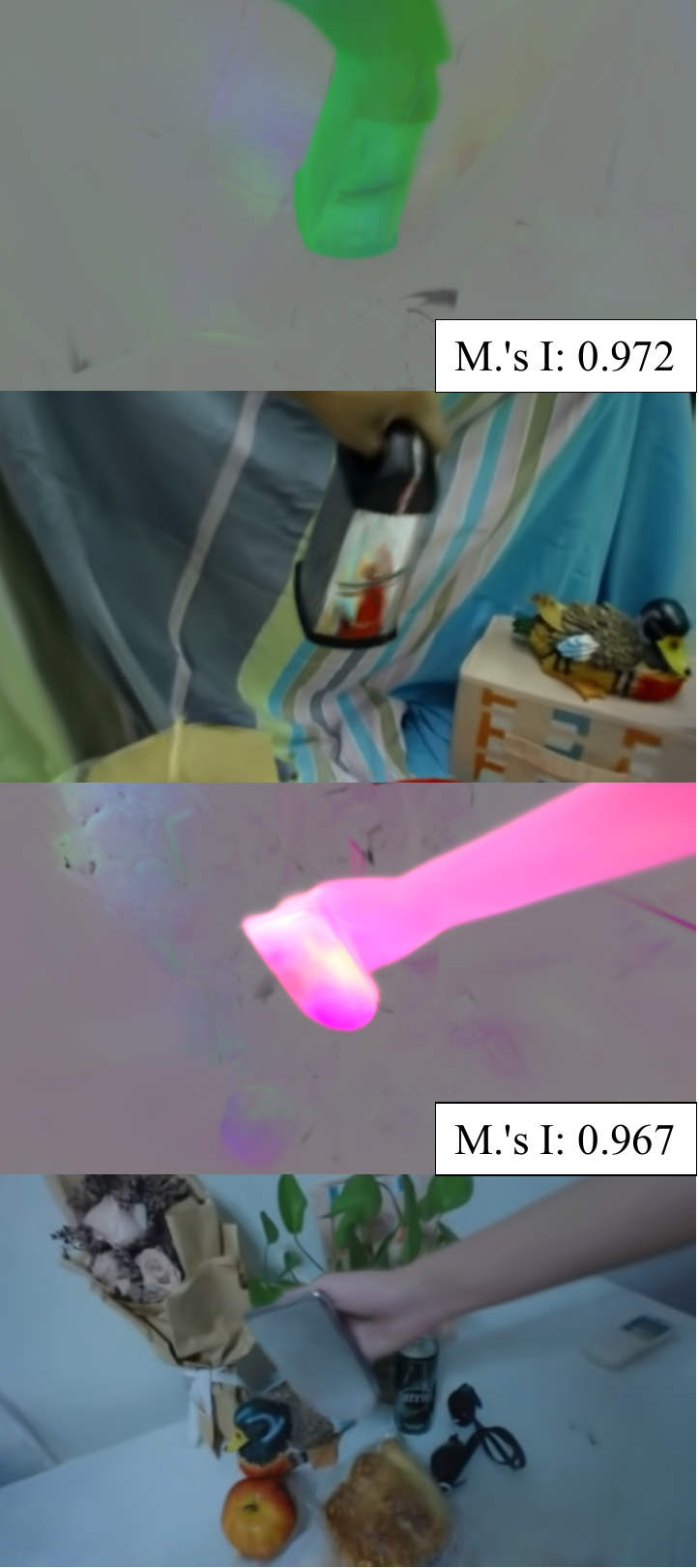} \\
\end{tabular}
\caption{Qualitative comparison of scene flows on NeRF-DS~\cite{yan2023nerf} dataset. With the derived regularizations in Sec.~\ref{sec:vel_acc_reg}, our representation effectively alleviates the high frequency caused by imprecise camera positions. We normalize all frames' motion vectors or velocities, so the gray color indicates static regions. We encourage readers to focus on the smoothness of motions.}
\label{fig:nerfds_scene_flow}
\end{figure*}

\begin{table*}
\centering
\small
\caption{Quantitative results on Hyper-NeRF~\cite{park2021hypernerf} dataset. Scenes in which most canonical-deformation-field-based methods fail are excluded. Results with abnormal floaters or diverged ones are excluded from the comparison (denoted as '-'). Jittery caused by inaccurate cameras is alleviated by applying \(\mathcal{L}_{acc}\) in our method, resulting in higher fidelity performance (PSNR) albeit reduced sharpness in some scenes.}
\begin{tabular}{lccc|ccc|ccc|ccc}
\hline
 & \multicolumn{3}{c|}{Chicken (vrig)} & \multicolumn{3}{c|}{3D Printer (vrig)} & \multicolumn{3}{c|}{Peal Banana (vrig)} & \multicolumn{3}{c}{Chickchicken (interp)} \\
\multirow{-2}{*}{Method} & PSNR\(\uparrow\) & LPIPS\(\downarrow\) & M.'s \textit{I}\(\uparrow\) & PSNR\(\uparrow\) & LPIPS\(\downarrow\) & M.'s \textit{I}\(\uparrow\) & PSNR\(\uparrow\) & LPIPS\(\downarrow\) & M.'s \textit{I}\(\uparrow\) & PSNR\(\uparrow\) & LPIPS\(\downarrow\) & M.'s \textit{I}\(\uparrow\) \\ \hline
4DGS~\cite{wu20234d} & 28.381 & 17.147 & 0.758 & 22.032 & 23.669 & 0.648 & 28.286 & 15.860 & 0.804 & \cellcolor[HTML]{FCE5CD}26.680 & 20.228 & 0.703 \\
Grid4D~\cite{xu2024grid4d} & \cellcolor[HTML]{F4CCCC}29.197 & 18.834 & 0.890 & \cellcolor[HTML]{F4CCCC}22.310 & 20.525 & 0.918 & \cellcolor[HTML]{F4CCCC}28.512 & \cellcolor[HTML]{FCE5CD}14.841 & 0.898 & \cellcolor[HTML]{F4CCCC}27.428 & 20.961 & 0.901 \\
Def.3D~\cite{yang2023deformable} & 21.816 & 18.742 & 0.943 & 18.814 & 27.319 & 0.897 & - & - & - & 25.040 & 19.507 & 0.890 \\ \hline
SF-Triplanes w/o acc & 24.393 & 18.604 & 0.898 & 21.786 & \cellcolor[HTML]{FCE5CD}19.802 & 0.902 & \cellcolor[HTML]{FCE5CD}28.334 & \cellcolor[HTML]{F4CCCC}14.363 & 0.947 & 26.037 & 17.878 & 0.954 \\
SF-Triplanes w/ acc & \cellcolor[HTML]{FCE5CD}28.829 & \cellcolor[HTML]{FCE5CD}16.889 & 0.877 & \cellcolor[HTML]{FCE5CD}22.169 & 22.448 & 0.866 & 28.083 & 17.031 & 0.894 & 26.154 & 19.280 & 0.945 \\
SF-PE-ResFields w/o acc & 23.204 & 18.465 & \cellcolor[HTML]{F4CCCC}0.960 & 20.219 & \cellcolor[HTML]{F4CCCC}18.696 & \cellcolor[HTML]{F4CCCC}0.975 & 26.164 & 17.289 & \cellcolor[HTML]{F4CCCC}0.975 & 26.278 & \cellcolor[HTML]{F4CCCC}15.788 & \cellcolor[HTML]{F4CCCC}0.985 \\
SF-PE-ResFields w/ acc & 28.751 & \cellcolor[HTML]{F4CCCC}16.305 & \cellcolor[HTML]{FCE5CD}0.949 & 21.934 & 24.114 & \cellcolor[HTML]{FCE5CD}0.946 & 27.579 & 17.175 & \cellcolor[HTML]{FCE5CD}0.960 & 26.543 & \cellcolor[HTML]{FCE5CD}15.788 & \cellcolor[HTML]{FCE5CD}0.983 \\ \hline
 & \multicolumn{3}{c|}{Cut Lemon (interp)} & \multicolumn{3}{c|}{Hand (interp)} & \multicolumn{3}{c|}{Slice Banana (interp)} & \multicolumn{3}{c}{Torch Chocolate (interp)} \\
\multirow{-2}{*}{Method} & PSNR\(\uparrow\) & LPIPS\(\downarrow\) & M.'s \textit{I}\(\uparrow\) & PSNR\(\uparrow\) & LPIPS\(\downarrow\) & M.'s \textit{I}\(\uparrow\) & PSNR\(\uparrow\) & LPIPS\(\downarrow\) & M.'s \textit{I}\(\uparrow\) & PSNR\(\uparrow\) & LPIPS\(\downarrow\) & M.'s \textit{I}\(\uparrow\) \\ \hline
4DGS~\cite{wu20234d} & 31.427 & 12.182 & 0.696 & 28.975 & 20.379 & 0.471 & 25.567 & 26.532 & 0.715 & 27.437 & 16.077 & 0.754 \\
Grid4D\cite{xu2024grid4d} & \cellcolor[HTML]{F4CCCC}32.101 & \cellcolor[HTML]{FCE5CD}11.862 & 0.746 & \cellcolor[HTML]{F4CCCC}31.338 & \cellcolor[HTML]{F4CCCC}12.656 & 0.694 & \cellcolor[HTML]{FCE5CD}25.869 & 25.887 & 0.801 & \cellcolor[HTML]{FCE5CD}28.256 & \cellcolor[HTML]{F4CCCC}10.842 & 0.779 \\
Def.3D~\cite{yang2023deformable} & 31.047 & 12.480 & 0.866 & 28.824 & 19.685 & 0.838 & - & - & - & 27.329 & 13.120 & 0.921 \\ \hline
SF-Triplanes w/o acc & 30.674 & 12.984 & 0.837 & 30.135 & 16.175 & 0.904 & 25.313 & \cellcolor[HTML]{FCE5CD}24.683 & 0.943 & 28.077 & 12.917 & 0.971 \\
SF-Triplanes w/ acc & 30.775 & 13.855 & 0.824 & 30.137 & 15.490 & 0.898 & 25.180 & 25.867 & 0.937 & 28.129 & 13.176 & 0.965 \\
SF-PE-ResFields w/o acc & 31.457 & 12.148 & \cellcolor[HTML]{FCE5CD}0.938 & \cellcolor[HTML]{FCE5CD}30.432 & \cellcolor[HTML]{FCE5CD}14.705 & \cellcolor[HTML]{F4CCCC}0.967 & 25.505 & \cellcolor[HTML]{F4CCCC}22.281 & \cellcolor[HTML]{F4CCCC}0.975 & 28.185 & \cellcolor[HTML]{FCE5CD}12.390 & \cellcolor[HTML]{F4CCCC}0.987 \\
SF-PE-ResFields w/ acc & \cellcolor[HTML]{FCE5CD}31.509 & \cellcolor[HTML]{F4CCCC}11.770 & \cellcolor[HTML]{F4CCCC}0.942 & 30.356 & 14.810 & \cellcolor[HTML]{FCE5CD}0.964 & \cellcolor[HTML]{F4CCCC}25.874 & 25.191 & \cellcolor[HTML]{FCE5CD}0.968 & \cellcolor[HTML]{F4CCCC}28.310 & 12.421 & \cellcolor[HTML]{FCE5CD}0.984 \\ \hline
\end{tabular}
\label{tab:hyper_comp}
\end{table*}

\begin{figure*}
\centering
\tiny
\setlength{\tabcolsep}{0.0pt}
\renewcommand{\arraystretch}{0.2}
\begin{tabular}{ccccccccccc}
    4DGS* & Def.3D & Grid4D & SF-Triplanes & SF-PE-ResFields & & 4DGS* & Grid4D & SF-PE-ResFields & SF-PE-ResFields\(\mathcal{x}\) \\
    \cite{wu20234d} & \cite{yang2023deformable} & \cite{xu2024grid4d} & w/ \(\mathcal{L}_{acc}\) & w/ \(\mathcal{L}_{acc}\) & Ground Truth & \cite{wu20234d} &  \cite{xu2024grid4d} & w/o \(\mathcal{L}_{acc}\) & w/ \(\mathcal{L}_{acc}\) & Ground Truth \\
    \includegraphics[width=0.09\linewidth]{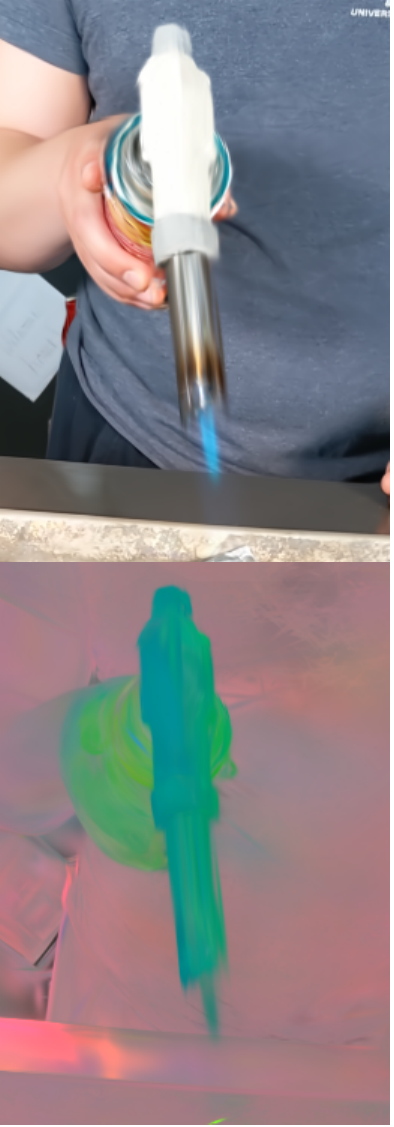} & \includegraphics[width=0.09\linewidth]{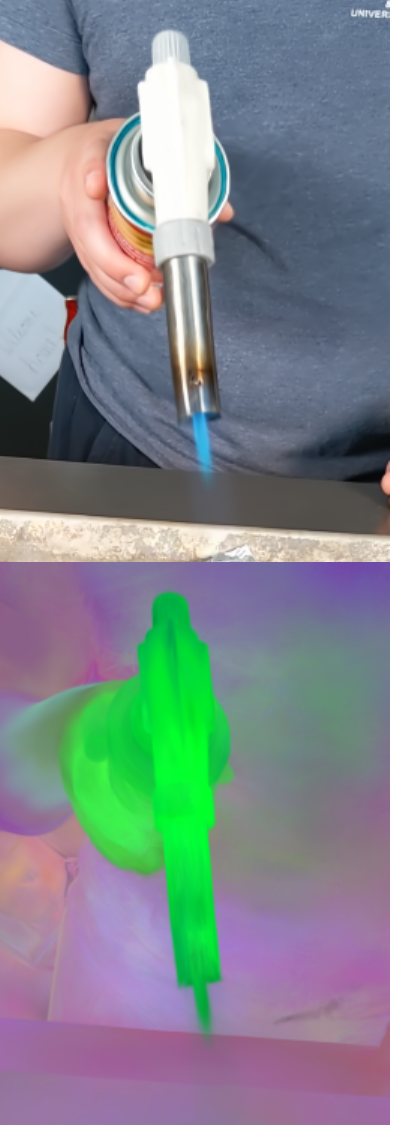} & \includegraphics[width=0.09\linewidth]{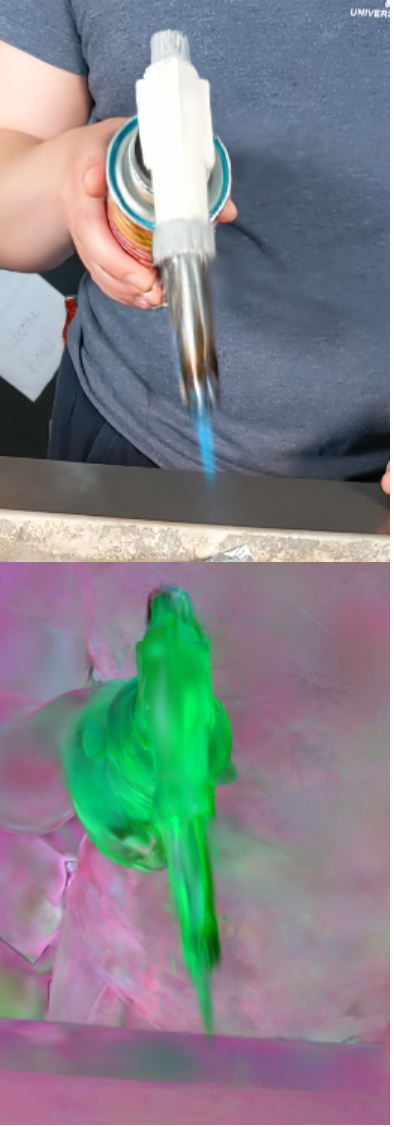} & \includegraphics[width=0.09\linewidth]{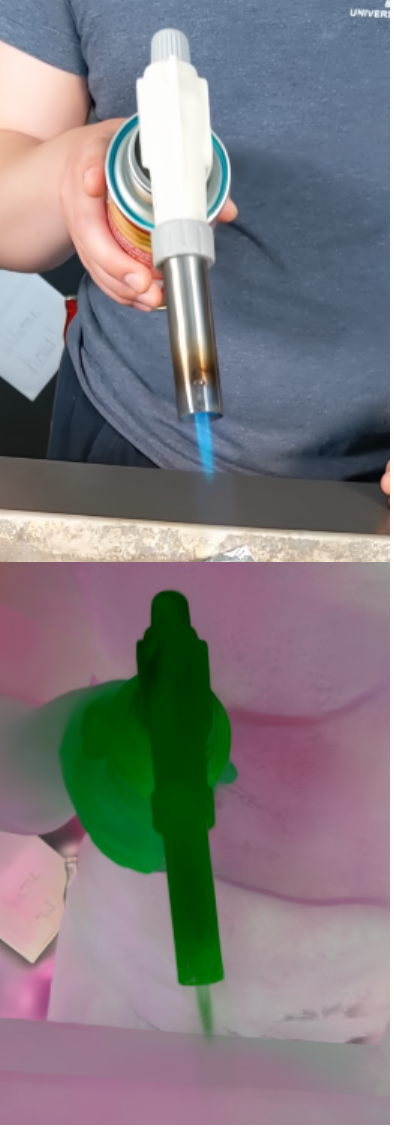} & 
    \includegraphics[width=0.09\linewidth]{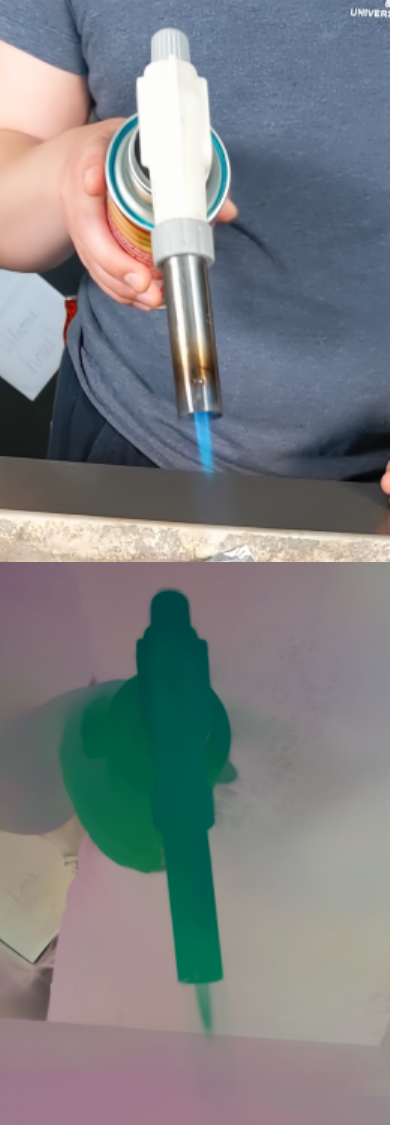} & \includegraphics[width=0.09\linewidth]{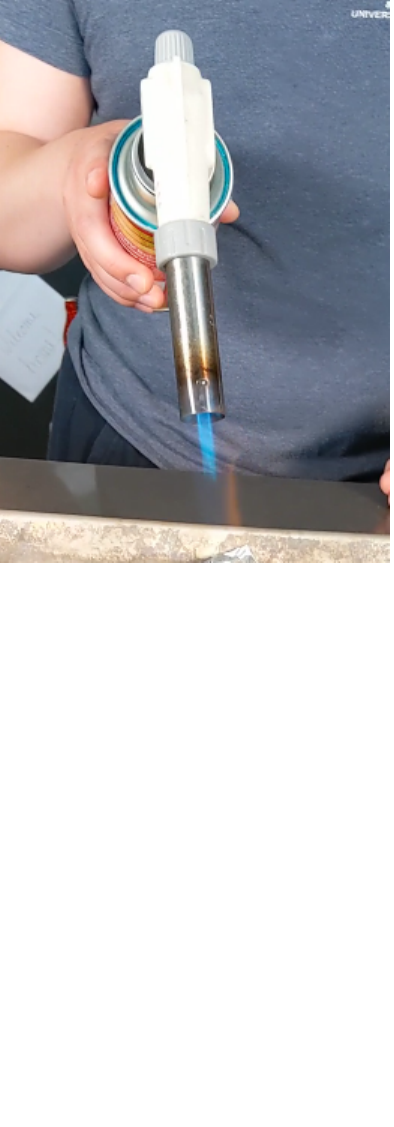} &
    \includegraphics[width=0.09\linewidth]{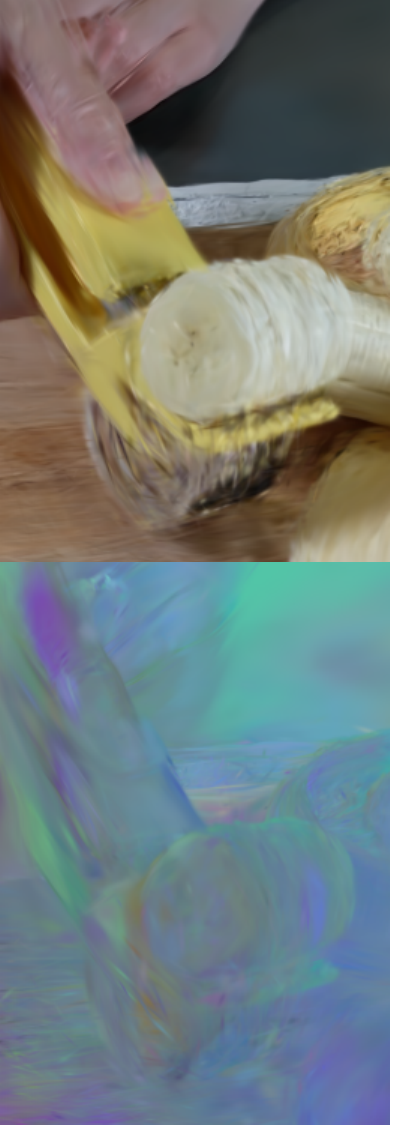} &
    \includegraphics[width=0.09\linewidth]{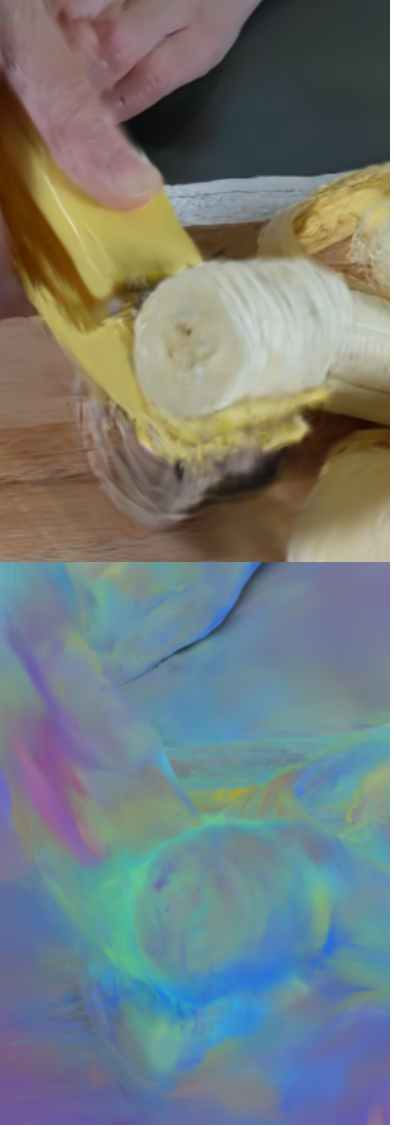} &
    \includegraphics[width=0.09\linewidth]{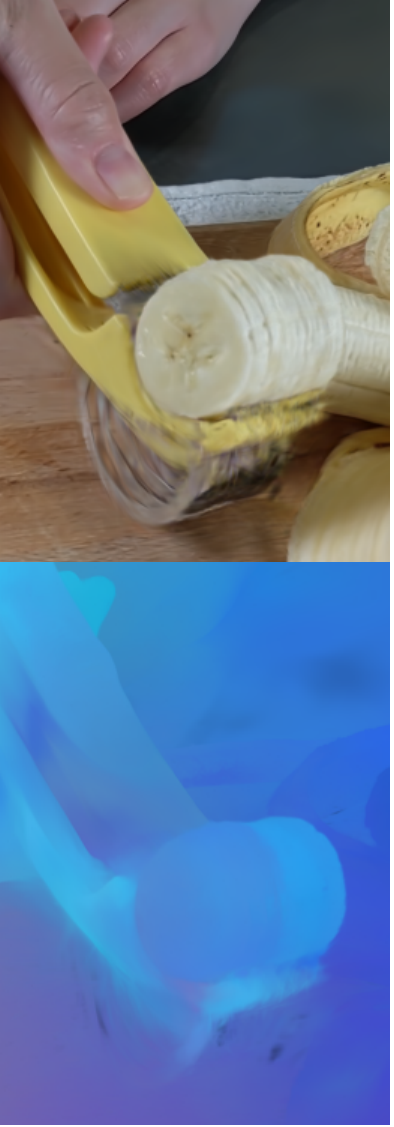} &
    \includegraphics[width=0.09\linewidth]{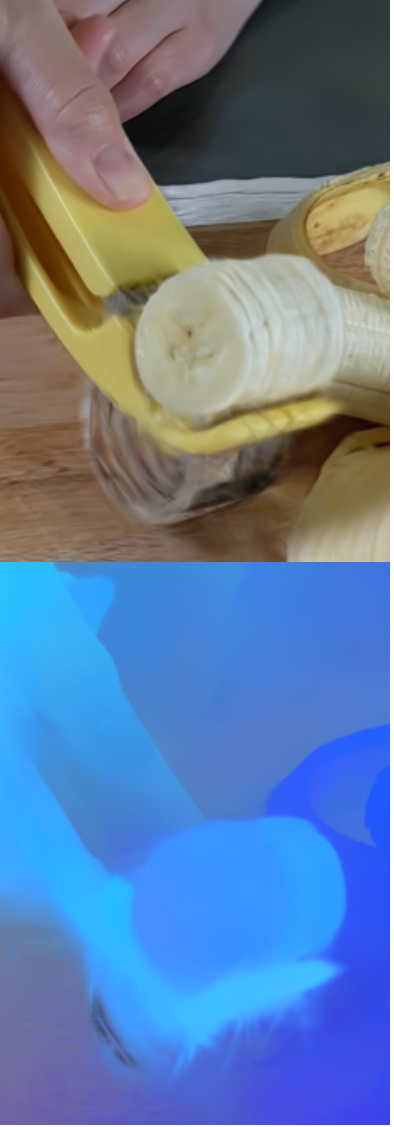} &
    \includegraphics[width=0.09\linewidth]{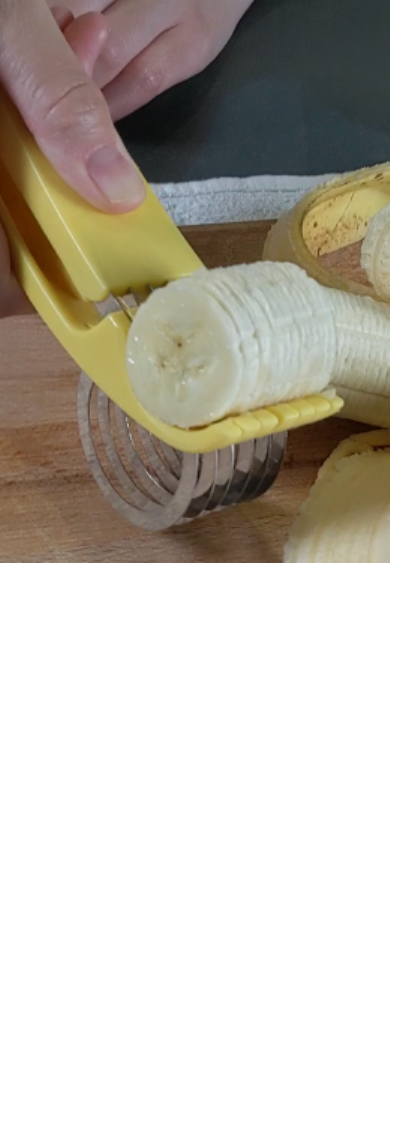}
\end{tabular}
\caption{Qualitative comparison of rendered images and scene flows on Hyper-NeRF~\cite{park2021hypernerf} dataset. Since the whole scene is jittery due to inaccurate cameras, static regions are also colored, compared with the NeRF-DS\cite{yan2023nerf} dataset. Additional results are included in Fig.~\ref{fig:extra_hyper_nerf_vis}.}
\label{fig:hyper_nerf_vis}
\end{figure*}

\subsubsection{Real-world datasets} NeRF-DS~\cite{yan2023nerf} features slow-moving objects with specular surfaces. In this dataset, only a single view is available per timestamp, and the camera positions are less accurate than those in synthetic datasets. To account for the motion characteristics in this dataset, we set \(N=30\) and \(rank=10\), showcasing the utility of explicit (\(N\)) regularization. As shown in Tab.~\ref{tab:nerfds_quantitative}, both grid-based and MLP-based methods benefit from our representation, achieving superior fidelity and perceptual quality than the baselines. Our method delivers competitive reconstruction quality compared to two-stage LBS-based SC-GS~\cite{huang2023sc}, with notably better temporal stability in static, textureless regions (white table, curtain in background). Additionally, we evaluate our method on another single-view-per-timestep dataset, Hyper-NeRF~\cite{park2021hypernerf}. It contains higher-resolution images but suffers from less accurate camera positions. We follow Eq.~\ref{eq:determined_eq} to determine \(N\) and set \(rank=16\). As shown in Tab.~\ref{tab:hyper_comp}, our method strikes a balance between fidelity and perceptual quality, with fewer abnormal floaters compared to Grid4D~\cite{xu2024grid4d}. As pointed out in Hyper-NeRF's original script, LPIPS (\textit{Alex}) aligns better with visual results, as illustrated in Fig.~\ref{fig:hyper_nerf_vis}. The imprecise camera poses in these two datasets lead to significant visual variations in sharpness across different views, a challenge most evident in grid-based methods without additional regularization. With strong explicit constraints, our variant SF-Triplanes reasonably preserve rigidity, as shown in the torch in Fig.~\ref{fig:hyper_nerf_vis}. As expected, MLP-based methods maintain better spatial coherence in scenes with simple motions due to their inherent global continuity but frequently fail in extreme cases.

\begin{table*}
\centering
\small
\setlength{\tabcolsep}{2.0pt}
\caption{Quantitative results on Neu3D~\cite{li2022neural}. 4DGS and our SF-Triplanes perform better than MLP-based methods in the long term with complex motion scenes. Grid4D~\cite{xu2024grid4d} deactivated or reduced weights for smooth regularization on this dataset in their official implementation.}
\begin{tabular}{lcccc|cccc|ccccllll}
\cline{1-13}
 & \multicolumn{4}{c|}{Coffee Martini} & \multicolumn{4}{c|}{Cook Spinach} & \multicolumn{4}{c}{Cut Roasted Beef} &  &  &  &  \\
\multirow{-2}{*}{Method} & PSNR\(\uparrow\) & SSIM\(\uparrow\) & LPIPS\(\downarrow\) & M.'s \textit{I}\(\uparrow\) & PSNR\(\uparrow\) & SSIM\(\uparrow\) & LPIPS\(\downarrow\) & M.'s \textit{I}\(\uparrow\) & PSNR\(\uparrow\) & SSIM\(\uparrow\) & LPIPS\(\downarrow\) & M.'s \textit{I}\(\uparrow\) &  &  &  &  \\ \cline{1-13}
4DGS~\cite{wu20234d} & 28.59 & 91.45 & 10.96 & 0.938 & \cellcolor[HTML]{FCE5CD}32.61 & 94.86 & \cellcolor[HTML]{FCE5CD}8.17 & 0.888 & 32.96 & 95.00 & 8.16 & 0.855 &  &  &  &  \\
Def.3D~\cite{yang2023deformable} & \cellcolor[HTML]{FCE5CD}28.81 & \cellcolor[HTML]{FCE5CD}91.64 & 11.07 & 0.915 & 32.06 & 94.83 & 8.46 & 0.842 & 32.84 & 94.97 & 8.43 & 0.809 &  &  &  &  \\
Grid4D~\cite{xu2024grid4d} & 28.30 & 90.02 & 13.59 & 0.418 & \cellcolor[HTML]{F4CCCC}32.64 & \cellcolor[HTML]{F4CCCC}94.96 & 8.62 & 0.700 & \cellcolor[HTML]{F4CCCC}33.22 & 94.47 & 8.75 & 0.669 &  &  &  &  \\ \cline{1-13} 
SF-Triplanes & \cellcolor[HTML]{F4CCCC}28.88 & \cellcolor[HTML]{F4CCCC}91.67 & \cellcolor[HTML]{F4CCCC}10.86 & \cellcolor[HTML]{FCE5CD}0.965 & 32.02 & 94.79 & \cellcolor[HTML]{F4CCCC}8.15 & \cellcolor[HTML]{FCE5CD}0.931 & \cellcolor[HTML]{FCE5CD}33.11 & \cellcolor[HTML]{F4CCCC}95.06 & \cellcolor[HTML]{FCE5CD}8.02 & \cellcolor[HTML]{FCE5CD}0.908 &  &  &  &  \\
SF-PE-Resfields & 28.74 & 91.48 & \cellcolor[HTML]{FCE5CD}10.86 & \cellcolor[HTML]{F4CCCC}0.967 & 32.47 & \cellcolor[HTML]{FCE5CD}94.89 & 8.54 & \cellcolor[HTML]{F4CCCC}0.939 & 32.57 & \cellcolor[HTML]{FCE5CD}95.04 & \cellcolor[HTML]{F4CCCC}7.99 & \cellcolor[HTML]{F4CCCC}0.913 &  &  &  &  \\ \hline
 & \multicolumn{4}{c|}{Flame Salmon} & \multicolumn{4}{c|}{Flame Steak} & \multicolumn{4}{c|}{Sear Steak} & \multicolumn{4}{c}{Average} \\
\multirow{-2}{*}{Method} & PSNR\(\uparrow\) & SSIM\(\uparrow\) & LPIPS\(\downarrow\) & M.'s \textit{I}\(\uparrow\) & PSNR\(\uparrow\) & SSIM\(\uparrow\) & LPIPS\(\downarrow\) & M.'s \textit{I}\(\uparrow\) & PSNR\(\uparrow\) & SSIM\(\uparrow\) & LPIPS\(\downarrow\) & \multicolumn{1}{c|}{M.'s \textit{I}\(\uparrow\)} & \multicolumn{1}{c}{PSNR\(\uparrow\)} & \multicolumn{1}{c}{SSIM\(\uparrow\)} & \multicolumn{1}{c}{LPIPS\(\downarrow\)} & \multicolumn{1}{c}{M.'s \textit{I}\(\uparrow\)} \\ \hline
4DGS~\cite{wu20234d} & \cellcolor[HTML]{FCE5CD}29.16 & \cellcolor[HTML]{FCE5CD}91.98 & 10.22 & 0.930 & \cellcolor[HTML]{FCE5CD}32.82 & 95.52 & \cellcolor[HTML]{FCE5CD}7.04 & 0.869 & \cellcolor[HTML]{F4CCCC}33.39 & 95.59 & \cellcolor[HTML]{FCE5CD}7.31 & \multicolumn{1}{c|}{\cellcolor[HTML]{F4CCCC}0.909} & \multicolumn{1}{c}{\cellcolor[HTML]{FCE5CD}31.59} & \multicolumn{1}{c}{94.04} & \multicolumn{1}{c}{8.64} & \multicolumn{1}{c}{0.898} \\
Def.3D~\cite{yang2023deformable} & 28.28 & 91.79 & 10.56 & 0.924 & 32.76 & 95.49 & 7.23 & 0.825 & 33.00 & 95.49 & 7.67 & \multicolumn{1}{c|}{0.760} & \multicolumn{1}{c}{31.29} & \multicolumn{1}{c}{94.04} & \multicolumn{1}{c}{8.90} & \multicolumn{1}{c}{0.846} \\
Grid4D~\cite{xu2024grid4d} & 29.12 & 90.90 & 12.91 & 0.452 & 32.56 & \cellcolor[HTML]{F4CCCC}95.71 & 7.31 & 0.676 & 33.16 & \cellcolor[HTML]{F4CCCC}95.68 & 7.47 & \multicolumn{1}{c|}{0.650} & \multicolumn{1}{c}{31.50} & \multicolumn{1}{c}{93.62} & \multicolumn{1}{c}{9.78} & \multicolumn{1}{c}{0.594} \\ \hline
SF-Triplanes & \cellcolor[HTML]{F4CCCC}29.21 & 91.90 & \cellcolor[HTML]{FCE5CD}10.13 & \cellcolor[HTML]{F4CCCC}0.961 & \cellcolor[HTML]{F4CCCC}33.21 & 95.65 & 7.05 & \cellcolor[HTML]{FCE5CD}0.914 & 33.15 & 95.43 & \cellcolor[HTML]{F4CCCC}7.31 & \multicolumn{1}{c|}{0.901} & \multicolumn{1}{c}{\cellcolor[HTML]{F4CCCC}31.60} & \multicolumn{1}{c}{\cellcolor[HTML]{FCE5CD}94.08} & \multicolumn{1}{c}{\cellcolor[HTML]{F4CCCC}8.59} & \multicolumn{1}{c}{\cellcolor[HTML]{FCE5CD}0.930} \\
SF-PE-Resfields & 28.97 & \cellcolor[HTML]{F4CCCC}91.99 & \cellcolor[HTML]{F4CCCC}10.00 & \cellcolor[HTML]{FCE5CD}0.960 & 32.82 & \cellcolor[HTML]{FCE5CD}95.65 & \cellcolor[HTML]{F4CCCC}6.95 & \cellcolor[HTML]{F4CCCC}0.934 & \cellcolor[HTML]{FCE5CD}33.23 & \cellcolor[HTML]{FCE5CD}95.61 & 7.45 & \multicolumn{1}{c|}{\cellcolor[HTML]{FCE5CD}0.908} & \multicolumn{1}{c}{31.47} & \multicolumn{1}{c}{\cellcolor[HTML]{F4CCCC}94.11} & \multicolumn{1}{c}{\cellcolor[HTML]{FCE5CD}8.63} & \multicolumn{1}{c}{\cellcolor[HTML]{F4CCCC}0.937} \\ \hline
\end{tabular}
\label{tab:neu3d_comp}
\end{table*}

\begin{figure*}
\centering
\tiny
\setlength{\tabcolsep}{0.0pt}
\renewcommand{\arraystretch}{0.2}
\begin{tabular}{ccccccccccc}
     & Def.3D & 4DGS & Grid4D &  & & Def.3D & 4DGS & Grid4D &  \\
     Ground Truth & \cite{yang2023deformable} & \cite{wu20234d} & \cite{xu2024grid4d} & SF-Triplanes & SF-PE-ResFields & \cite{yang2023deformable} & \cite{wu20234d} & \cite{xu2024grid4d} & SF-Triplanes & SF-PE-ResFields \\
     \includegraphics[width=0.09\linewidth]{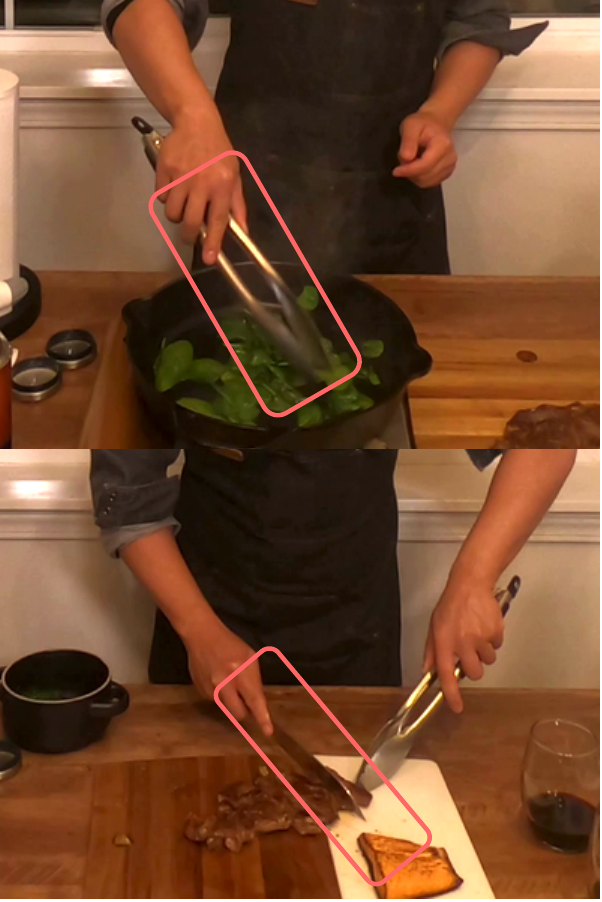} & 
     \includegraphics[width=0.09\linewidth]{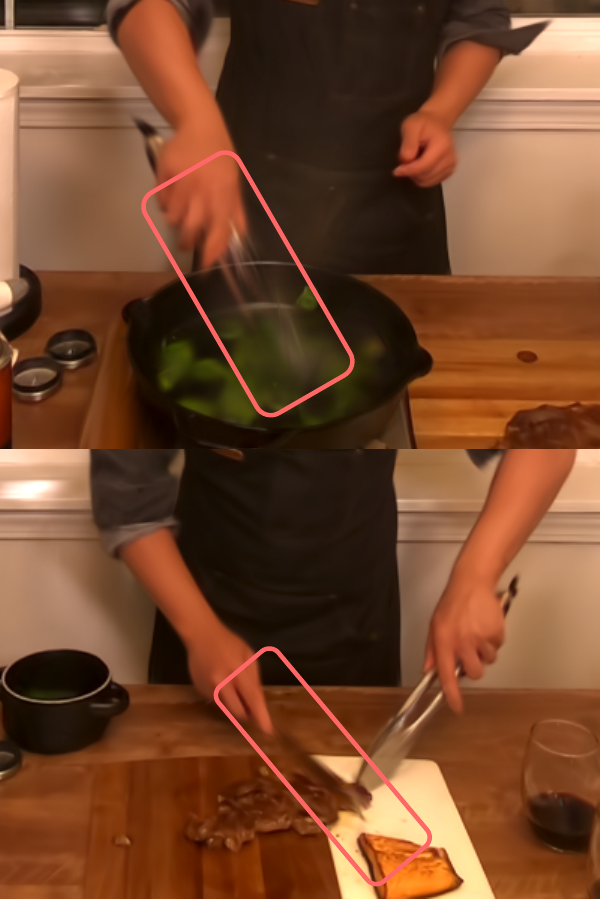} & 
     \includegraphics[width=0.09\linewidth]{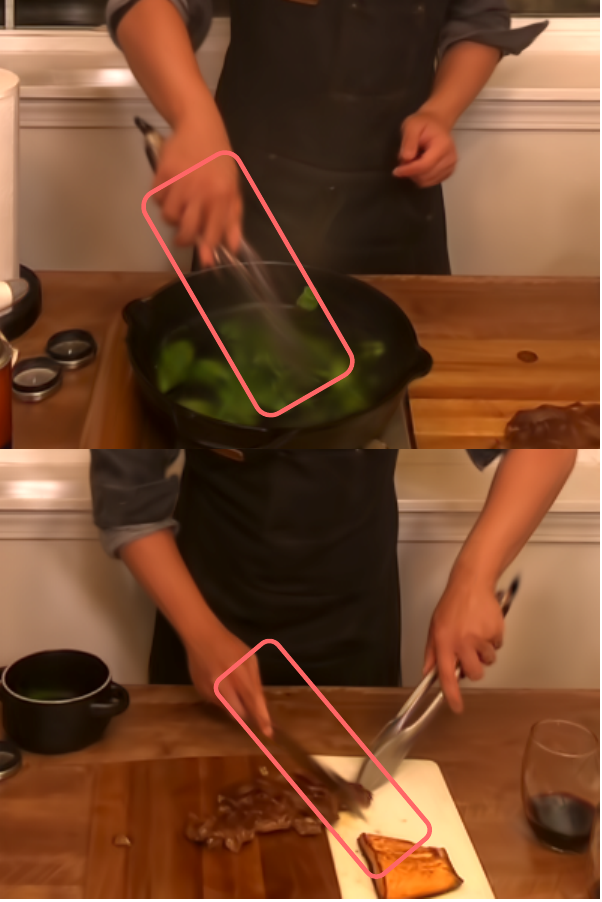} & 
     \includegraphics[width=0.09\linewidth]{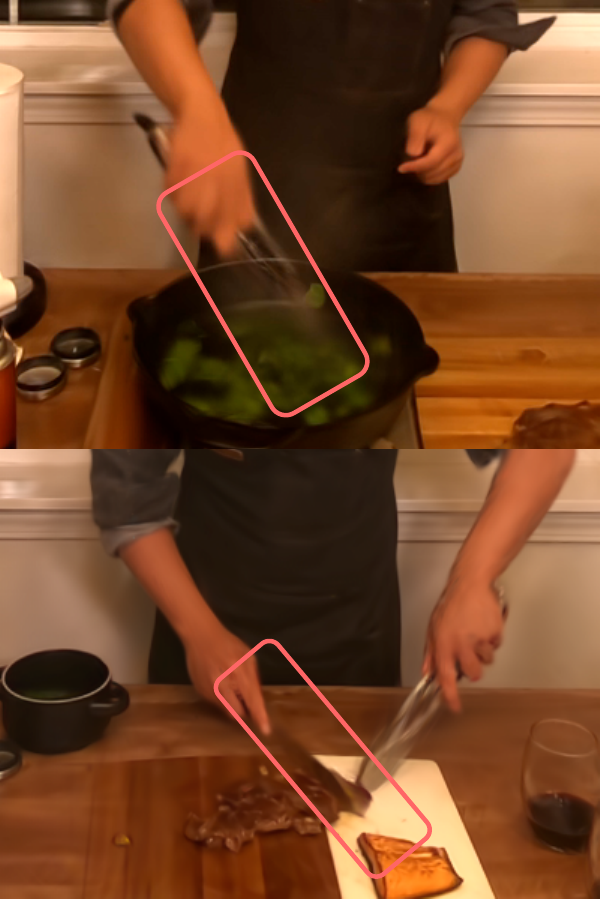} & 
     \includegraphics[width=0.09\linewidth]{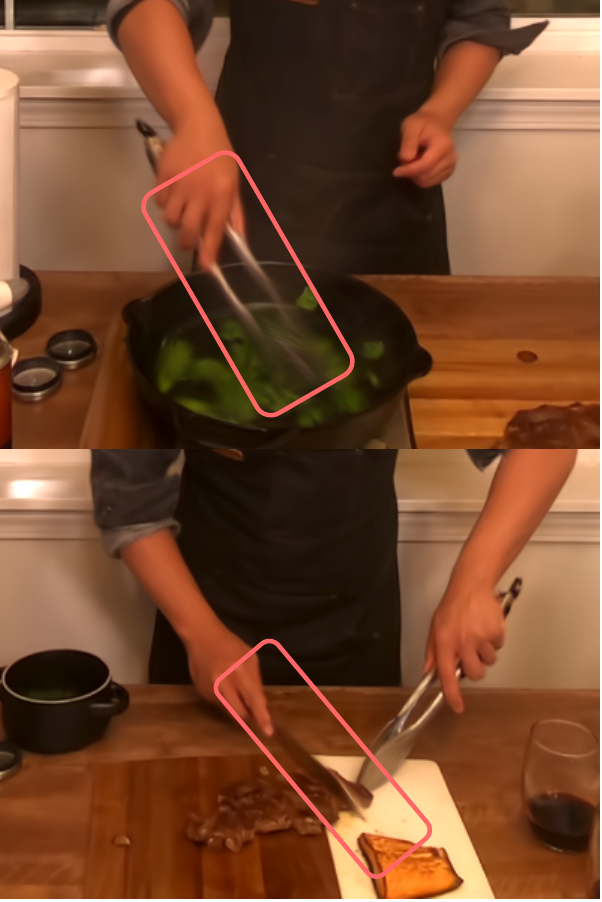} & 
     \includegraphics[width=0.09\linewidth]{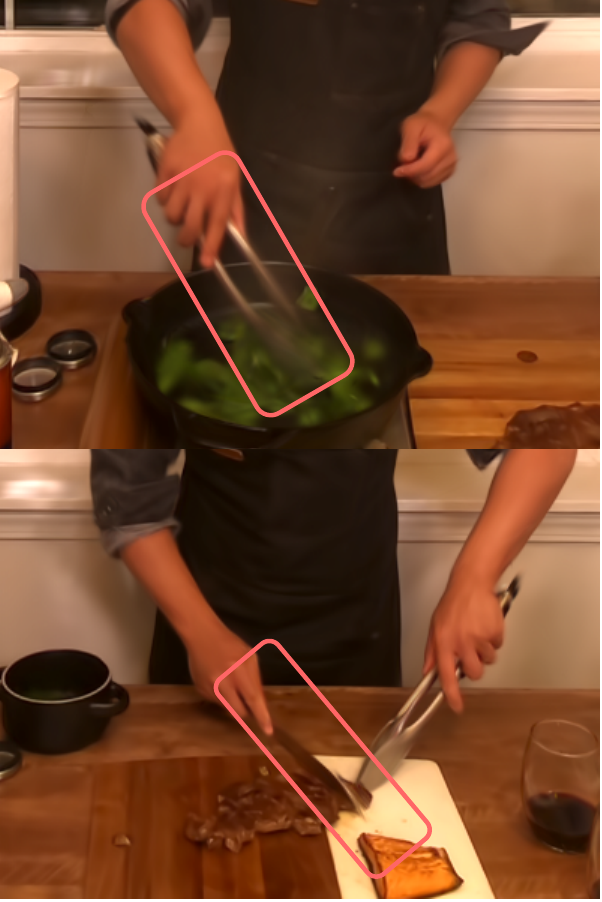} & 
     \includegraphics[width=0.09\linewidth]{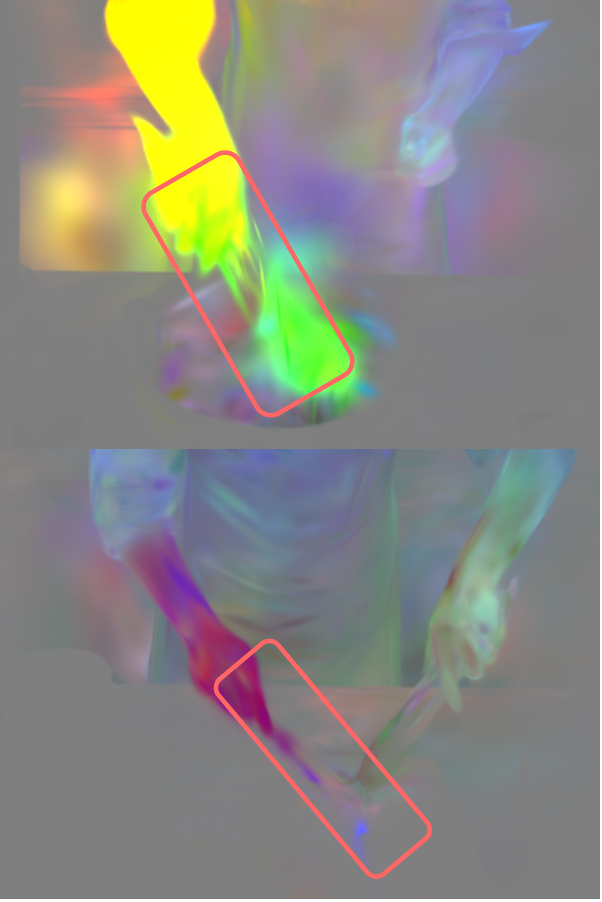} & 
     \includegraphics[width=0.09\linewidth]{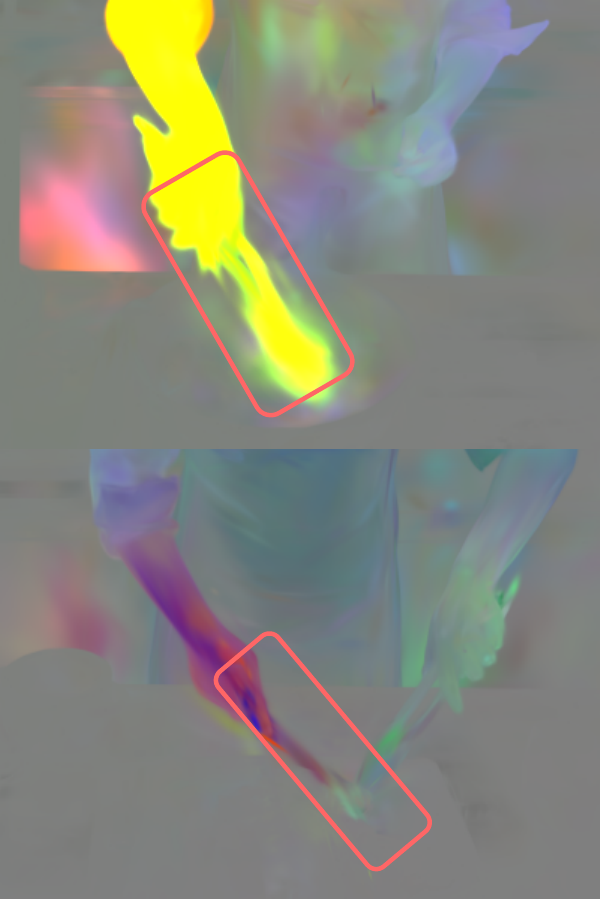} & 
     \includegraphics[width=0.09\linewidth]{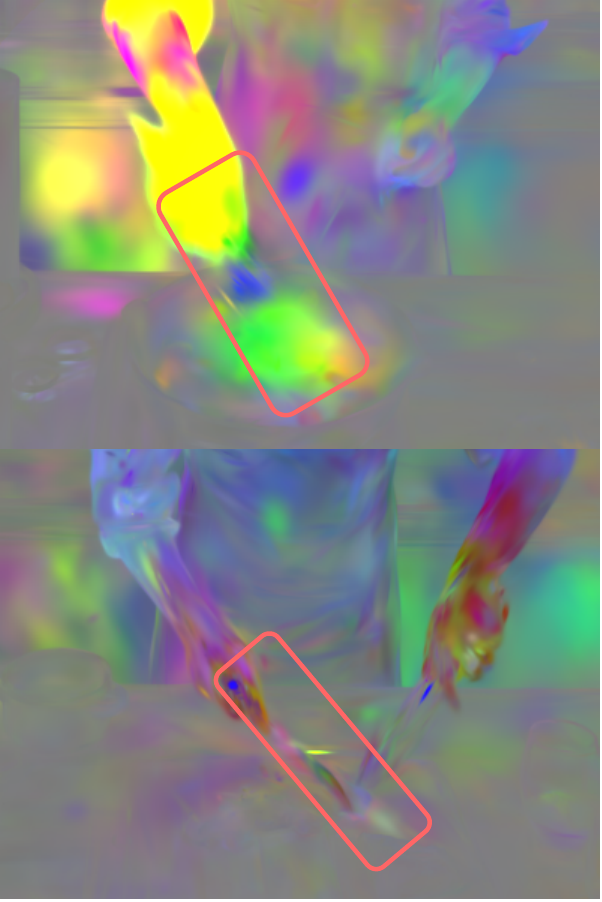} & 
     \includegraphics[width=0.09\linewidth]{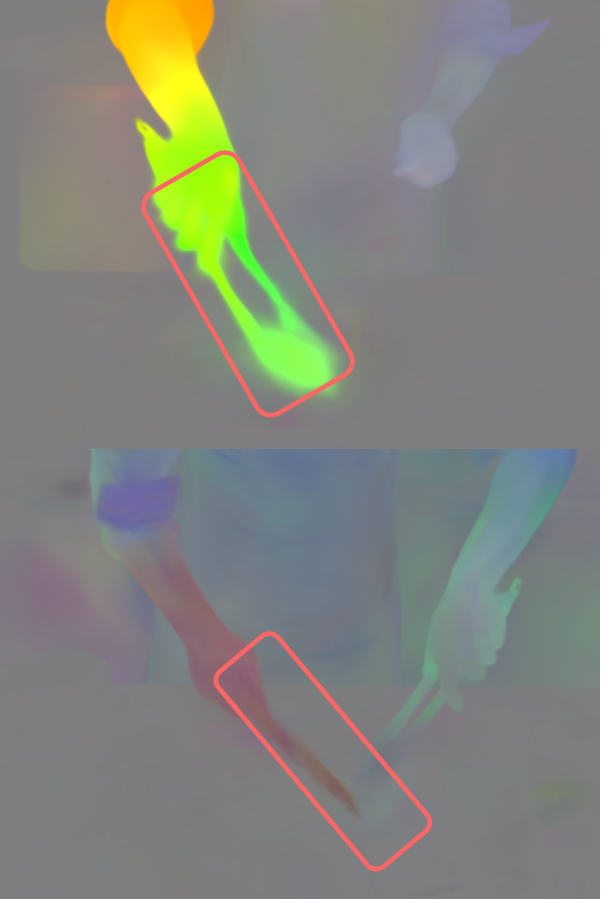} & 
     \includegraphics[width=0.09\linewidth]{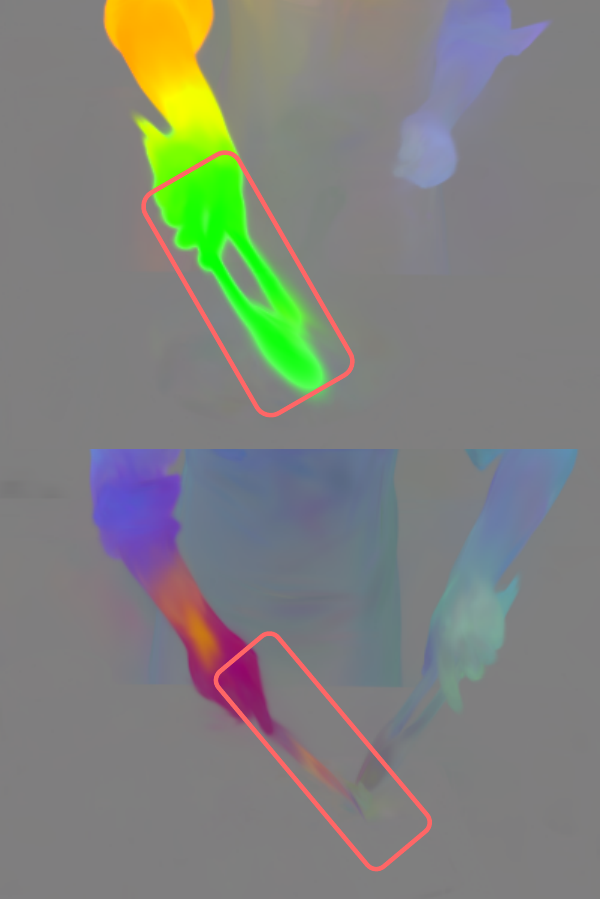}
\end{tabular}
\caption{Qualitative comparison of rendered images and scene flows on Neu3D~\cite{li2022neural} dataset. Our velocity regularizations preserve the handle with reflection batter than the baselines. The differences are more evident in rendered motion vectors when photometric reconstructions are mixed.}
\label{fig:neu3d_vis}
\end{figure*}

Neu3D~\cite{li2022neural} is a multi-camera-per-timestamp dataset with significantly more accurate camera positions and features more complex motions than the previously mentioned datasets. As shown in Tab.~\ref{tab:neu3d_comp}, the well-defined camera setup and large static regions yield comparable reconstruction metrics across all methods, while our approach achieves notable improvements in spatial coherence, enhancing the reconstruction of challenging motions, as illustrated in Fig.~\ref{fig:neu3d_vis}. The accurate multi-view camera setup reduces the performance gap of photometric metrics between MLP and grid-based methods. Notably, as evidenced by Moran's \textit{I} comparison between 4DGS~\cite{wu20234d} and Def.3D~\cite{yang2023deformable}, grid-based methods demonstrate superior spatial coherence when supplemented with total variation~\cite{fridovich2023k, wu20234d} loss. This issue likely stems from the limited capacity and temporal fluctuation bias of a single MLP with coupled 4D inputs to model long-term signals, aligning with the experimental observations in Sec.~\ref{sce:continuous_deformation_field}. However, our approach effectively mitigates this limitation by leveraging decoupling, low-rank factorization, and velocity regularization.

\subsubsection{Synthetic dataset} We further evaluate our method on the synthetic, one-camera-per-timestep dataset, D-NeRF~\cite{pumarola2021d}. The quantitative results are included in the supplementary material. Our method slightly lags behind state-of-the-art methods regarding fidelity. This performance gap can be attributed to minor temporal misalignments, which we believe are due to biases in our representation. A detailed discussion of this issue is provided in the supplementary material, and we leave its resolution as future work. Despite the slightly reduced fidelity performance, our representation consistently generates spatially coherent motions, and the misalignment is visually negligible, as shown in Fig.~\ref{fig:extra_dnerf}.

\section{Discussions \& Limitations}
\label{sec:discussion}
To further reason on our choice of \(\gamma_{\varphi(t)}(\cdot)\) over \(\gamma(\cdot)\), we illustrate the temporal weights, as shown in Fig.~\ref{fig:temporal_weights_visual}. The weights, aligned with the actual motions, demonstrated obvious symmetricity and periodicity. We expect this evidence to provide insights for the following works. We briefly discuss the potential applications based on our representation. Firstly, we qualitatively demonstrate the possibility of advecting points to verify the reasonableness of the derived velocity in Eq.~\ref{eq:velocity}. Specifically, we calculate the velocity of the last timestamp and propagate points with such constant speed. Secondly, we show the possibility of editing in the trajectory manifold. Furthermore, the established coherency enables us to propagate appearance modification in one frame to the whole sequence. The results are qualitatively shown in Fig.~\ref{fig:ep_motion} and \ref{fig:style_transfer}. 

Explicit spline interpolation demonstrates superior performance in large-scale interpolation and smooth motion modeling, which aligns with our intuition and is confirmed by our experimental findings. Since reconstruction errors are minimized through gradient descent and cannot be reduced to precisely zero, the misalignment issue caused by deviations from the ground truth is exacerbated by explicit spline interpolation. More specifically, we set the number of knots to half the number of training signals when no reliable prior on motion is available. This ensures a clean pipeline and a well-determined fit, avoiding the need to heuristically choose the number of knots or optimize interval lengths. To address the above issue, we apply a regularization term akin to penalized splines. This inevitably introduces slight offsets from the ground-truth trajectories, though this discrepancy is imperceptible in the absence of residual maps. Stemming from it, whether other curves (e.g., hybrid circular and elliptical~\cite{yuksel2020class}) can solve such unexpected bias is a worthy topic.

\section{Conclusion}
In this work, we demonstrate that splines, though a classical tool from applied mathematics, remain effective in modern settings, achieving strong results across several datasets with our implementation. In early experiments, we attempted to model deformations using independent splines without the support of Coord.-NNs, but achieved only modest results, even with strong regularization. Interestingly, our findings reveal that splines and Coord.-NNs complement each other well. Coord.-NNs provide coherent guidance for splines through their smoothness inductive bias, while the explicit regularization derived from spline interpolation reinforces this smoothness, particularly in challenging or ill-constrained scenarios. Moreover, our improved performances also come from the decoupling of spatial and temporal signals, which is a consideration often overlooked. Finally, we provide a principled and robust evaluation on spatial coherency concept on Moran's \textit{I}, which offers more informative insights in cases where photometric scores are ambiguous or inconsistent.

\begin{figure*}
    \centering
    \tiny
    \setlength{\tabcolsep}{0.0pt}
    \begin{tabular}{cc}
        \includegraphics[height=4.64cm]{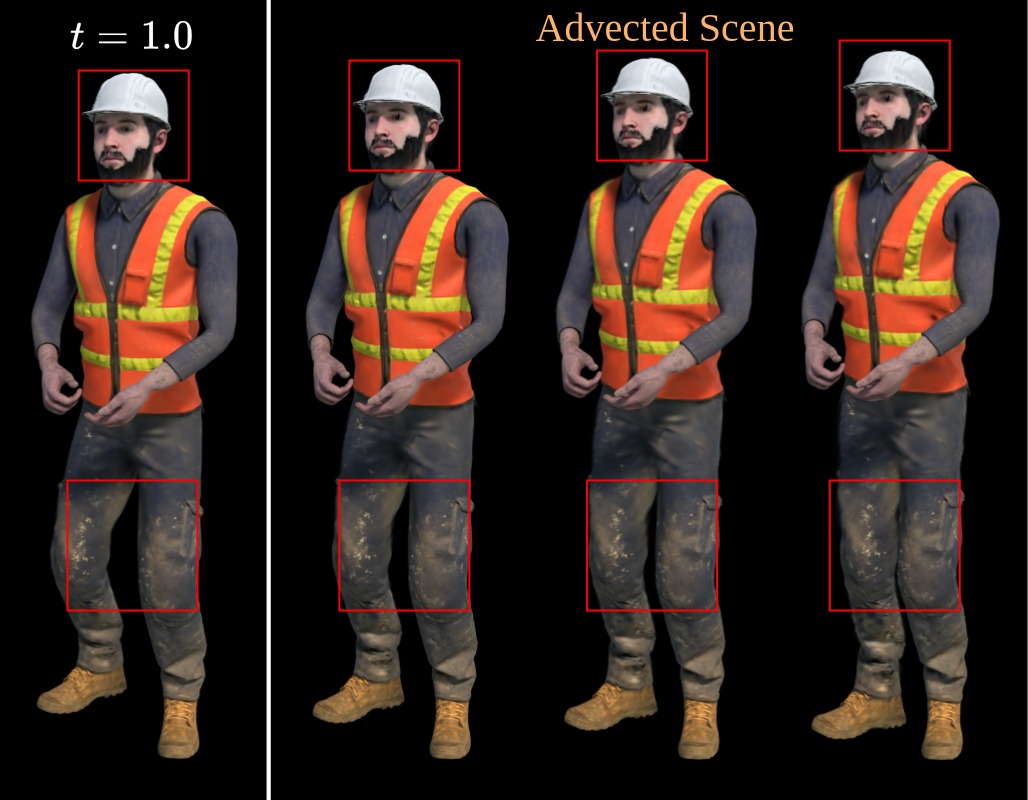} &
        \includegraphics[height=4.64cm]{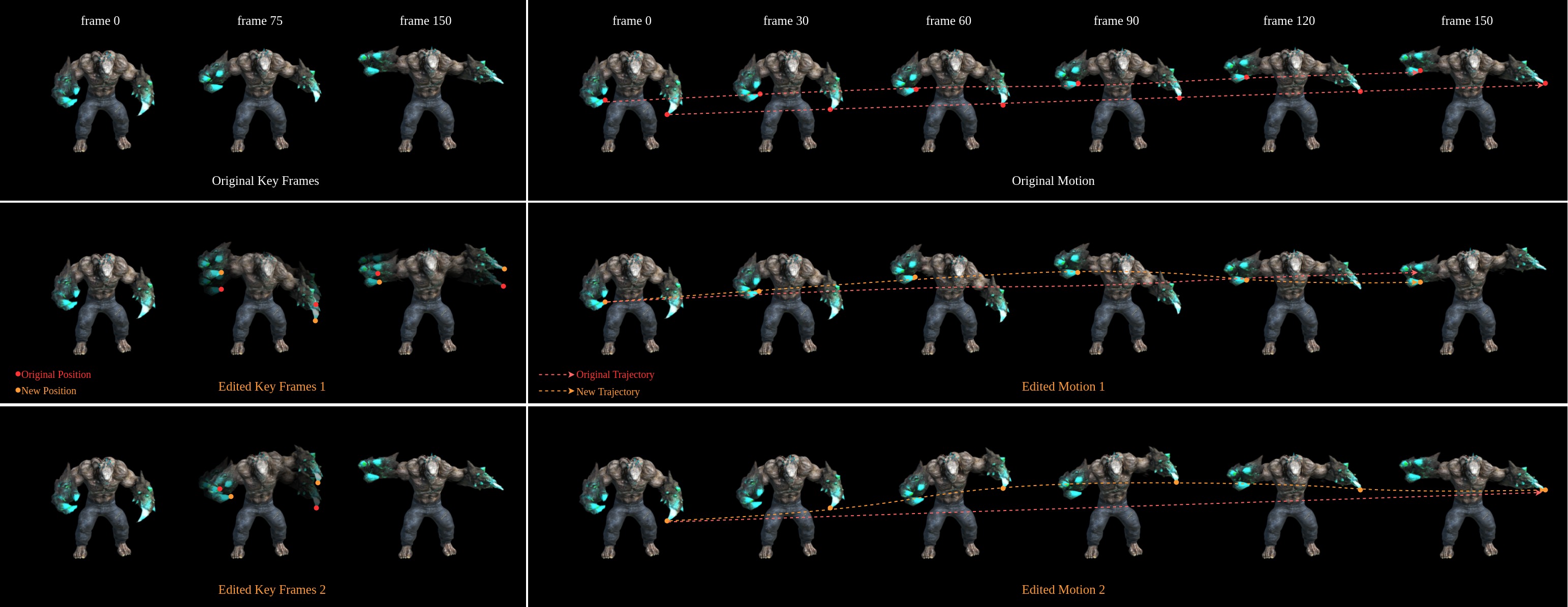}
    \end{tabular}
    \caption{Examples of motion advection and editing. \textbf{Left}: We propagate points in the last frame with the derived velocity. The plausible visual results showcase the effectiveness of \(\mathcal{L}_{v}\). \textbf{Right}: We adjust motions by only editing keyframes. In this work, we can only neatly achieve trifling modifications since the definition of '\textit{motion} editing' is still ambiguous, and LBS-based deformation is tedious for users. We leave a clearer description of such a task and a streamlined editing pipeline as future work (e.g., cage-based deformation).}
    \label{fig:ep_motion}
\end{figure*}
\begin{figure*}
\centering
\tiny
\setlength{\tabcolsep}{0.0pt}
\renewcommand{\arraystretch}{0.2}
\begin{tabular}{cccccccccccc}
    4DGS* & Def.3D & Grid4D & SF-PE-ResFields & SF-PE-ResFields & & 4DGS* & Def.3D & Grid4D & SF-PE-ResFields & SF-PE-ResFields\(\mathcal{x}\) & \\
    \cite{wu20234d} & \cite{yang2023deformable} & \cite{xu2024grid4d} & w/o \(\mathcal{L}_{acc}\) & w/ \(\mathcal{L}_{acc}\) & Ground Truth & \cite{wu20234d} & \cite{yang2023deformable} &  \cite{xu2024grid4d} & w/o \(\mathcal{L}_{acc}\) & w/ \(\mathcal{L}_{acc}\) & Ground Truth \\
    \includegraphics[width=0.083\linewidth]{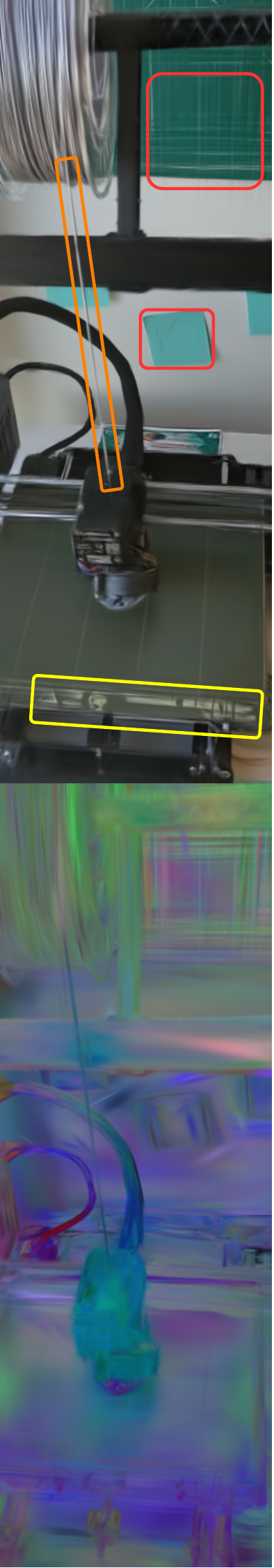} & \includegraphics[width=0.083\linewidth]{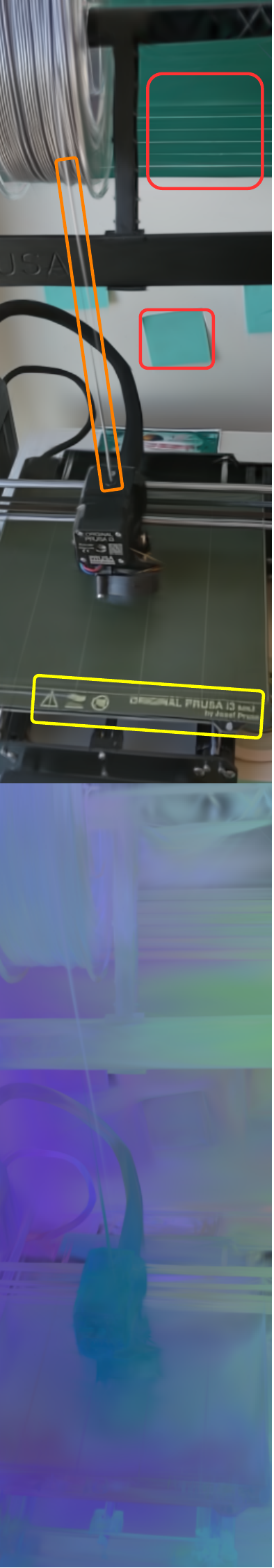} & \includegraphics[width=0.083\linewidth]{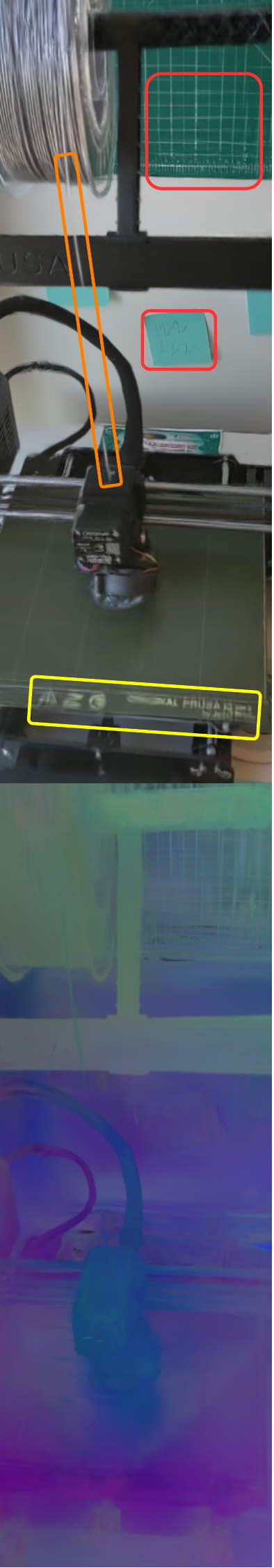} & \includegraphics[width=0.083\linewidth]{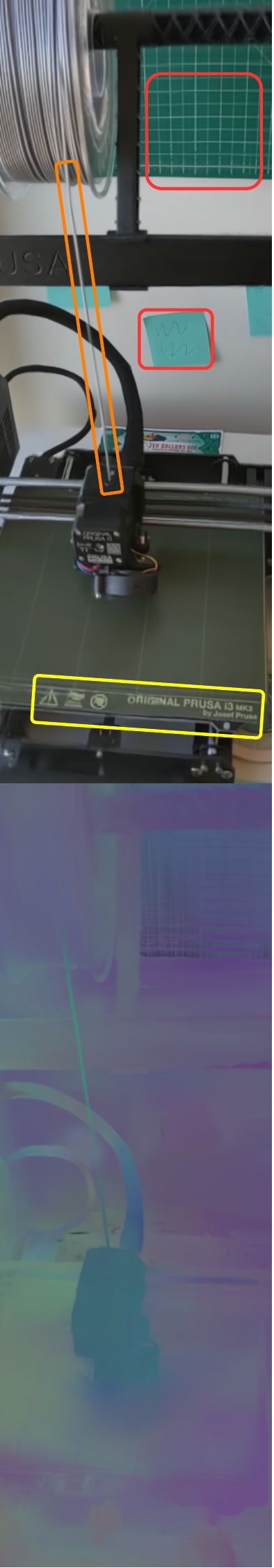} & 
    \includegraphics[width=0.083\linewidth]{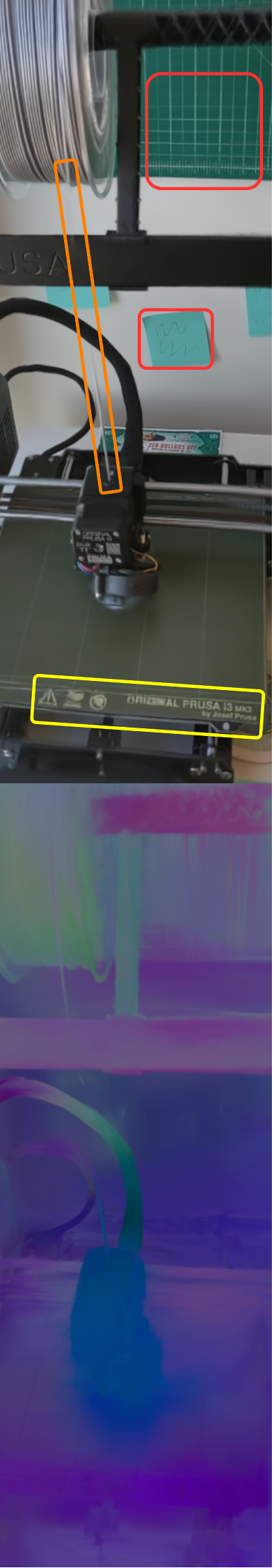} & \includegraphics[width=0.083\linewidth]{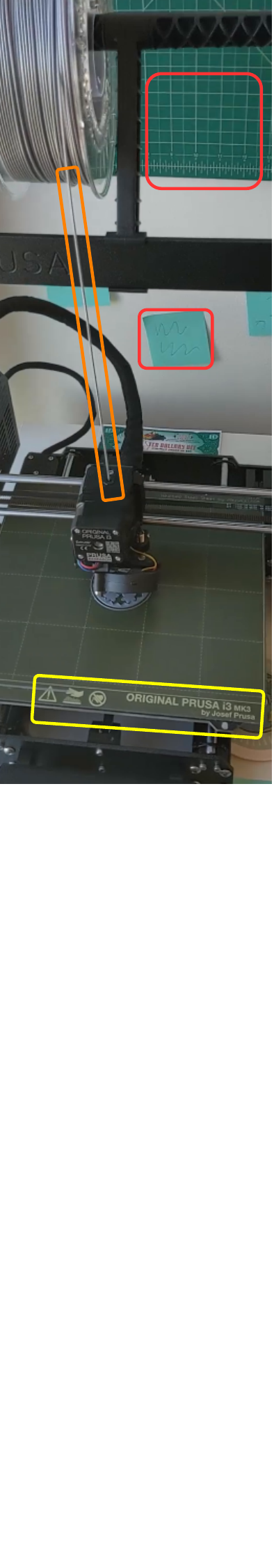} &
    \includegraphics[width=0.083\linewidth]{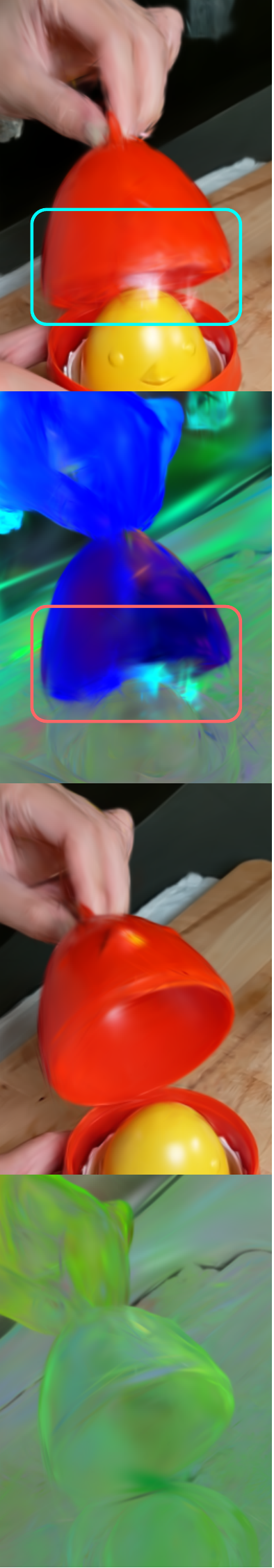} &
    \includegraphics[width=0.083\linewidth]{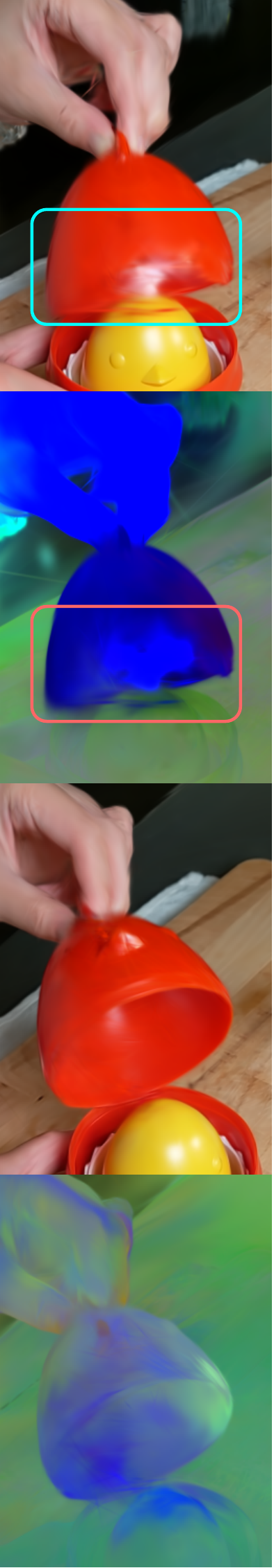} &
    \includegraphics[width=0.083\linewidth]{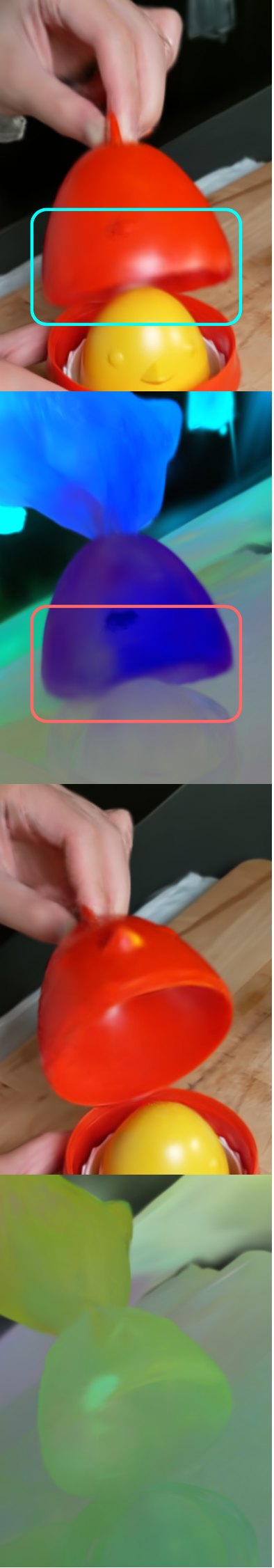} &
    \includegraphics[width=0.083\linewidth]{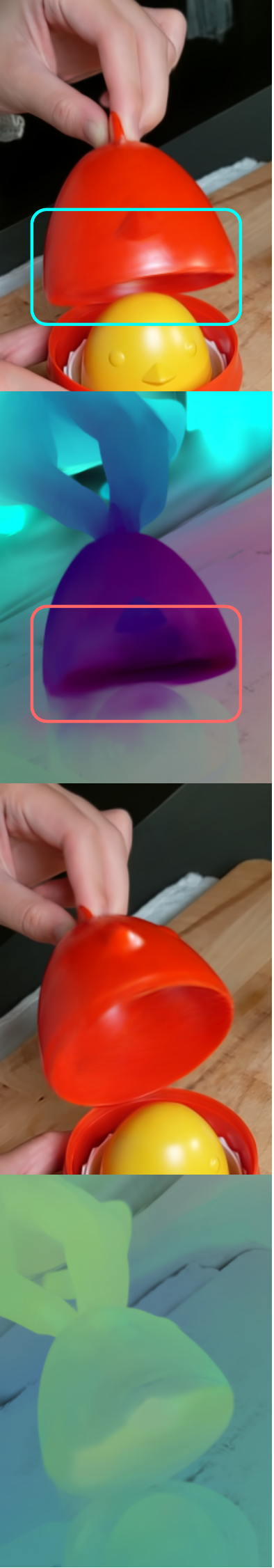} &
    \includegraphics[width=0.083\linewidth]{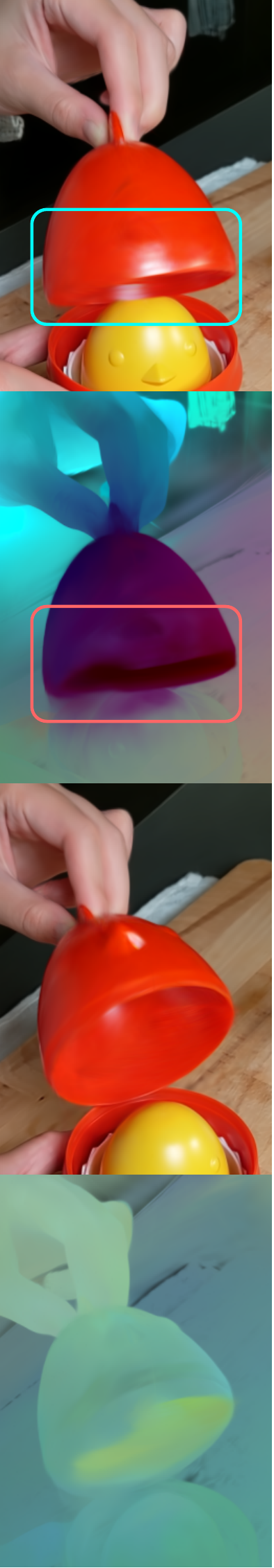} &
    \includegraphics[width=0.083\linewidth]{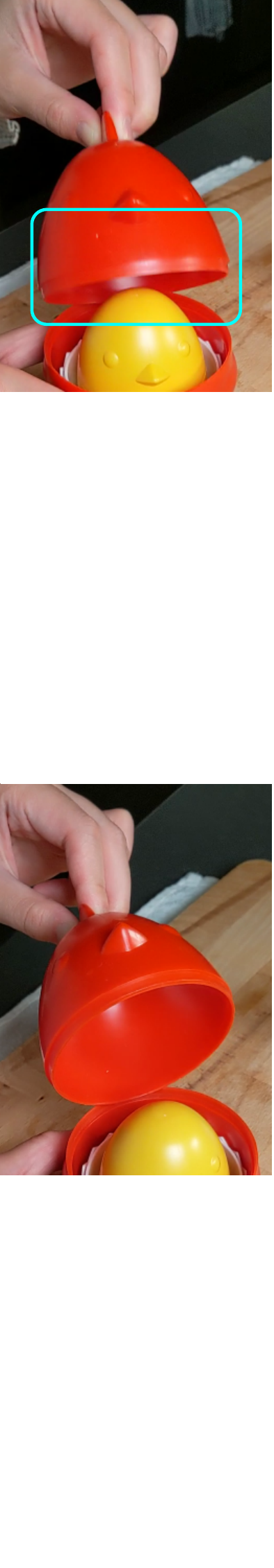}
\end{tabular}
\caption{Additional qualitative comparison of rendered images and scene flows on Hyper-NeRF~\cite{park2021hypernerf} dataset. Bounding boxes highlight major differences. We provide artifacts and failure cases of our method on this dataset in the supplementary material.}
\label{fig:extra_hyper_nerf_vis}
\end{figure*}
\begin{figure}
    \centering
    \includegraphics[width=0.80\linewidth]{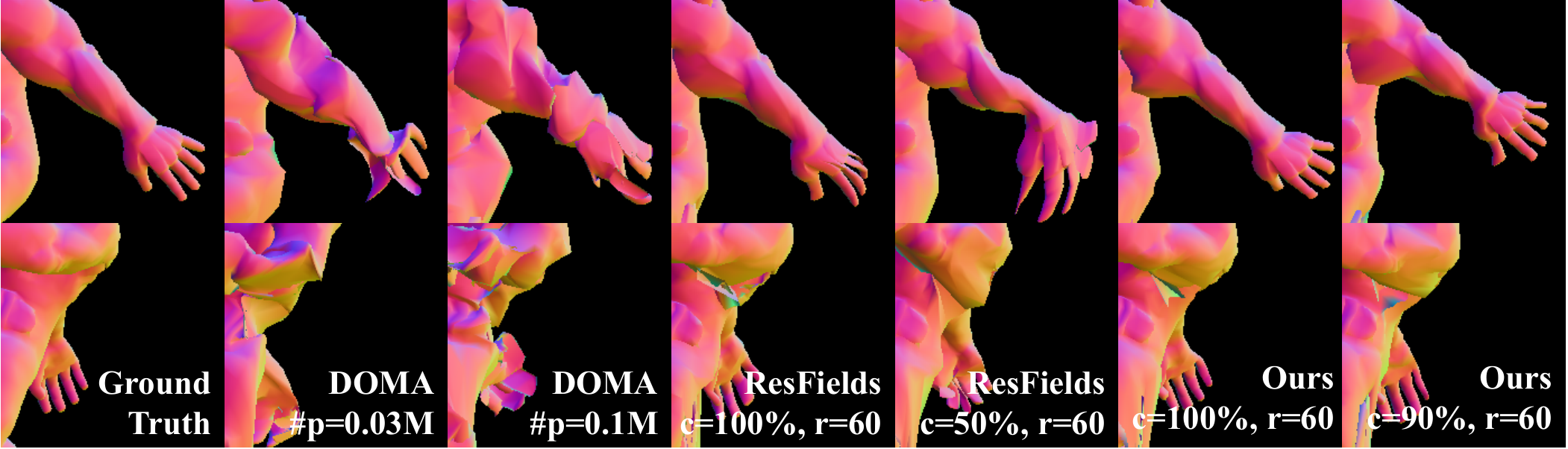}
    \caption{We present additional scene flow interpolation results, a task with significant practical value yet remains underexplored. Our model demonstrates strong performance and can serve as a baseline for future research.}
    \label{fig:scene_flow_interp_extra}
\end{figure}
\begin{figure}
    \centering
    \includegraphics[width=0.75\linewidth]{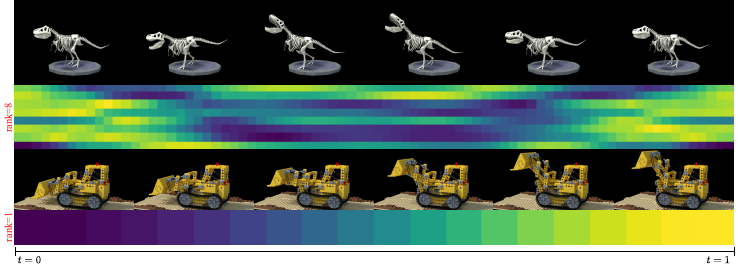}
    \caption{Visualization of \(\textbf{v}\in\mathbb{R}^{rank}\) in Eq.~\ref{eq:res_weights}. We concatenate the temporal weights of SF-Triplanes along the temporal dimension, i.e., each \textbf{column} in the heat map is a vector representing a deformed state. Not surprisingly, scene \textit{lego} can be reconstructed with \(rank=1\).}
    \label{fig:temporal_weights_visual}
\end{figure}
\begin{figure*}
\centering
\small
\setlength{\tabcolsep}{0.0pt}
\renewcommand{\arraystretch}{0.2}
\begin{tabular}{ccccccc}
    & Grid4D & & Grid4D & & Grid4D & \\
    Ground Truth & \cite{xu2024grid4d} & Ours & \cite{xu2024grid4d} & Ours & \cite{xu2024grid4d} & Ours \\
    \includegraphics[width=0.142\linewidth]{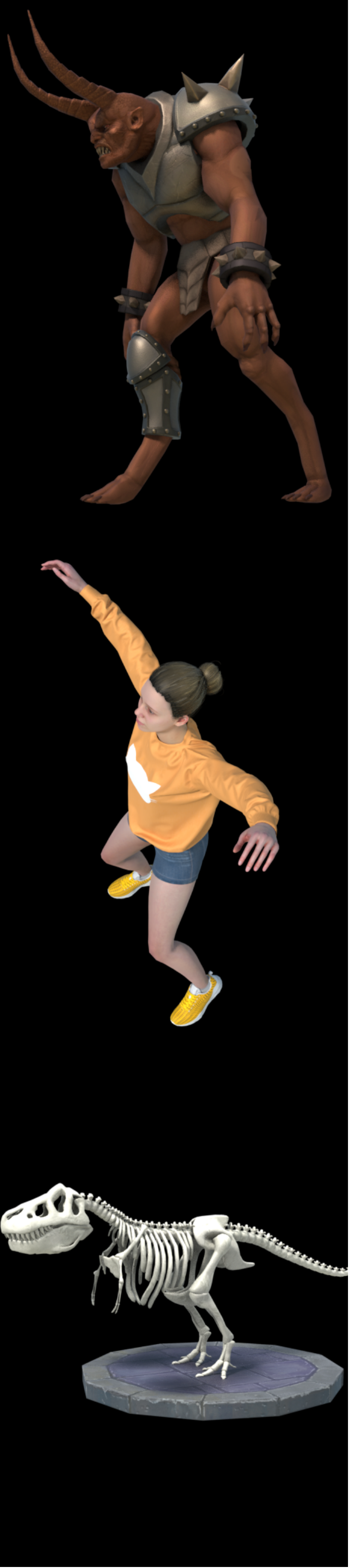} & 
    \includegraphics[width=0.142\linewidth]{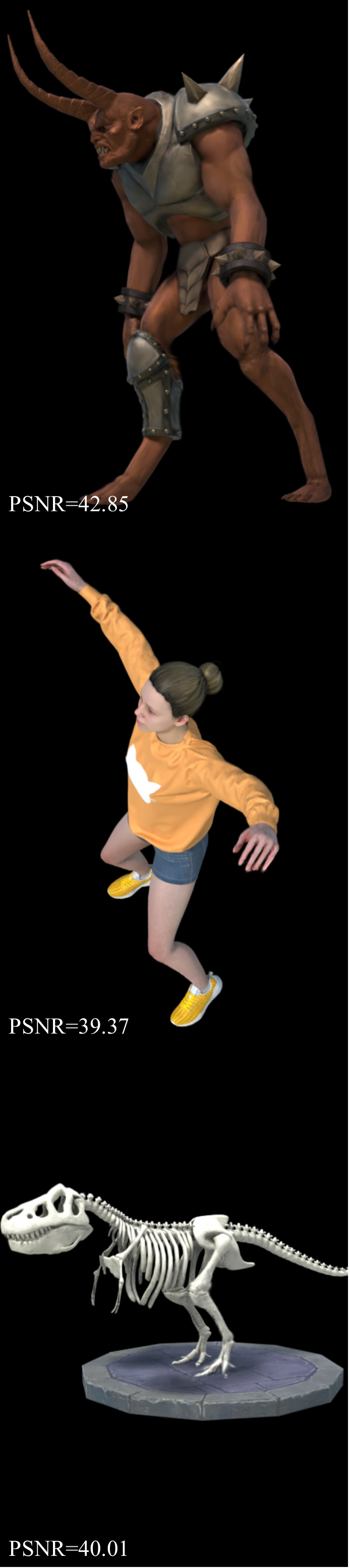} & 
    \includegraphics[width=0.142\linewidth]{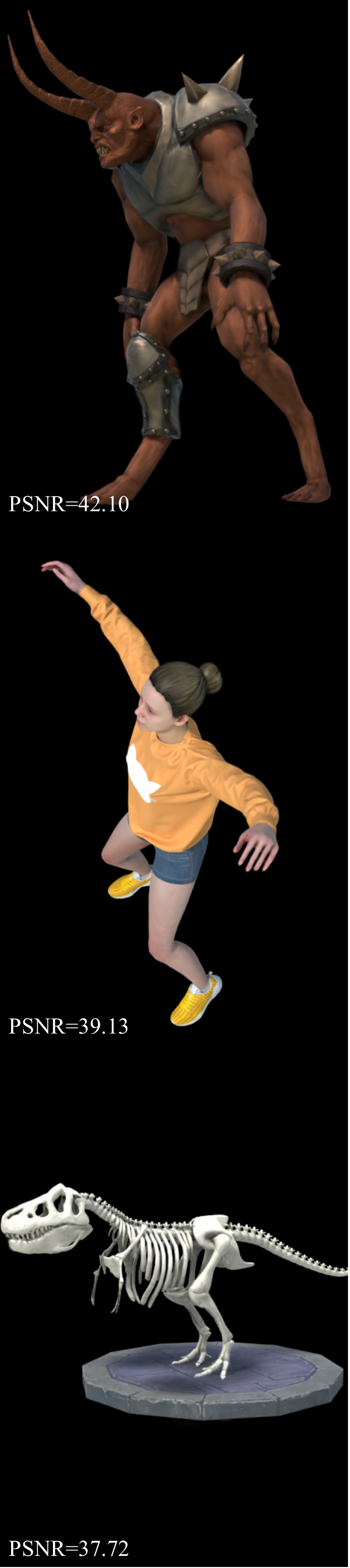} & 
    \includegraphics[width=0.142\linewidth]{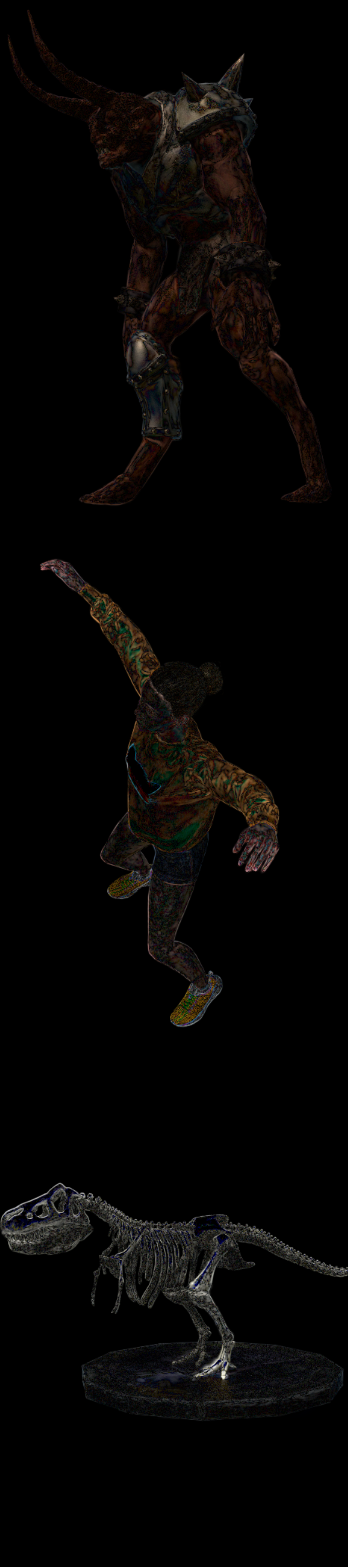} & 
    \includegraphics[width=0.142\linewidth]{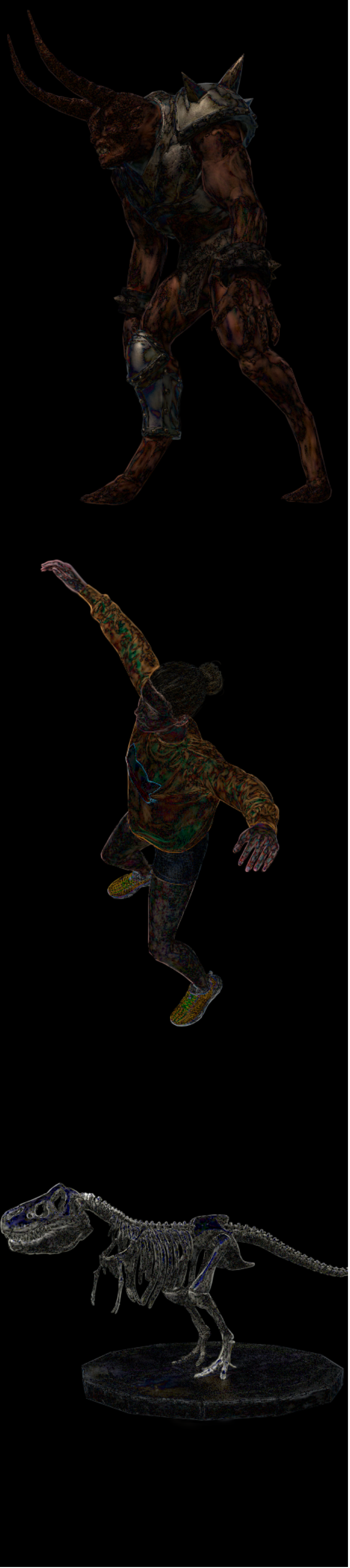} & 
    \includegraphics[width=0.142\linewidth]{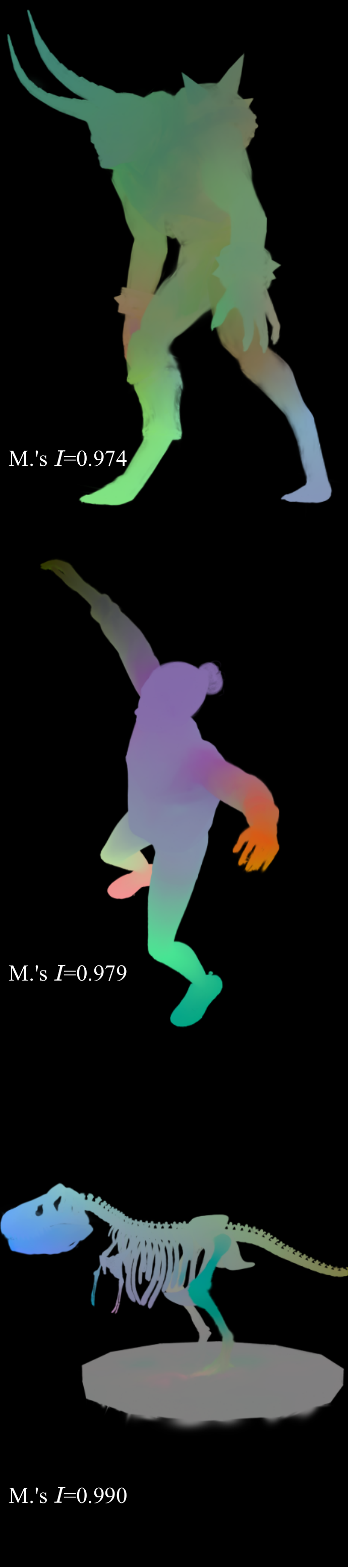} & 
    \includegraphics[width=0.142\linewidth]{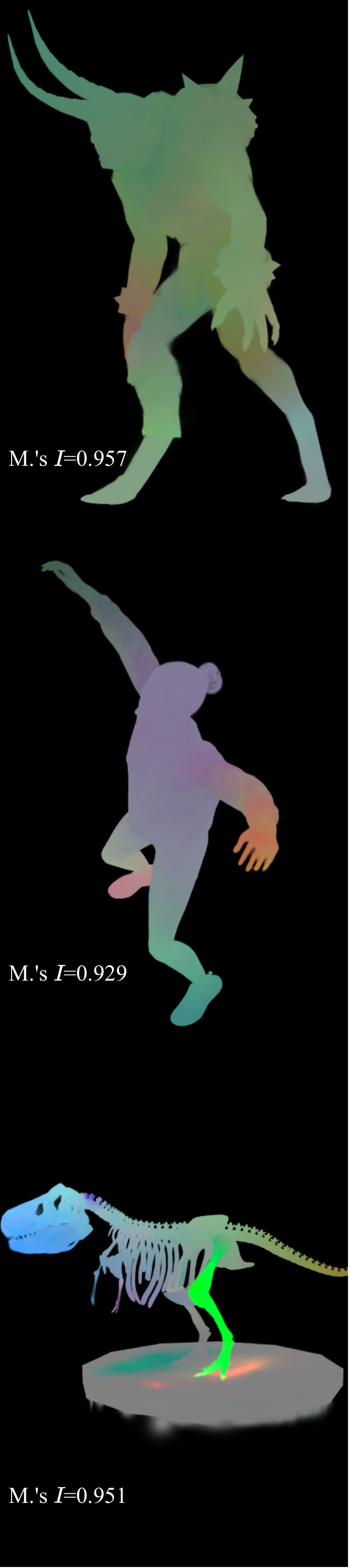} 
\end{tabular}
\caption{We present the comparison between our method and Grid4D~\cite{xu2024grid4d}, the SOTA method on synthetic D-NeRF~\cite{pumarola2021d} dataset. While our results lag slightly behind on this dataset, our method is capable of reasonably reconstructing challenging regions. Our evaluations of spatial coherency by using Moran's \textit{I} align with these observations, verifying informativeness of this metric. As highlighted in the main paper, our explicit regularization demonstrates superior performance in real-world scenarios. We provide artifacts of our method on this dataset in the supplementary material.}
\label{fig:extra_dnerf}
\end{figure*}
\begin{figure*}
    \centering
    \includegraphics[width=0.972\linewidth]{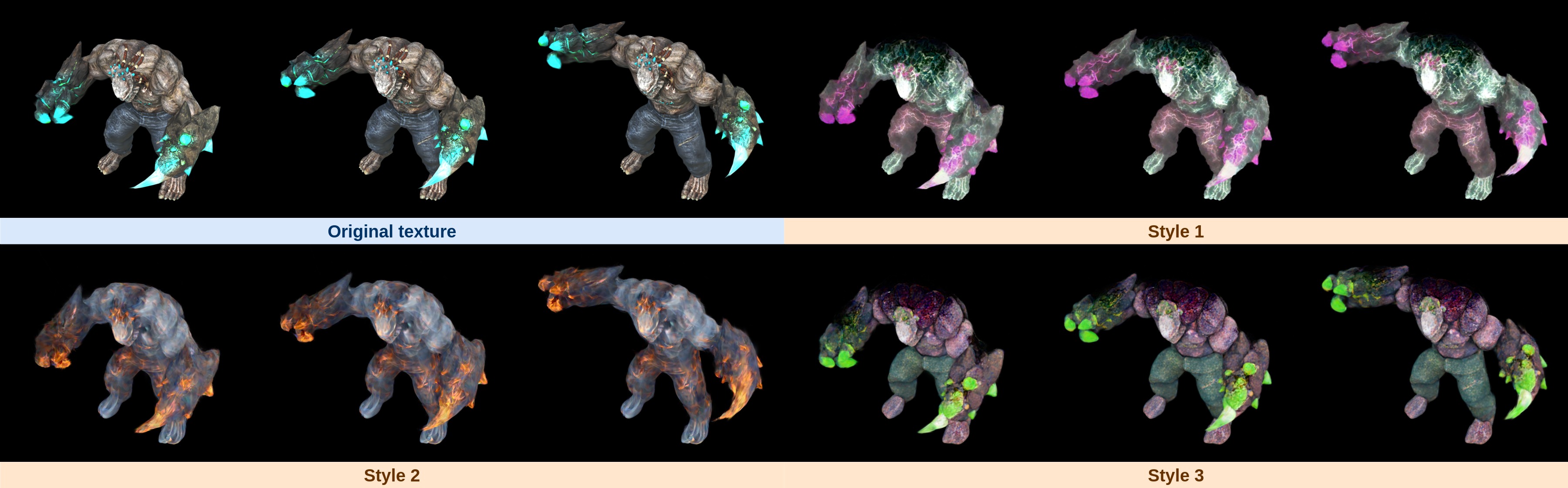}
    \caption{The canonical space of our variant, SF-PE-ResFields, achieves improved consistency with dynamic observations. As a result, we can perform style transfer on the avatar's texture in a straightforward manner. Specifically, by transferring the style in the static canonical space, the new texture can be seamlessly propagated throughout the entire sequence. While new points are densified without constraint during style transfer, the continuity of the deformation field and regularization on the rendered depth map allow us to achieve these appealing results without the need for additional registration (e.g., Chamfer distance).}
    \label{fig:style_transfer}
\end{figure*}

\begin{acks}
We gratefully acknowledge the creators of the open-source datasets used in this work. The Deer, Astra SambaDancing, and Pumpkinhulk SwingDancing scenes shown in Fig.~\ref{fig:teaser}, \ref{fig:scene_flow_interp_main}, and \ref{fig:scene_flow_interp_extra} were created by Yang Li et al.~\cite{li20214dcomplete}. The Flame Steak and Cook Spinach scenes in Fig.~\ref{fig:teaser} and \ref{fig:neu3d_vis} were created by Zhaoyang Lv et al.~\cite{li2022neural}. The Bell and Press scenes in Fig.\ref{fig:nerfds_scene_flow} were created by Zhiwen Yan et al.~\cite{yan2023nerf}. The Torch Chocolate, Slice Banana, 3D Printer, and ChickChicken scenes in Fig.~\ref{fig:hyper_nerf_vis} and \ref{fig:extra_hyper_nerf_vis} were created by Keunhong Park et al.~\cite{park2021hypernerf}. The Standup, Mutant, T-Rex, and Lego scenes in Fig.~\ref{fig:ep_motion}, \ref{fig:temporal_weights_visual}, \ref{fig:extra_dnerf}, and \ref{fig:style_transfer} were created by Albert Pumarola et al.~\cite{pumarola2021d}.
\end{acks}

\bibliographystyle{ACM-Reference-Format}
\bibliography{sample-bibliography}

\clearpage
\appendix

\section{Implementation Details}
\subsection{Motion Correlation}
We use spatial autocorrelation~\cite{moran1950notes} of motion vectors to quantitatively assess spatial coherency. To compute the motion vectors for a scene with \(T_{test}\) frames in the testing set, we first calculate the offsets between consecutive frames, which can be formulated as:
\begin{equation}
    \hat{v}_i^t=x_i(t+1)-x_i(t), i=1,...,N_p, t=1,...,T_{test}-1
    \label{eq:motion_vect}
\end{equation}
where we use \(\hat{v}_i(t)\) to represent the discrete motion vector, distinguishing it from the closed-form velocity \(v_i\). Next, we calculate Moran's \textit{I} for each point and timestep, denoted as \(I^t(x_i)\), and then compute the average across all timesteps.

For each \(\hat{v}_i(t)\) we gather its \(K\) nearest neighbors based on the coordinates of its starting point \(x_i(t)\), and then measure \(I^t(x_i)\) with the following equations:
\begin{equation}
    I^t(x_i)=\frac{K}{\sum_{i=1}^K\sum_{j=1}^Kw^t_{ij}}\frac{\sum_{i=1}^K\sum_{j=1}^Kw^t_{ij}\hat{v}^t_i\hat{v}^t_j}{\sum_i^K\hat{v}_i^t},
    \label{eq:m1_eq_1}
\end{equation}

\begin{equation}
    w^t_{ij}=
    \begin{cases}
        ||x_i(t)-x_j(t)||_2^{-1} & \text{if i \(\neq\) j}, \\
        0 & \text{otherwise}.
    \end{cases}
\end{equation}

We set \(K=10\) across all experiments, and \(T_{test}\) is the number of frames in testing sequence following the official train/test splits. We refer readers to the supplementary video for an intuitive explanation.

\subsection{Time-Variant Spatial Encoding (TVSE)}
We first provide a detailed overview of our design and related functions, which were omitted in the paper due to space constraints, in Algorithm~\ref{alg:sdf}. Notably, \(t_{start}\) and \(t_{end}\) are integer indices which are used to query the time-specific vectors \(\textbf{v}_{t_{start}}\in\mathbb{R}^{rank}\) and \(\textbf{v}_{t_{end}}\in\mathbb{R}^{rank}\). For a more complete understanding, we recommend that readers refer to Eqs. 2–3 in the main paper.

\begin{algorithm}
\SetAlgoNoLine
\KwIn{Canonical coordinates: \(\textbf{X}^{c}=\{\text{x}_i^c\in\mathbb{R}^3\}_{i=1,...,N_p}\); query time: \(t_{query}\in[0.0, 1.0]\),}

\KwOut{Deformed coordinates at timestep \(t_{query}\): \(\textbf{X}(t_{query})=\{\text{x}_i(t_{query})\in\mathbb{R}^3\}_{i=1,...,N_p}\),}

\textbf{Step 1.} Calculate the length of time interval \(\tau\): \\
\qquad \(\tau=1/(N-1)\) where \(N\) is the number of knots,

\textbf{Step 2.} Determine the starting and ending temporal index: \\
\qquad \(t_{start}=\text{CLAMP}(\lfloor t_{query}/\tau\rfloor,\text{min}=0,\text{max}=N-2)\), \\
\qquad \(t_{end}=t_{start}+1\),

\textbf{Step 3.} Normalize to relative time: \\
\qquad \(\bar{t}=t_{query}\times(N-1)-t_{start}\),

\textbf{Step 4.} Calculate offset and tangent of starting timestep: \\
\qquad \(\triangle \text{x}_i(t_{start}),\dot{\text{x}}_i(t_{start})=\Phi_{\theta}(\text{x}_i^c,t_{start})\),\\
\qquad \(\text{x}_i(t_{start})=\text{x}_i^c+\triangle \text{x}_i(t_{start})\),

\textbf{Step 5.} Calculate offset and tangent of ending timestep: \\
\qquad \(\triangle \text{x}_i(t_{end}),\dot{\text{x}}_i(t_{end})=\Phi_{\theta}(\text{x}_i^c,t_{end})\),\\
\qquad \(\text{x}_i(t_{end})=\text{x}_i^c+\triangle \text{x}_i(t_{end})\),

\textbf{Step 6.} Interpolate cubic polynomial: \\
\qquad \(\text{x}_i(t_{query})=\mathcal{F}(\bar{t},\text{x}_i(t_{start}),\dot{\text{x}}_i(t_{start}),\text{x}_{i}(t_{end}),\dot{\text{x}}(t_{end}))\) \\
\qquad\qquad\quad \(=(2\bar{t}^3-3\bar{t}^2+1)\text{x}_i(t_{start})+(\bar{t}^3-2\bar{t}^2+\bar{t})\dot{\text{x}}_i(t_{start})+\) \\
\qquad\qquad\qquad \((-2\bar{t}^3+3\bar{t}^2)\text{x}_i(t_{end})+(\bar{t}^3-\bar{t}^2)\dot{\text{x}}(t_{end})\),

\textbf{Step 7.} Derive velocity: \\
\qquad \(v_i(t_{query})=(6\bar{t}^2-6\bar{t})\text{x}_i(t_{start})+(3\bar{t}^2-4\bar{t}+1)\dot{\text{x}}(t_{start})+\) \\
\qquad\qquad\qquad\quad \((-6\bar{t}^2+6\bar{t})\text{x}_i(t_{end})+(3\bar{t}^2-2\bar{t})\dot{\text{x}}_i(t_{end})\),

\textbf{Step 8.} Derive acceleration: \\
\qquad \(a_i(t_{query})=(12\bar{t}-6)\text{x}_i(t_{start})+(6\bar{t}-4)\dot{\text{x}}_i(t_{start})+\) \\
\qquad\qquad\qquad\quad \((-12\bar{t}+6)\text{x}_i(t_{end})+(6\bar{t}-2)\dot{\text{x}}_i(t_{end})\).

\caption{Spline Deformation Field}
\label{alg:sdf}
\end{algorithm}

For reproducibility, we provide a detailed formulation of TVSE and recommend that readers refer to Fig. 3 and Eq. 5 in the main paper for notation. The signature at each timestep is determined by a vector \(\textbf{v}_t\). Given \(N\) knots in the temporal domain, we obtain \(N\) such vectors, collectively denotes as (\(\textbf{V}=\{\textbf{v}_t|\textbf{v}_t\in\mathbb{R}^{rank}, t=1,2,\cdots,N\}\in\mathbb{R}^{N\times rank}\)), which represent the entire dynamic scene. For a more intuitive understanding of the role and meaning of \(\textbf{V}\), we refer readers to Fig. 11 in the main paper.

\subsubsection{PE-ResFields}
\begin{figure}
    \centering
    \includegraphics[width=0.6\linewidth]{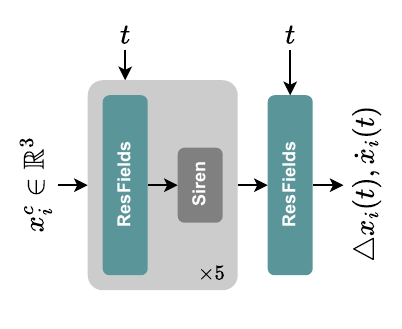}
    \caption{The network architecture of SF-Siren-ResFields follows the same design as the compared ResFields~\cite{mihajlovic2023resfields}, with the only difference being that we predict tangents (velocities) in addition to offsets.}
    \label{fig:sceen_flow_net_arch}
\end{figure}

\begin{figure}
    \centering
    \includegraphics[width=\linewidth]{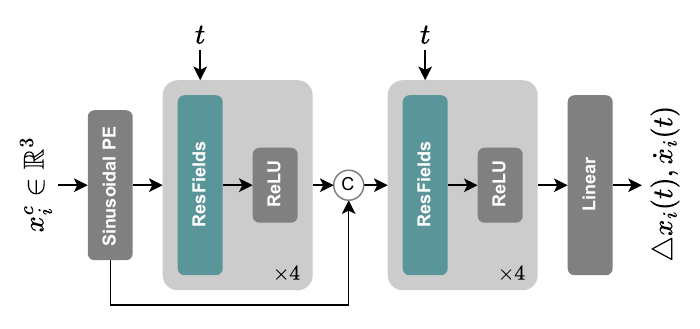}
    \caption{The network architecture of SF-PE-ResFields replaces the periodic activation layers with ReLU and incorporates sinusoidal positional encoding at the beginning of the neural network. The architecture design, such as concatenation, and the width and depth of the MLP, is derived from Def.3D~\cite{yang2023deformable}.}
    \label{fig:dy_gs_net_arch}
\end{figure}

We adopt the off-the-shelf design from ResFields~\cite{mihajlovic2024ResFields} for our SF-Siren-ResFields and SF-PE-ResFields. The overall architecture is intuitively illustrated in Fig.\ref{fig:sceen_flow_net_arch} and Fig.\ref{fig:dy_gs_net_arch}. The formulation of the linear layer in ResFields~\cite{mihajlovic2024ResFields} can be formulated as:
\begin{equation}
    \begin{array}{c}
        \text{feature}_{out}=\text{feature}_{in}\cdot W(t) + \text{bias},\\
        W(t)=\text{W}_{base}+\sum_{r=1}^{rank}\text{v}_t[r]\cdot W_{res}[r] \\
        W(t)\in\mathbb{R}^{C_{in}\times C_{out}},
    \end{array}
\end{equation}
where \(\text{bias}\) is same to the bias in \lstinline{torch.nn.Linear}. We refer readers to the original script and official implementation of ResFields~\cite{mihajlovic2024ResFields} for further details.

\subsubsection{Time-variant Triplanes/Triaxes}
We begin by outlining our motivation for adopting spline representations over grid-based methods and providing a detailed description of their implementation.

We demonstrate that the three main components of our method, spline interpolation, decoupled spatiotemporal encoding, and explicit regularization, are independent of the specific encoding mechanism, as shown in Tables 2–4 of the main paper. As a result, our approach can benefit from future advancements in encoding strategies.

\begin{figure}
    \centering
    \includegraphics[width=\linewidth]{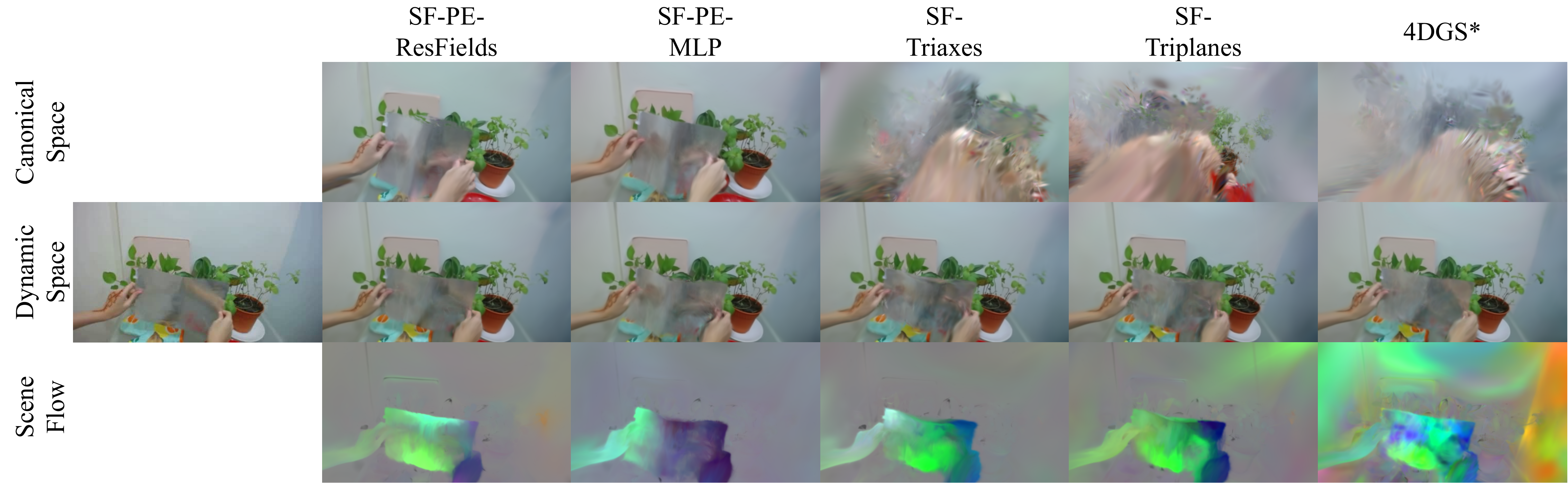}
    \caption{Comparison of MLP and grid-based methods in canonical space: MLP's canonical space demonstrates better structure and alignment with dynamic observations due to its global continuity. However, such consistency is not critical for maintaining spatial coherence. As shown in 4DGS* and our methods, SF-Triaxes and Triplanes, even noisy canonical spaces can achieve comparable photometric performance in dynamic spaces. Nonetheless, relying solely on implicit perturbation is not sufficient to fully resolve the artifacts introduced by discrete grids.}
    \label{fig:canon_comp_nerfds}
\end{figure}

\begin{figure}
    \centering
    \includegraphics[width=\linewidth]{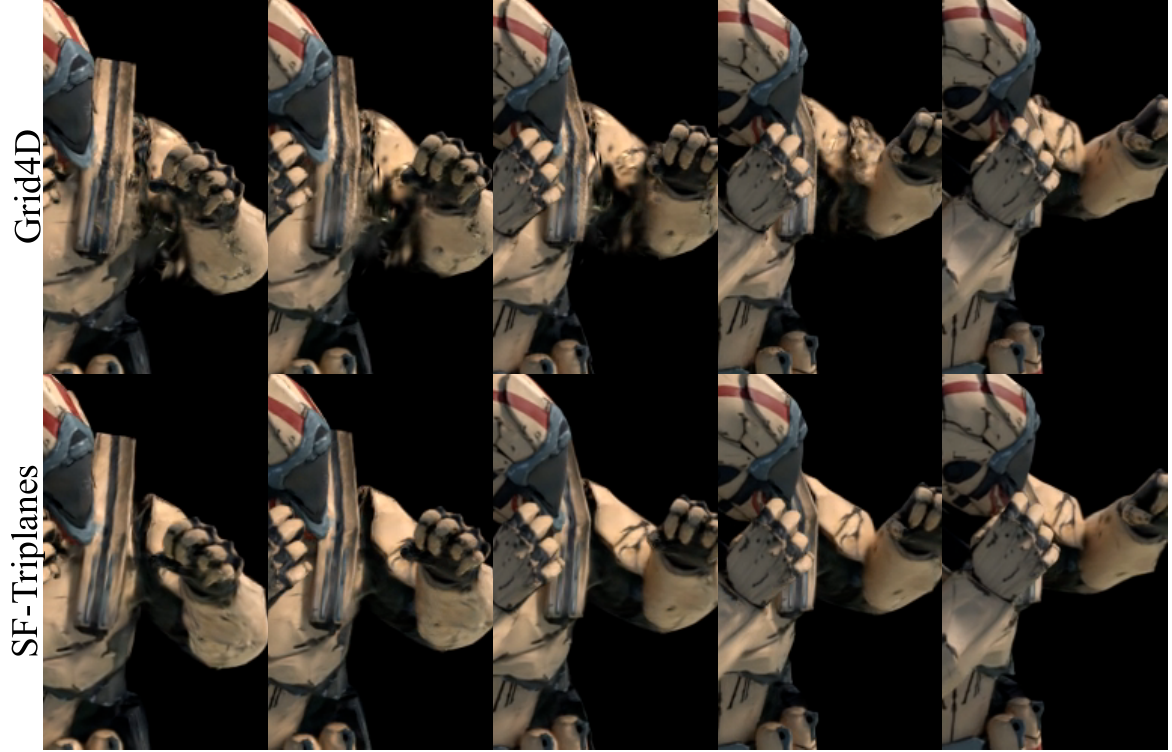}
    \caption{While Grid4D~\cite{xu2024grid4d} effectively enhances spatial coherence compared to 4DGS~\cite{wu20234d} in this synthetic dataset, abnormal floaters still appear in certain extreme cases. Furthermore, as shown in Figures 6–7 of the main paper, the effectiveness of perturbation or total variation loss remains limited in addressing the challenges posed by real-world scenarios.}
    \label{fig:hook_floater}
\end{figure}

Additionally, we aim to highlight a key distinction between MLP and grid-based methods that contributes to the observed differences in performance, as illustrated in Fig.\ref{fig:canon_comp_nerfds}. Grid-based methods tend to produce irregular canonical points. While interpolation is smooth within each grid voxel, the independently optimized grid values increase the risk of overfitting to training views. This issue has been effectively addressed by Grid4D\cite{xu2024grid4d} on the synthetic D-NeRF~\cite{pumarola2021d} dataset, as shown in Fig. 12 of the main paper. However, abnormal floaters are still visible in corner cases. While using triplanes as the backbone, which is sparser and less advanced than Hash encoding, our explicit regularization proves more effective in mitigating block-like artifacts in grid-based methods compared to directly regularizing the grid values, as shown in Fig.\ref{fig:hook_floater}.

We formally describe the operation of Time-variant Triplanes with the following equations:
\begin{equation}
\begin{array}{c}
    \text{P}^{XY}(t)=\text{P}^{XY}_{base}+\sum_{r=1}^{rank}\text{v}_t[r]\cdot \text{P}^{XY}_{res}[r], \\
    \text{P}^{YZ}(t)=\text{P}^{YZ}_{base}+\sum_{r=1}^{rank}\text{v}_t[r]\cdot \text{P}^{YZ}_{res}[r], \\
    \text{P}^{XZ}(t)=\text{P}^{XZ}_{base}+\sum_{r=1}^{rank}\text{v}_t[r]\cdot \text{P}^{XZ}_{res}[r], \\
    \text{P}^{\cdot\cdot}(t)\in\mathbb{R}^{D\times D}
\end{array}
\end{equation}
where \(\text{P}^{XY}(t)\) represents the plane storing features at timestep \(t\) to be queried by the \((x,y)\) coordinates. Here, \(\textbf{v}_t\in\mathbb{R}^{rank}\) is the temporal code corresponding to timestep \(t\), and \(\textbf{P}^{XY}_{res}\in\mathbb{R}^{rank\times D\times D}\) denotes the temporal templates which will be aggregated by \(\textbf{v}_t\) via weighted summation. \(D\) denotes the resolution of planes. The same explanation applies to \(\text{P}^{YZ}(t)\) and \(\text{P}^{XZ}(t)\).

We are akin to 4DGS~\cite{wu20234d} and K-Planes~\cite{fridovich2023k} for the multi-level grid sampling and feature aggregation. More specifically, we use element-wise multiplication to aggregate features interpolated from the three planes:
\begin{equation}
\begin{array}{c}
    \text{feat}(x,y,t)=\text{P}^{XY}(t)[x,y], \\
    \text{feat}(y,z,t)=\text{P}^{YZ}(t)[y,z], \\
    \text{feat}(x,z,t)=\text{P}^{XZ}(t)[x,z], \\
    \text{feat}(x,y,z,t)=\text{feat}(x,y,t)\odot\text{feat}(y,z,t)\odot\text{feat}(x,z,t), \\
    \text{feat}(\cdots)\in\mathbb{R}^{d},
    \label{eq:triplane_agg}
\end{array}
\end{equation}
where \(\text{P}^{XY}(t)[x,y]\) represents getting feature values at position \((x,y)\) by interpolating its neighboring grid vertices. \(d\) denotes the dimension of the features stored in grid vertices.

Similarly, the derivation of Time-Variant Axes is formulated as:
\begin{equation}
\begin{array}{c}
\text{A}^{X}(t)=\text{A}^{X}_{base}+\sum_{r=1}^{rank}\text{v}_t[r]\cdot \text{A}^{X}_{res}[r], \\
\text{A}^{Y}(t)=\text{A}^{Y}_{base}+\sum_{r=1}^{rank}\text{v}_t[r]\cdot \text{A}^{Y}_{res}[r], \\
\text{A}^{Z}(t)=\text{A}^{Z}_{base}+\sum_{r=1}^{rank}\text{v}_t[r]\cdot \text{A}^{Z}_{res}[r], \\
\text{A}^{\cdot}(t)\in\mathbb{R}^{D},
\end{array}
\end{equation}
where \(\text{A}^{X}(t)\) represents the axis storing features to be queried by the \(x\) coordinates. \(\textbf{A}^{X}_{res}\in\mathbb{R}^{rank\times D}\) denotes the temporal template axes, where \(D\) is the resolution of axis. The same explanation applies to \(\text{A}^{Y}(t)\) and \(\text{A}^{Z}(t)\).

Same to Eq.~\ref{eq:triplane_agg}, we use element-wise multiplication to aggregate features queried form different axes:
\begin{equation}
\begin{array}{c}
    \text{feat}(x,t)=\text{A}^X(t)[x],\\
    \text{feat}(y,t)=\text{A}^Y(t)[y],\\
    \text{feat}(z,t)=\text{A}^Z(t)[z],\\
    \text{feat}(x,y,z,t)=\text{feat}(x,t)\odot\text{feat}(y,t)\odot\text{feat}(z,t), \\
    \text{feat}(x,y,z,t)\in\mathbb{R}^{d}.
\end{array}
\end{equation}

We concatenate \(\text{feat}(x,y,z,t)\) from multiple levels along the channel dimension. The aggregated features from triplanes or triaxes are then decoded by a two-layer MLP into offsets and tangents. Apart from the construction of time-varying triplanes/triaxes, the rest of the pipeline follows the design of Hexplane~\cite{cao2023hexplane, fridovich2023k} and 4DGS~\cite{wu20234d}.

We follow the protocol of ResFields~\cite{mihajlovic2024ResFields} to initialize the temporal codes \(\textbf{v}_t\) using a normal distribution \(\textbf{V}\sim\mathcal{N}(0,10^{-2})\) to ensure training stability. We adopt the protocol outlined in K-Planes~\cite{fridovich2023k} for the initialization of \(\textbf{P}^{XY}\) and \(\textbf{A}^{X}\). The spatial resolution of the planes for each scene is determined following the approach described in 4DGS~\cite{wu20234d}.

\subsection{Experiment Details}
We adhere to the evaluation protocol provided by the dataset for dynamic scene reconstruction tasks. For the densification schedule, we follow the approach used in vanilla 3DGS~\cite{kerbl20233d}. With strong explicit regularization, we are able to use a larger learning rate. Specifically, we set the learning rate of the grids (Triplanes/Triaxes) to fifty times that of the Gaussian primitive's spatial learning rate, and the learning rate of the MLPs to five times that value. Additionally, we address several issues in the original implementation of 4DGS~\cite{wu20234d}, resulting in consistent improvements. We use the same reconstruction loss as vanilla 3DGS, and the hyperparameter settings from Eq. 10 in the main paper are listed in Tab.~\ref{tab:hyper_alpha_beta}. Specifically, the final loss function for dynamic scene reconstruction can be formulated as:
\begin{equation}
    \mathcal{L}=(1-\lambda)\text{L1}(I,\hat{I})+\lambda\mathcal{L}_{SSIM}(I,\hat{I})+\alpha\mathcal{L}_{v}+\beta\mathcal{L}_{acc},
\end{equation}
where \(I\) and \(\hat{I}\) represent ground truth and rendered images, respectively, and \(\lambda\) is set to 0.2 across all datasets.

Due to the high flexibility of 3DGS and the stochastic nature of the training process, differences between various motion representations are less pronounced. To address this and to reduce uncertainty while making comparisons more evident, we propose using scene flow interpolation to evaluate both spatial coherence and temporal consistency. We follow the evaluation protocol described in DOMA~\cite{zhang2024degrees} for long-term scene flow interpolation. Specifically, only 25\% of vertices and fewer than 25\% of timesteps are sampled as the training set, leaving ample room to validate the spatial and temporal performance of different methods. The names of selected scenes in DeformThings4D~\cite{li20214dcomplete} are listed in Tab.~\ref{tab:deform_4d_names}. Experimentally and consistent with DOMA's results, sinusoidal positional encoding (PE) performs poorly in this setup. Consequently, all compared methods, including ours, adopt Siren~\cite{sitzmann2020implicit} as the backbone. The reconstruction loss is defined as the L1 distance between predictions and ground truth, and the hyperparameter settings from Eq. 10 in the main paper are listed in Tab.~\ref{tab:hyper_alpha_beta}. Specifically, the final loss function of scene flow interpolation can be formulated as follows:
\begin{equation}
    \mathcal{L}=\text{L1}(\textbf{X},\hat{\textbf{X}})+\alpha\mathcal{L}_{v}+\beta\mathcal{L}_{acc},
\end{equation}
where \(\textbf{X}\) and \(\hat{\textbf{X}}\) represent the ground truth and predicted dynamic vertices positions, respectively.

\begin{table}
    \caption{The names of utilized sequences in DeformThings4D~\cite{li20214dcomplete} (humanoids) dataset.}
    \centering
    \begin{tabular}{l}
    \hline
    astra\_SambaDancing \\
    ironman\_BellyDance \\
    ironman\_LockingHipHopDance \\
    jackie\_SlideHipHopDance \\
    jody\_ThrillerPart3 \\
    kachujin\_HouseDancing \\
    michelle\_SlideHipHopDance \\
    nightshade\_ThrillerPart2 \\
    pumpkinhulk\_SwingDancing \\
    \hline
    \end{tabular}

    \label{tab:deform_4d_names}
\end{table}

\begin{table}
\centering
\caption{Values of the regularization hyperparameters in our final loss function are provided. Notably, adjusting these scales across datasets is natural to account for variations in spatial scales.}
\begin{tabular}{lcccc}
\hline
 & Scene Flow & NeRF-DS & Hyper-NeRF & Neu3D \\ \hline
\(\alpha\) & 1.0 & 0.01 & 0.05 & 0.0001 \\
\(\beta\) & 0.01 & 0.1 & 0.05 & 0.05 \\ \hline
\end{tabular}
\label{tab:hyper_alpha_beta}
\end{table}

\section{Additional Comparison}
\subsection{Efficiency}
\begin{table}
\centering
\small
\setlength{\tabcolsep}{2.3pt}
\caption{Comparison of render speed (FPS), GPU usage (Mem.) in megabytes, and number of points (\#G) in \(1\times10^3\) on NeRF-DS~\cite{yan2023nerf} dataset.}
\begin{tabular}{lcccccc}
\hline
Method & FPS & \begin{tabular}[c]{@{}c@{}}Mem. \\ (MB)\end{tabular} & \#G (K) & PSNR & LPIPS & M.'s \textit{I} \\ \hline
Def.3D~\cite{yang2023deformable} & 32.1 & 3664 & 245.2 & 23.75 & \cellcolor[HTML]{FCE5CD}13.28 & 0.706 \\
4DGS*~\cite{wu20234d} & 184.3 & \cellcolor[HTML]{F4CCCC}990 & 137.2 & 23.56 & 14.36 & 0.788 \\ \hline
SF-Triaxes & \cellcolor[HTML]{FCE5CD}429.3 & 1357 & 161.1 & \cellcolor[HTML]{FCE5CD}24.36 & 14.01 & 0.926 \\
SF-Triplanes & \cellcolor[HTML]{F4CCCC}449.6 & \cellcolor[HTML]{FCE5CD}1343 & 160.6 & 24.32 & 13.42 & \cellcolor[HTML]{FCE5CD}0.931 \\
SF-PE-ResFields & 295.2 & 1686 & 176.6 & \cellcolor[HTML]{F4CCCC}24.40 & \cellcolor[HTML]{F4CCCC}12.85 & \cellcolor[HTML]{F4CCCC}0.964 \\ \hline
\end{tabular}
\label{tab:nerfds_efficiency}
\end{table}

\begin{table}
\centering
\small
\setlength{\tabcolsep}{2.3pt}
\caption{Comparison of render speed (FPS), GPU usage (Mem.) in megabytes, and number of points (\#G) in \(1\times10^3\) on Neu3DS~\cite{li2022neural} dataset.}
\begin{tabular}{lcccccc}
\hline
Method & FPS & \begin{tabular}[c]{@{}c@{}}Mem.\\ (MB)\end{tabular} & \#G (K) & PSNR & LPIPS & M.'s \textit{I} \\ \hline
4DGS~\cite{wu20234d} & \cellcolor[HTML]{F4CCCC}148.8 & \cellcolor[HTML]{F4CCCC}1687 & 150.4 & \cellcolor[HTML]{FCE5CD}31.59 & 8.64 & 0.898 \\
Def.3D~\cite{yang2023deformable} & 40.1 & \cellcolor[HTML]{FCE5CD}2012 & 144.5 & 31.29 & 8.90 & 0.846 \\
Grid4D~\cite{xu2024grid4d} & 117.6 & 2945 & 236.8 & 31.50 & 9.78 & 0.594 \\ \hline
SF-Triplanes & \cellcolor[HTML]{FCE5CD}135.9 & 2178 & 148.7 & \cellcolor[HTML]{F4CCCC}31.60 & \cellcolor[HTML]{F4CCCC}8.59 & \cellcolor[HTML]{FCE5CD}0.930 \\
SF-PE-ResFields & 30.2 & 2327 & 146.3 & 31.47 & \cellcolor[HTML]{FCE5CD}8.63 & \cellcolor[HTML]{F4CCCC}0.937 \\ \hline
\end{tabular}
\label{tab:neu3d_efficiency}
\end{table}

We provide efficiency comparisons in Tables~\ref{tab:nerfds_efficiency}–\ref{tab:neu3d_efficiency}. Our variant, SF-Triplanes, achieves a frames per second (FPS) rate that is sufficient for real-time applications. SF-PE-ResFields exhibits slower inference speed, as shown in Tab.\ref{tab:neu3d_efficiency}. We believe this limitation can potentially be addressed by reducing the number of parameters, an advantage of ResFields\cite{mihajlovic2024ResFields} that we have not fully explored in this work.


\subsection{Results on D-NeRF dataset}
\begin{table*}
\centering
\small
\setlength{\tabcolsep}{2.5pt}
\caption{Supplemental comparison on the D-NeRF~\cite{pumarola2021d} dataset (\(800\times800\) resolution): As shown in the comparison between 4DGS and its variant 4DGS*, particularly in Moran's \textit{I} evaluation, perturbation in Grid4D serves as an effective regularization for grid-based methods in this synthetic dataset. As mentioned, we use the number of testing frames, \(T\), to calculate Moran's \textit{I} results. However, due to this dataset's limited number of frames, we set \(T_{test}=150\) for all scenes in this comparison.}
\begin{tabular}{lccc|ccc|ccc|ccclll}
\cline{1-13}
 & \multicolumn{3}{c|}{Bouncingballs} & \multicolumn{3}{c|}{Hellwarrior} & \multicolumn{3}{c|}{Hook} & \multicolumn{3}{c}{Jumpingjacks} &  &  &  \\
\multirow{-2}{*}{Method} & PSNR\(\uparrow\) & SSIM\(\uparrow\) & M.'s \textit{I}\(\uparrow\) & PSNR\(\uparrow\) & SSIM\(\uparrow\) & M.'s \textit{I}\(\uparrow\) & PSNR\(\uparrow\) & SSIM\(\uparrow\) & M.'s \textit{I}\(\uparrow\) & PSNR\(\uparrow\) & SSIM\(\uparrow\) & M.'s \textit{I}\(\uparrow\) &  &  &  \\ \cline{1-13}
4DGS~\cite{wu20234d} & 36.28 & 99.0 & 0.543 & 39.08 & 97.9 & 0.635 & 33.61 & 97.6 & 0.635 & 36.32 & 98.8 & 0.686 &  &  &  \\
4DGS*~\cite{wu20234d} & 37.63 & 99.2 & 0.633 & 39.83 & 98.2 & 0.725 & 34.64 & 98.0 & 0.731 & 36.77 & 98.9 & 0.767 &  &  &  \\
Def.3D~\cite{yang2023deformable} & \cellcolor[HTML]{FFF2CC}40.70 & \cellcolor[HTML]{FFF2CC}99.5 & 0.715 & 41.41 & 98.7 & 0.876 & 37.21 & 98.6 & 0.882 & 37.60 & 98.9 & 0.867 &  &  &  \\
Grid4D~\cite{xu2024grid4d} & \cellcolor[HTML]{F4CCCC}42.62 & \cellcolor[HTML]{F4CCCC}99.6 & \cellcolor[HTML]{F4CCCC}0.974 & \cellcolor[HTML]{F4CCCC}42.85 & \cellcolor[HTML]{F4CCCC}99.1 & \cellcolor[HTML]{F4CCCC}0.974 & \cellcolor[HTML]{F4CCCC}38.89 & \cellcolor[HTML]{FCE5CD}99.0 & \cellcolor[HTML]{F4CCCC}0.979 & \cellcolor[HTML]{F4CCCC}39.37 & \cellcolor[HTML]{F4CCCC}99.3 & \cellcolor[HTML]{F4CCCC}0.979 &  &  &  \\ \cline{1-13}
SF-Tripanes & 40.17 & 99.5 & \cellcolor[HTML]{FFF2CC}0.845 & \cellcolor[HTML]{FFF2CC}41.95 & \cellcolor[HTML]{FFF2CC}98.9 & \cellcolor[HTML]{FFF2CC}0.945 & \cellcolor[HTML]{FCE5CD}38.44 & \cellcolor[HTML]{F4CCCC}99.0 & \cellcolor[HTML]{FFF2CC}0.928 & \cellcolor[HTML]{FCE5CD}39.13 & \cellcolor[HTML]{FCE5CD}99.3 & \cellcolor[HTML]{FFF2CC}0.929 &  &  &  \\
SF-PE-ResFields & \cellcolor[HTML]{FCE5CD}41.13 & \cellcolor[HTML]{FCE5CD}99.5 & \cellcolor[HTML]{FCE5CD}0.926 & \cellcolor[HTML]{FCE5CD}42.10 & \cellcolor[HTML]{FCE5CD}98.9 & \cellcolor[HTML]{FCE5CD}0.957 & \cellcolor[HTML]{FFF2CC}38.03 & \cellcolor[HTML]{FFF2CC}99.0 & \cellcolor[HTML]{FCE5CD}0.955 & \cellcolor[HTML]{FFF2CC}38.69 & \cellcolor[HTML]{FFF2CC}99.2 & \cellcolor[HTML]{FCE5CD}0.941 &  &  &  \\ \hline
 & \multicolumn{3}{c|}{Mutant} & \multicolumn{3}{c|}{Standup} & \multicolumn{3}{c|}{T-Rex} & \multicolumn{3}{c|}{Average} & \multicolumn{1}{c}{} & \multicolumn{1}{c}{} & \multicolumn{1}{c}{} \\
\multirow{-2}{*}{Method} & PSNR\(\uparrow\) & SSIM\(\uparrow\) & M.'s \textit{I}\(\uparrow\) & PSNR\(\uparrow\) & SSIM\(\uparrow\) & M.'s \textit{I}\(\uparrow\) & PSNR\(\uparrow\) & SSIM\(\uparrow\) & M.'s \textit{I}\(\uparrow\) & PSNR\(\uparrow\) & SSIM\(\uparrow\) & \multicolumn{1}{c|}{M.'s \textit{I}\(\uparrow\)} & \multicolumn{1}{c}{\multirow{-2}{*}{FPS}} & \multicolumn{1}{c}{\multirow{-2}{*}{Mem.}} & \multicolumn{1}{c}{\multirow{-2}{*}{\#G(K)}} \\ \hline
4DGS~\cite{wu20234d} & 39.43 & 99.2 & 0.876 & 41.73 & 99.3 & 0.694 & 34.22 & FALSE & 0.73 & 37.24 & 98.6 & \multicolumn{1}{c|}{0.686} & \multicolumn{1}{c}{\cellcolor[HTML]{F4CCCC}} & \multicolumn{1}{c}{\cellcolor[HTML]{F4CCCC}} & \multicolumn{1}{c}{} \\
4DGS*~\cite{wu20234d} & 40.32 & 99.3 & 0.926 & 42.42 & 99.4 & 0.790 & 35.51 & 99.1 & 0.820 & 38.16 & 98.9 & \multicolumn{1}{c|}{0.770} & \multicolumn{1}{c}{\multirow{-2}{*}{\cellcolor[HTML]{F4CCCC}180}} & \multicolumn{1}{c}{\multirow{-2}{*}{\cellcolor[HTML]{F4CCCC}583}} & \multicolumn{1}{c}{\multirow{-2}{*}{58}} \\
Def.3D~\cite{yang2023deformable} & \cellcolor[HTML]{FCE5CD}42.65 & \cellcolor[HTML]{FFF2CC}99.5 & \cellcolor[HTML]{FFF2CC}0.965 & \cellcolor[HTML]{FCE5CD}44.25 & \cellcolor[HTML]{FFF2CC}99.5 & 0.891 & \cellcolor[HTML]{FCE5CD}38.17 & \cellcolor[HTML]{FFF2CC}99.3 & \cellcolor[HTML]{FFF2CC}0.912 & \cellcolor[HTML]{FCE5CD}40.28 & 99.1 & \multicolumn{1}{c|}{0.873} & \multicolumn{1}{c}{60} & \multicolumn{1}{c}{2852} & \multicolumn{1}{c}{133} \\
Grid4D~\cite{xu2024grid4d} & \cellcolor[HTML]{F4CCCC}43.94 & \cellcolor[HTML]{F4CCCC}99.6 & \cellcolor[HTML]{F4CCCC}0.986 & \cellcolor[HTML]{F4CCCC}46.28 & \cellcolor[HTML]{F4CCCC}99.7 & \cellcolor[HTML]{F4CCCC}0.983 & \cellcolor[HTML]{F4CCCC}40.01 & \cellcolor[HTML]{F4CCCC}99.5 & \cellcolor[HTML]{F4CCCC}0.990 & \cellcolor[HTML]{F4CCCC}42.00 & \cellcolor[HTML]{F4CCCC}99.4 & \multicolumn{1}{c|}{\cellcolor[HTML]{F4CCCC}0.981} & \multicolumn{1}{c}{\cellcolor[HTML]{FCE5CD}152} & \multicolumn{1}{c}{3165} & \multicolumn{1}{c}{162} \\ \hline
SF-Tripanes & \cellcolor[HTML]{FFF2CC}41.68 & \cellcolor[HTML]{FCE5CD}99.5 & 0.956 & \cellcolor[HTML]{FFF2CC}43.75 & \cellcolor[HTML]{FCE5CD}99.6 & \cellcolor[HTML]{FFF2CC}0.932 & 36.81 & 99.2 & 0.906 & \cellcolor[HTML]{FFF2CC}40.28 & \cellcolor[HTML]{FCE5CD}99.3 & \multicolumn{1}{c|}{\cellcolor[HTML]{FFF2CC}0.920} & \multicolumn{1}{c}{\cellcolor[HTML]{FFF2CC}140} & \multicolumn{1}{c}{\cellcolor[HTML]{FCE5CD}644} & \multicolumn{1}{c}{59} \\
SF-PE-ResFields & 41.47 & 99.5 & \cellcolor[HTML]{FCE5CD}0.967 & 42.07 & 99.4 & \cellcolor[HTML]{FCE5CD}0.952 & \cellcolor[HTML]{FFF2CC}37.72 & \cellcolor[HTML]{FCE5CD}99.4 & \cellcolor[HTML]{FCE5CD}0.951 & 40.17 & \cellcolor[HTML]{FFF2CC}99.3 & \multicolumn{1}{c|}{\cellcolor[HTML]{FCE5CD}0.950} & \multicolumn{1}{c}{34} & \multicolumn{1}{c}{\cellcolor[HTML]{FFF2CC}880} & \multicolumn{1}{c}{78} \\ \hline
\end{tabular}
\label{tab:dnerf}
\end{table*}

As mentioned at the end of Sec. 3.2 in the main paper, we provide comparisons on the synthetic D-NeRF~\cite{pumarola2021d} dataset, as shown in Tab.\ref{tab:dnerf}. The perturbation introduced by Grid4D\cite{xu2024grid4d} effectively resolves artifacts in 4DGS~\cite{wu20234d}, achieving even better evaluations of Moran's \textit{I} than our method. We interpret this as a result of improved regularization and reduced overlap, with hash encoding enabling much higher resolution grids than the standard grid used in Hexplane~\cite{cao2023hexplane, fridovich2023k}.

However, as demonstrated in the main paper, this regularization shows limited effectiveness in real-world scenarios. We interpret this limitation as stemming from the fact that, while 4D signals are factorized to avoid hash collisions, the coupling of spatial and temporal signals still persists (e.g., grid \(XYT\)). The perturbation works effectively in the synthetic dataset, where precise camera positions, high frame rates, and simple motions provide ideal conditions. However, in long-term cases with complex motions, the hashing behavior in the mixed spatiotemporal encoding becomes unpredictable. To address this, we believe that time-variant spatial hash encoding, combined with a more advanced spline representation, holds promise for overcoming these challenges.

\subsection{Comparison with SC-GS}
\begin{figure}
    \centering
    \includegraphics[width=\linewidth]{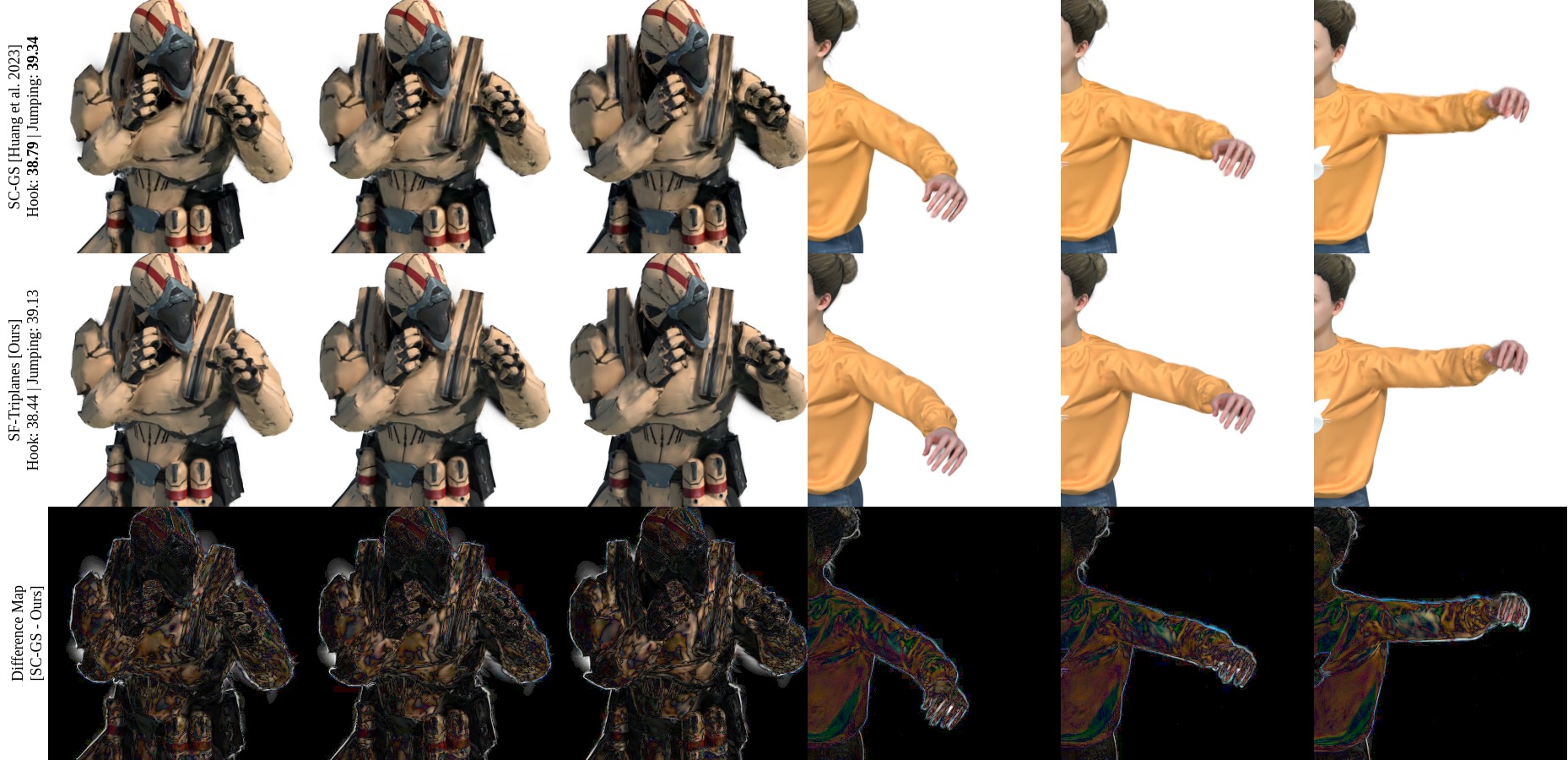}
    \caption{We present a visual comparison between our SF-Triplanes and SC-GS~\cite{huang2023sc}. The images for SC-GS are sourced from videos available on its official website, and the PSNR evaluation is adopted from Grid4D~\cite{xu2024grid4d}. We want to note that while PSNR (Peak Signal-to-\textbf{Noise} Ratio) is well-suited for assessing additive degradations, it may not be the most appropriate metric for evaluating GS-based scene reconstruction, where discrepancies in shape and texture are more prevalent, as shown in the difference maps.}
    \label{fig:sc_gs_dnerf}
\end{figure}

As shown in Fig.\ref{fig:sc_gs_dnerf}\footnote{https://yihua7.github.io/SC-GS-web/}, SC-GS~\cite{huang2023sc} outperforms our method on the D-NeRF~\cite{pumarola2021d} dataset. When modeling a single avatar, SC-GS~\cite{huang2023sc}’s sparse control paradigm proves to be both suitable and principled. We believe that combining SC-GS~\cite{huang2023sc} with our spline-based representation holds strong potential for motion editing applications (as illustrated in Fig. 8 of the main paper), extending beyond static object deformation or frame-by-frame processing.

\begin{figure}
    \centering
    \setlength{\tabcolsep}{0.0pt}
    \begin{tabular}{cc}
        SC-GS~\cite{huang2023sc} & SF-PE-ResFields (ours) \\
        \includegraphics[width=0.5\linewidth]{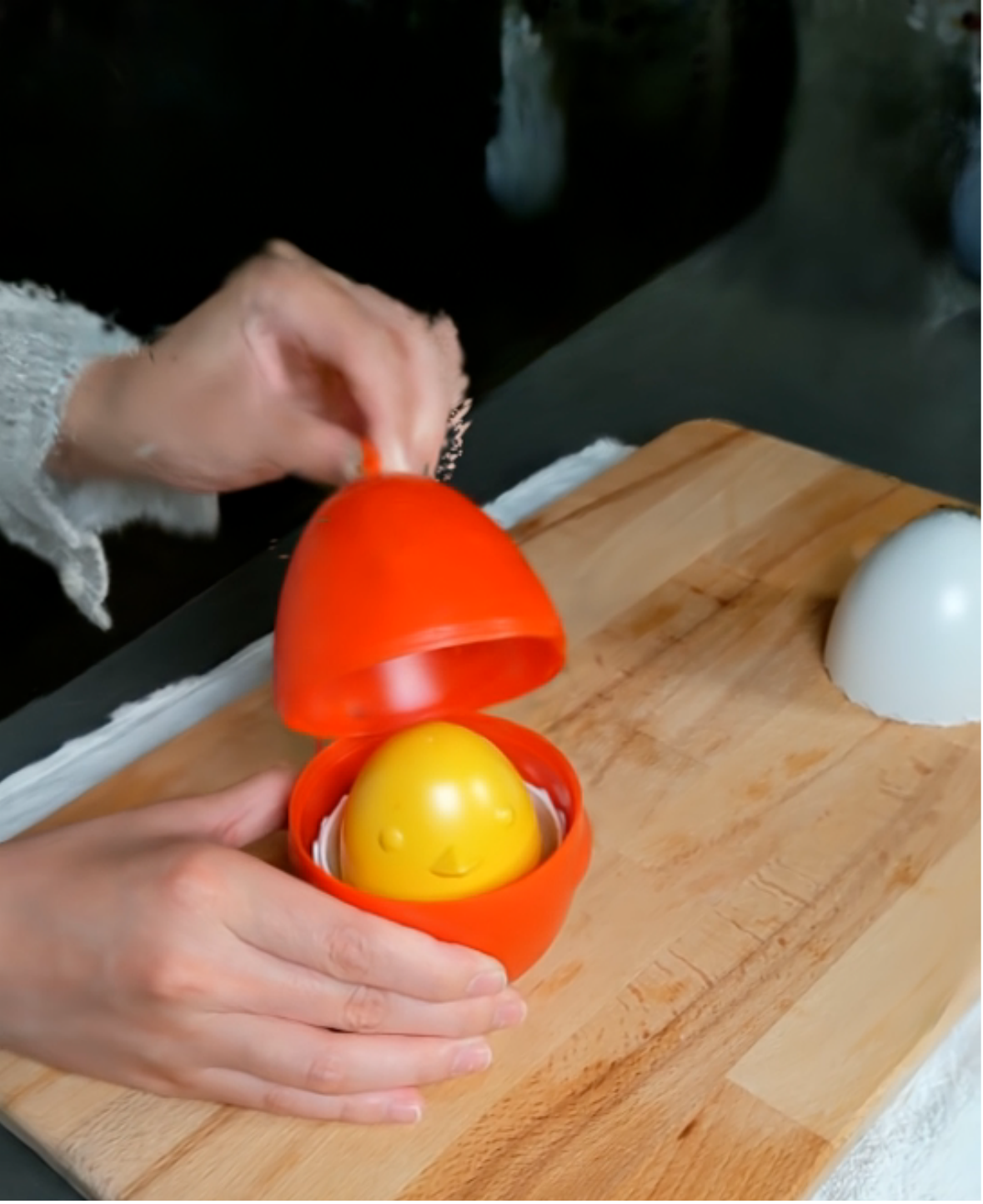} &
        \includegraphics[width=0.5\linewidth]{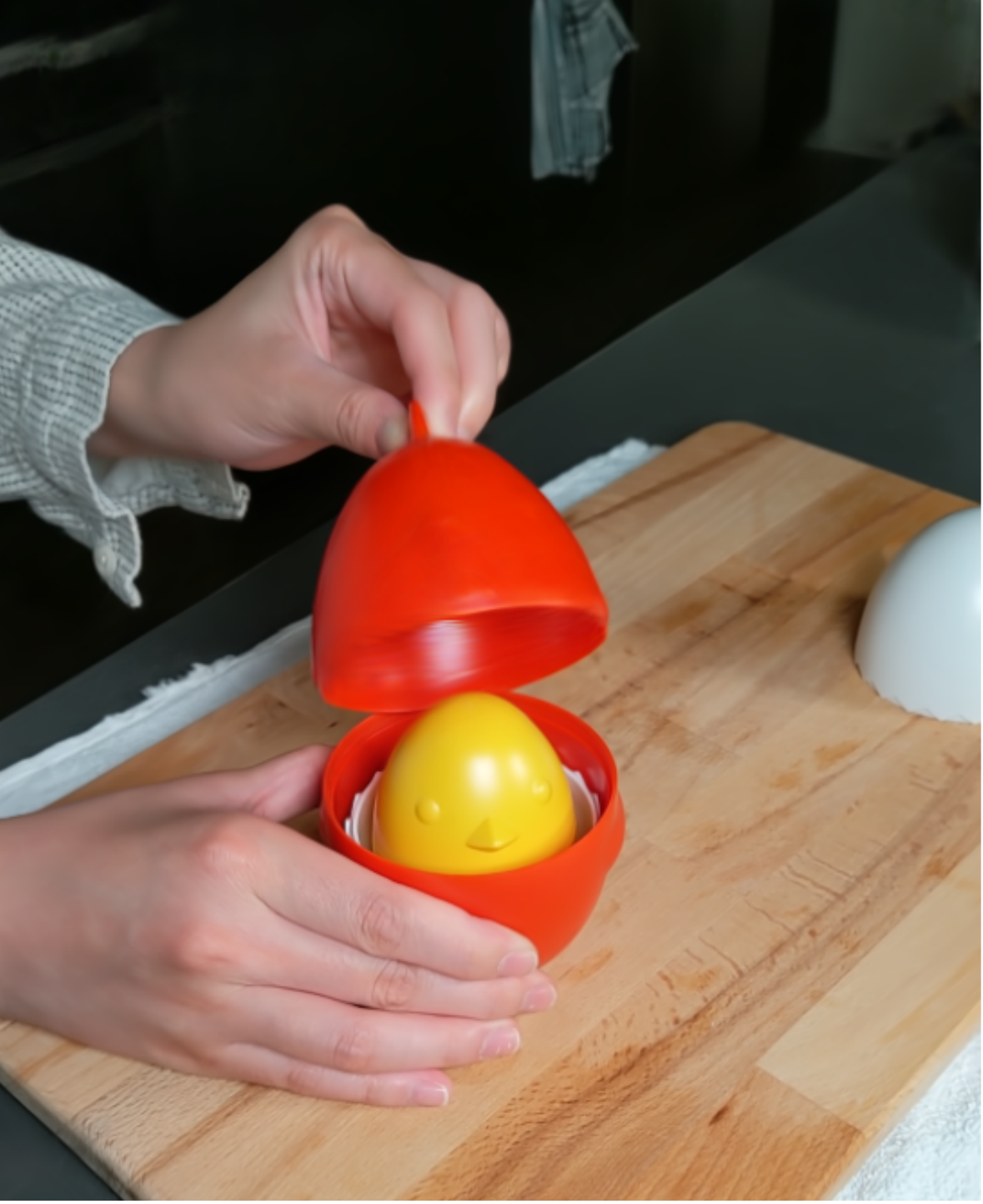}
    \end{tabular}
    \caption{Qualitative comparison of SC-GS~\cite{huang2023sc} and our method. The compared frame is taken from the official rendered video.}
    \label{fig:sc_gs_chick}
\end{figure}

An additional visual comparison between our method and SC-GS~\cite{huang2023sc} on the real-world dataset HyperNeRF~\cite{park2021hypernerf} is presented in Fig.~\ref{fig:sc_gs_chick}\footnote{https://yihua7.github.io/SC-GS-web/materials/videos/hypernerf3.mp4}. We direct readers to the supplementary video for a demonstration of the imprecise camera issue in Hyper-NeRF~\cite{park2021hypernerf}, which becomes evident from the fixed camera view. 

\section{Failure cases}
\begin{figure}
    \centering
    \includegraphics[width=\linewidth]{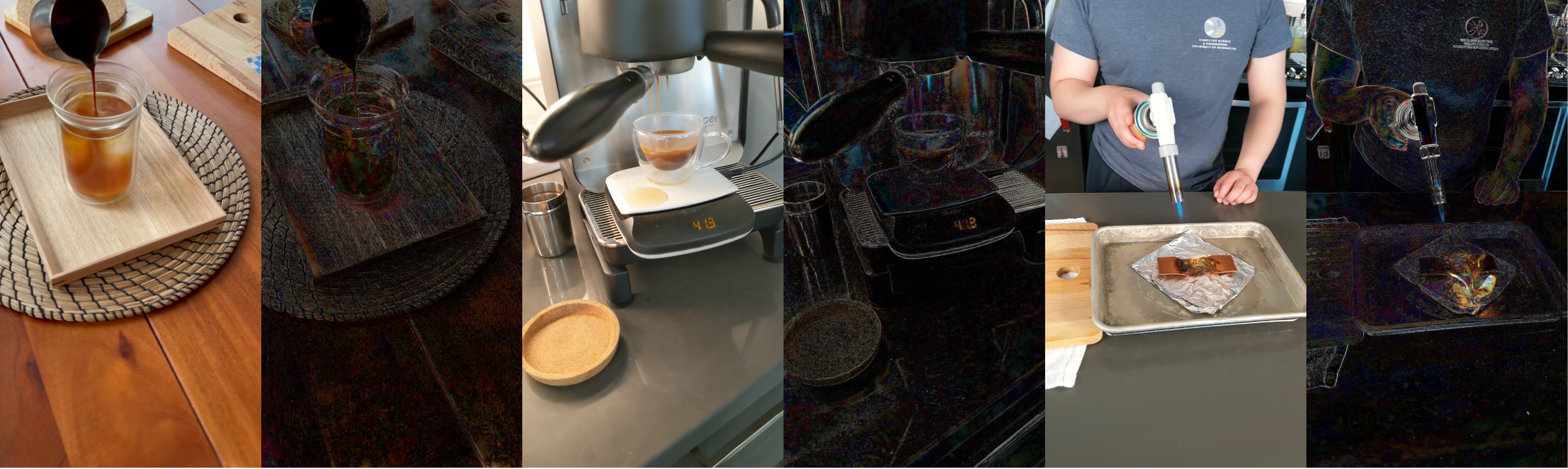}
    \caption{We present rendered results of scenes with non-rigid motions and their differences from the ground truth. As demonstrated in the scenes \textit{americano} and \textit{espresso} from the Hyper-NeRF~\cite{park2021hypernerf} (misc) dataset, fluids with solid colors can still be effectively reconstructed using \(\mathcal{L}_{v}\) constraints without additional fine-tuning of \(\alpha\) in Eq.10 of the main paper. However, our method struggles with semi-transparent and fine-grained volumes, as seen in the \textit{torch\_chocolate} scene. Notably, the flames in the \textit{flame\_steak} scene are well-reconstructed, as shown in the teaser in the main paper.}
    \label{fig:fluid}
\end{figure}

\begin{figure}
    \centering
    \includegraphics[width=\linewidth]{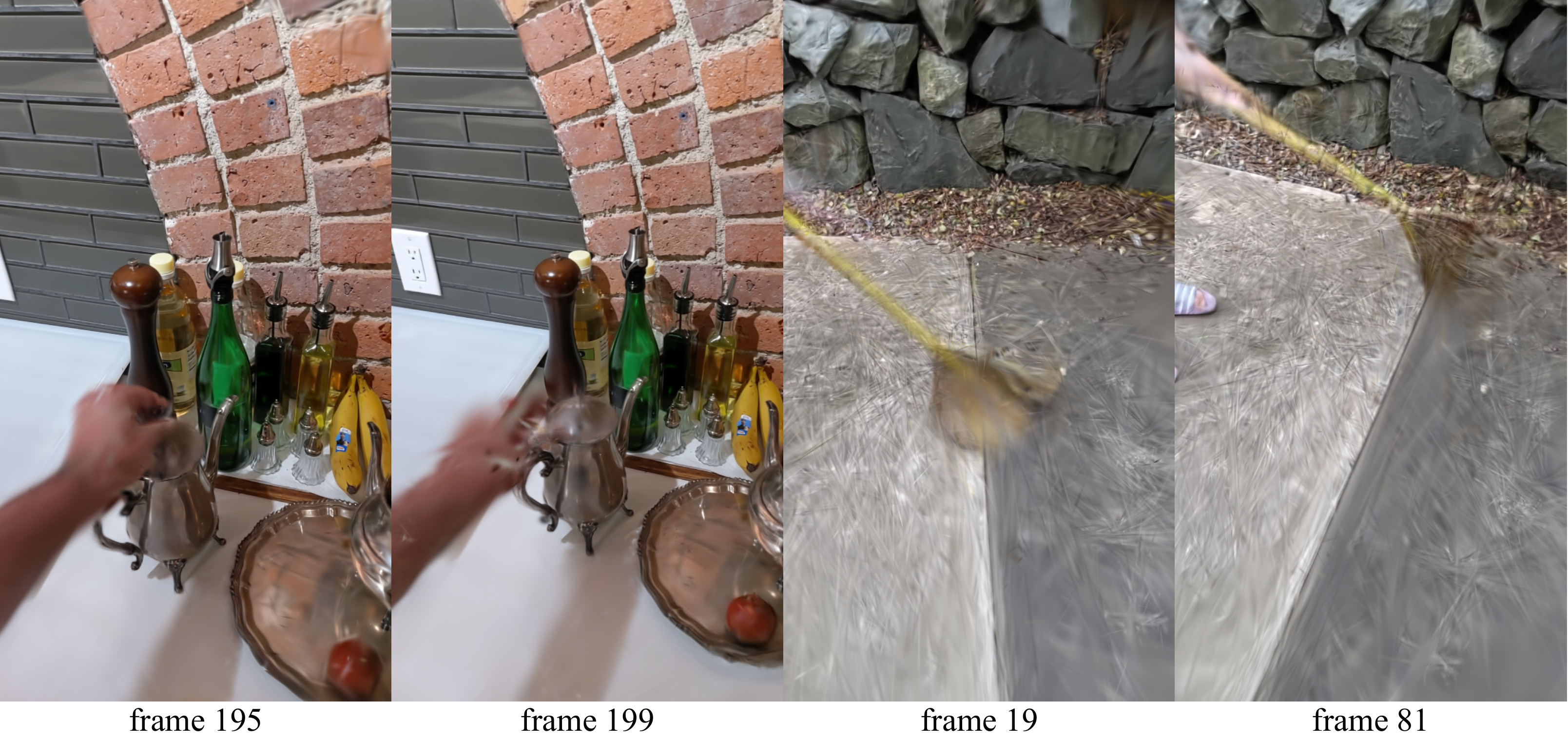}
    \caption{Our failure cases from the Hyper-NeRF~\cite{park2021hypernerf} dataset. While the hand in \textit{aleks-teapot} can be reasonably reconstructed when it pauses over the teapot, it turns into floaters when moving out of the scene. Similarly, in the \textit{broom2} scene, the moving broom fails to be reconstructed accurately, and the static floor remains unreconstructed.}
    \label{fig:hyper_fail}
\end{figure}

\begin{figure}
    \centering
    \includegraphics[width=\linewidth]{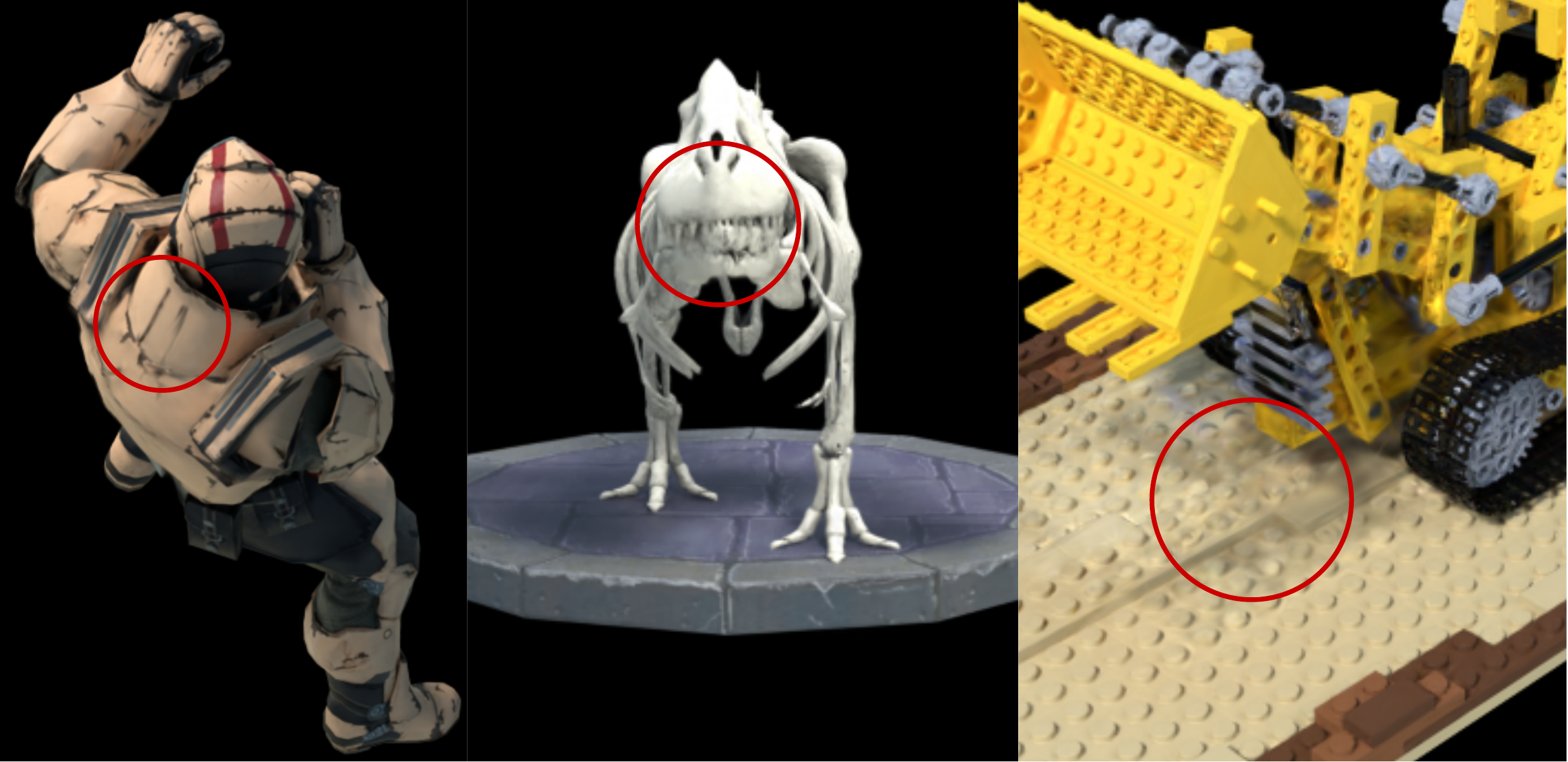}
    \caption{Our failure cases from the D-NeRF~\cite{pumarola2021d} dataset highlight challenges with specific view directions. We anticipate that future work will address these difficulties.}
    \label{fig:dnerf_fail}
\end{figure}

We present typical artifacts and failure cases of our method. Consistent with most canonical-deformation-based approaches, our method faces challenges in scenes from Hyper-NeRF~\cite{park2021hypernerf} with extremely imprecise camera settings (\textit{broom2}) and objects moving in and out of the scene (\textit{aleks-teapot}), as shown in Fig.\ref{fig:hyper_fail}. Additionally, as illustrated in Fig.\ref{fig:dnerf_fail}, abnormal floaters may still appear even with strong explicit regularization. Sudden motions in \textit{cook\_spinach} pose difficulties for most existing methods. We attribute this to the balanced allocation of training resources across frames, which leads to insufficient attention being given to difficult-to-fit frames during training. While we are able to reconstruct non-rigid motions, such as fluids (as shown in Fig.~\ref{fig:fluid}), high-frequency flames remain a challenge for our method.

\section{Additional Discussion}

While we eliminate heuristics in determining the number and temporal positions of knots, the dense number of knots increases the risk of jittering in certain corner cases. Although the regularization term on acceleration, \(\mathcal{L}_{acc}\), effectively mitigates such temporal artifacts, it also introduces the risk of over-smoothing and underfitting. While we experimented with quintic polynomials (\(C_2\) smooth in theory), we observed only slight improvements in practical cases. For completeness, we provide the formulation of explicit quintic spline:
\begin{equation}
\begin{array}{r}
     p(\bar{t})=(-6\bar{t}^5+15\bar{t}^4-10\bar{t}^3+1)p_0\\
     +(-3\bar{t}^5+8\bar{t}^4-6\bar{t}^3+\bar{t})m_0\\
     +(-\frac{1}{2}\bar{t}^5+\frac{3}{2}\bar{t}^4-\frac{3}{2}\bar{t}^3+\frac{1}{2}\bar{t}^2)a_0\\
     +(6\bar{t}^5-15\bar{t}^4+10\bar{t}^3)p_1\\
     +(-3\bar{t}^5+7\bar{t}^4-4\bar{t}^3)m_1\\
     +(\frac{1}{2}\bar{t}^5-\bar{t}^4+\frac{1}{2}\bar{t}^3)a_1,
\end{array}
\end{equation}
where \(a_0\) and \(a_1\) denote the acceleration of the starting and ending knots. We believe that representing trajectories by blending multiple local bases, addressing both temporal sparsity and fitting accuracy simultaneously, is a promising direction for future work.

While we achieve real-time rendering performance with minimal GPU memory overhead, our training process is slower than per-timestep estimation methods, primarily due to the need for two inferences of the starting and ending knots. We believe this issue can be resolved by optimizing the backpropagation of gradients, such as through a fused operation.